\PassOptionsToPackage{dvipsnames,table}{xcolor}
\documentclass{article} % For LaTeX2e
\usepackage{iclr2025_conference,times}

\usepackage{amsmath,amsfonts,bm}

\def\eqref#1{equation~\ref{#1}}
\def\1{\bm{1}}

\DeclareMathAlphabet{\mathsfit}{\encodingdefault}{\sfdefault}{m}{sl}
\SetMathAlphabet{\mathsfit}{bold}{\encodingdefault}{\sfdefault}{bx}{n}

\newcommand{\R}{\mathbb{R}}

\usepackage{hyperref}
\usepackage{url}

\hypersetup{
  breaklinks,
  colorlinks,
  linkcolor=BlueViolet,
  citecolor=BlueViolet,
  urlcolor=BlueViolet,
}

\usepackage[T1]{fontenc}    % use 8-bit T1 fonts
\usepackage{booktabs}       % professional-quality tables
\usepackage{amsfonts}       % blackboard math symbols
\usepackage{nicefrac}       % compact symbols for 1/2, etc.
\usepackage{microtype}      % microtypography
\usepackage[dvipsnames]{xcolor}
\usepackage{tcolorbox}      % For colored boxes

\usepackage{amsmath}
\usepackage{amssymb}
\usepackage{mathtools}
\usepackage{amsthm}
\usepackage{parskip}

\usepackage{graphicx}
\usepackage{subfigure}
\usepackage{wrapfig}
\usepackage{tabularx}
\usepackage{adjustbox}

\usepackage{multirow}

\usepackage[capitalize]{cleveref}
\usepackage[utf8]{inputenc}

\usepackage{enumitem}

\usepackage[small]{caption}
\usepackage{titlesec}

\newenvironment{takeaway}{\begin{center}\begin{tcolorbox}[width=\textwidth, colback=ForestGreen!9!white, colframe=black, boxrule=1pt, arc=2mm, auto outer arc, left=5pt, right=5pt, top=2pt, bottom=2pt]}{\end{tcolorbox}\end{center}}

\title{Exploring the Effectiveness of Object-Centric Representations in Visual Question Answering: Comparative Insights with Foundation Models}

\author{%
    Amir Mohammad Karimi Mamaghan$^1$\quad
    Samuele Papa$^2$\quad
    Karl Henrik Johansson$^{1,3}$ \\
    \textbf{Stefan Bauer$^{4,5}$\quad
    Andrea Dittadi$^{4,5,6}$} \vspace{0.8em} \\
    \normalfont $^1$KTH Royal Institute of Technology\quad\ 
    $^2$University of Amsterdam\quad\ 
    $^3$Digital Futures \\
    $^4$Helmholtz AI\quad\ 
    $^5$Technical University of Munich\quad\ 
    $^6$MPI for Intelligent Systems, Tübingen
}

\iclrfinalcopy % Uncomment for camera-ready version, but NOT for submission.
\begin{document}

\maketitle

\begin{abstract}
Object-centric (OC) representations, which model visual scenes as compositions of discrete objects, have the potential to be used in various downstream tasks to achieve systematic compositional generalization and facilitate reasoning. However, these claims have yet to be thoroughly validated empirically.
Recently, foundation models have demonstrated unparalleled capabilities across diverse domains, from language to computer vision, positioning them as a potential cornerstone of future research for a wide range of computational tasks.
In this paper, we conduct an extensive empirical study on representation learning for downstream Visual Question Answering (VQA), which requires an accurate compositional understanding of the scene. 
We thoroughly investigate the benefits and trade-offs of OC models and alternative approaches including large pre-trained foundation models on both synthetic and real-world data, ultimately identifying a promising path to leverage the strengths of both paradigms. 
The extensiveness of our study, encompassing over 600 downstream VQA models and 15 different types of upstream representations, also provides several additional insights that we believe will be of interest to the community at large.
\end{abstract}

\section{Introduction}
\label{sec:introduction}
Object-centric (OC) learning aims to represent the physical world's inherent structure, assuming visual scenes consist of entities or objects and employing this as an inductive bias for neural networks \citep{goyal2019recurrent,locatello2020object, lin2020space, burgess2019monet, singh2022illiterate, singh2022simple, seitzer2023bridging, wu2023slotdiffusion, jiang2023object, lowe2023rotating}. Applied in various domains like visual reasoning \citep{zfchen2021iclr, Ding2021, wu2022slotformer, ding2021dynamic, santoro2017simple, webb2023systematic, mondal2023learning} and image and video generation \citep{chen2021roots, singh2022illiterate, singh2022simple, elsayed2022savi++, jabri2023dorsal}, these representations play a crucial role in capturing compositional and causal structures, with the potential to improve the generalizability and interpretability of AI algorithms \citep{lake2017building, scholkopf2021toward, brady2023provably, kim2023imagine, jung2024learning}. Breaking down scenes into conceptual elements corresponding to causal factors aligns with the idea that causal models play a crucial role in achieving human-level generalization \citep{pearl2009causality, peters2017elements, mansouri2023object, liu2023causal}.

While OC representations thus provide a way of representing the state of a visual scene, a comprehensive understanding of these representations is still an ongoing exploration. 
Recently, there have been several works on evaluating OC representations.
Some studies evaluate object-centric models in terms of reconstruction and segmentation accuracy, and quantify the quality and information content of object representations via a downstream object property prediction task \citep{Dittadi2021, Papa2022a}. Arguing that a major goal of representation learning is to facilitate downstream tasks, \citet{yoon2023investigation} focuses on the evaluation of the representations in reinforcement learning, which requires a thorough understanding of the environment in terms of objects and the relations between them. However, a more direct quantification of the role of object-centric representations for reasoning is still missing.\looseness=-1

In the rapidly evolving landscape of deep learning, foundation models, often characterized by self-supervised and large-scale pre-training, have demonstrated unparalleled capabilities in generalization and zero-shot learning, showcasing their prowess in tasks across diverse domains from natural language processing to computer vision \citep{kirillov2023segany, wen2023foundationpose, touvron2023llama, brown2020language, chowdhery2023palm, huang2023stu, borsos2023audiolm, yang2023dawn, jumper2021highly, rombach2022high}.
Despite their widespread success, foundation models have not been comprehensively analyzed and compared with OC models. 

In this paper, we aim to take one step towards understanding the relevance of object-centric representations in the era of foundation models by evaluating them on visual reasoning tasks through a simple framework.
More specifically, \textbf{our main contributions} are the following: 
\begin{itemize}[itemsep=-4pt,topsep=-2pt,leftmargin=10pt]
    \item We conduct a large empirical study on representation learning for downstream Visual Question Answering (VQA) \citep{antol2015vqa, johnson2017clevr} on three synthetic and two real-world multi-object datasets. 
    In our extensive evaluation, we train overall 684 downstream transformer models for VQA, involving 15 different types of upstream representation models, ranging from VAEs to state-of-the-art OC methods to large pre-trained foundation models.
    \item We identify and investigate the trade-offs between large foundation models and OC models. We observe that, without any fine-tuning or hyperparameter adjustment, foundation models perform comparably to the top-performing OC models. On the other hand, they typically require more compute and larger downstream models.
    We find that applying the OC inductive bias to foundation models effectively achieves the best of both worlds, 
    reducing the downstream computational needs while achieving comparable or better performance and obtaining more explicit representations.
    \item We present several additional insights regarding, among other things, the correlation between performances on VQA and a simpler downstream task, the relationship between upstream and downstream performance of OC models, the effect of training set size on VQA performance, the difference between different question types, and a deeper analysis of the global (single-vector) representations of traditional VAEs.
\end{itemize}

\section{Related Works}
\paragraph{Object-Centric Learning.}
Object-centric (OC) learning has gained attention over the past few years \citep{goyal2019recurrent, singh2021structured, singh2022illiterate, singh2022simple, seitzer2023bridging, wu2021generative, wu2023slotdiffusion, jiang2023object, eslami2016attend, crawford2019spatially, kosiorek2018sequential, JiangJanghorbaniDeMeloAhn2020SCALOR, dittadi2019lavae, engelcke2019genesis, engelcke2021genesis, lin2020improving, lin2020space, greff2017neural, gregor2015draw, yuan2019generative, locatello2020object, burgess2019monet, jabri2023dorsal, chen2021roots, kipf2019contrastive, kipf2022conditional, elsayed2022savi++, lowe2023rotating, kori2023grounded, sajjadi2022object, daniel2022unsupervised}. OC models aim to learn visual representations without supervision by treating each image as a composition of objects. Among them, Slot Attention \citep{locatello2020object} stands out as a popular model and a crucial component in several recent state-of-the-art models. Numerous enhancements have been proposed, including improvements of the Slot Attention module \citep{biza2023invariant, jia2023improving, Majellaro2024explicitly} or adding additional modules on top \citep{kim2023shepherding}, using a transformer decoder instead of the original mixture-based decoder \citep{singh2021structured, singh2022simple}, replacing the CNN backbone with a pre-trained model \citep{seitzer2023bridging}, and integrating diffusion models with Slot Attention \citep{jiang2023object, wu2023slotdiffusion, jabri2023dorsal}.

\paragraph{Evaluation of Object-Centric Representations.}
OC methods have been applied in several works in visual reasoning \citep{zfchen2021iclr, Ding2021, wu2022slotformer, ding2021dynamic, santoro2017simple, webb2023systematic, mondal2023learning, driess2023palm} and some of these works try to address the Visual Question Answering task itself. \citet{Ding2021} propose a new method to address the VQA in videos and run a transformer over slots obtained from a pre-trained MONet \citep{burgess2019monet}, and text tokens of the question, and applies an MLP on top to predict the answer. The method proposed by \citet{wu2022slotformer} reasons over the object representations of Slot Attention to model spatiotemporal relationships, and predicts future object states. Their framework is also applied to a VQA downstream task. 

In addition, a few works focus more specifically on the evaluation of OC representations. \citet{weis2021benchmarking} designs a benchmark over only OC video models and analyzes their performance over different tracking scenarios relevant to natural videos. \citet{yang2024benchmarking} evaluates OC representations and shows their shortcomings in segmenting objects in a real-world dataset.
\citet{Dittadi2021} evaluates the representations indirectly in the context of reconstruction loss, segmentation quality, and object property prediction, and analyzes their generalization and robustness. \citet{Papa2022a} uses the same evaluation metrics on a dataset with complex textures. \citet{yoon2023investigation} evaluates the representations on more practically relevant downstream tasks in reinforcement learning and includes a wider range of methods compared to the previous works.
Finally, \citet{driess2023palm} demonstrates the suitability of OC representations in planning and VQA tasks within a robotic environment. However, the assessment is done on a single OC baseline in the presence of a Large Language Model (LLM) and the VQA setup is restricted to particular scenarios.
In our work, we are interested in investigating the suitability of different types of representation, including object-centric ones, for reasoning tasks. To this end, we opt to more directly assess the suitability of representations for reasoning through VQA.

\section{Experimental Setup}
\label{sec: experimental setup}

In this section, we provide an overview of our experimental setup. First, we introduce the downstream task used in our experiments to evaluate representations. 
We then outline the upstream representation models, the datasets and metrics, and the concrete setup for learning the downstream task.

\subsection{Visual Question Answering}
In this paper, we evaluate the performance attainable on a Visual Question Answering (VQA) task \citep{antol2015vqa} from different representations of the visual scenes. 
With questions that can involve any number of objects from just one to all the objects in an image, VQA presents a more demanding challenge compared to object-level tasks. It requires a thorough understanding of the image and complex reasoning about objects and their relationships. We therefore choose VQA as a benchmark to directly assess the suitability of different representations for reasoning.

Given an image, the task is to answer a natural language question such as \textit{``How many tiny green objects are made of the same material as the purple cube?''}. The questions are usually about how many objects there are, whether an object with a specific attribute exists, and what properties they have in relation to another set of objects in an image. The possible answers include ``yes'', ``no'', and various numerical and categorical values. Further details are provided in \cref{appendix:question_generation}.

\begin{figure*}[t]
    \centering
    \vskip -0.15in
    \includegraphics[width=0.87\linewidth]{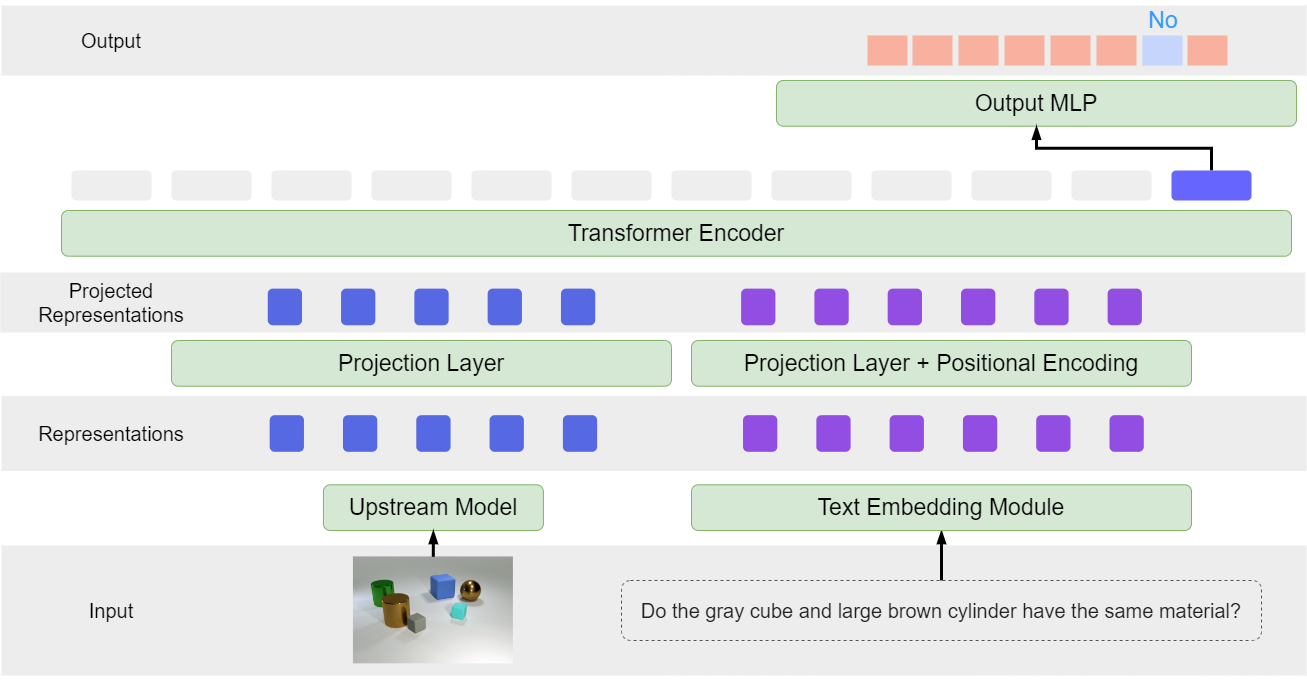}
    \caption{An overview of our framework. Starting from an image and a question, we first extract image and question representations, by applying the upstream model and the text embedding module, respectively. The obtained representations are then passed to the projection layer and then, a positional encoding is applied to the text representations. Next, both are concatenated and a transformer model is applied to the combined sequence. Finally, The answer to the question is obtained by an MLP that takes the transformed value of the \textit{CLS} token and produces a probability vector over all possible answers.}
    \label{fig:framework}
    \vskip -0.12in
\end{figure*}

Our framework, summarized in \cref{fig:framework}, consists of: (1) an upstream model that provides high-level representations of an image, (2) a fixed pre-trained text embedding model that converts a question in natural language to text embeddings, and (3) a downstream model that takes as input the image representation and the text embedding and outputs the answer to the question. We will elaborate on each part in the following sections.

\subsection{Upstream Models}
To investigate OC representations, we consider three types of representations: global, fixed-region, and object-centric. Global representations encode the image into a single vector which contains high-level information about the image. Fixed-region representations consist of a fixed number of vectors, each loosely corresponding to a specific region within the image. OC representations consist of a set of vectors, each ideally corresponding to a single object.

The evaluated models are summarized in \cref{table:models_summary}. As OC baselines, we use \textit{MONet} \citep{burgess2019monet}, \textit{SPACE} \citep{lin2020space}, and \textit{Slot Attention} (\textit{SA}) \citep{locatello2020object}. We also include \textit{ResNet SA} \citep{biza2023invariant}, an improved version of the standard SA autoencoder with the following 
\begin{wraptable}[13]{r}{8.5cm}
    \caption{Summary of models included in our study.}
    \label{table:models_summary}
    \resizebox{8.5cm}{!}{%
        \begin{tabular}{lcr}
            \toprule
            \textbf{Model} & \textbf{Representation Type} & \textbf{Training Regime} \\
            \midrule
            DINOv2 \citep{oquab2023dinov2} & Fixed-Region & Pre-training \\
            MAE \citep{he2022masked} & Fixed-Region & Pre-training \\
            CLIP \citep{radford2021learning} & Fixed-Region & Pre-training \\
            VQ-AE \citep{rombach2022high} & Fixed-Region & Pre-training \\
            KL-AE \citep{rombach2022high} & Fixed-Region & Pre-training \\
            ResNet50 \citep{he2016deep} & Fixed-Region & Pre-training \\
            CNN \citep{zambaldi2018relational} & Fixed-Region & End-to-End Training \\
            MultiCNN \citep{kipf2019contrastive} & Object-Centric & End-to-End Training \\
            Slot Attention \citep{locatello2020object} & Object-Centric & Dataset-Specific Pre-training \\
            ResNet Slot Attention \citep{biza2023invariant} & Object-Centric & Dataset-Specific Pre-training \\
            MONet \citep{burgess2019monet} & Object-Centric & Dataset-Specific Pre-training \\
            SPACE \citep{lin2020space} & Object-Centric & Dataset-Specific Pre-training \\
            STEVE \citep{singh2022simple} & Object-Centric & Dataset-Specific Pre-training \\
            DINOSAURv2 \citep{seitzer2023bridging} & Object-Centric & Dataset-Specific Pre-training \\
            VAE \citep{watters2019spatial} & Global & Dataset-Specific Pre-training \\
            \bottomrule
        \end{tabular}
    }
\end{wraptable}
modifications: the backbone is replaced by a ResNet34 \citep{he2016deep} without pre-training; a larger feature map resolution is used in both the encoder and the decoder; and the slot initializations are learnable. We also consider \textit{STEVE} \citep{singh2022simple}, a state-of-the-art OC video model for complex and naturalistic videos. STEVE is a more robust version of \textit{SLATE} \citep{singh2022illiterate} combining the SLATE decoder with a standard slot-level recurrent model. To adapt STEVE to images, we simply consider images as 1-frame videos, following the authors' recommendation. Furthermore, as the last OC baseline, we consider \textit{DINOSAUR} \citep{seitzer2023bridging}, a state-of-the-art OC image model, and replace its pre-trained DINO~\citep{caron2021emerging} backbone with DINOv2~\citep{oquab2023dinov2}---we refer to this model as \textit{DINOSAURv2}.
Following previous work, we also consider multiple CNNs \citep{kipf2019contrastive, watters2017visual, yoon2023investigation}, each CNN being expected to capture one object in the image, and train all of them end-to-end together with the downstream model---we refer to this approach as \textit{MultiCNN}.

As a classic benchmark for fixed-region representations \citep{yoon2023investigation, seitzer2023bridging}, we include a pre-trained \textit{ResNet50} \citep{he2016deep}. We also utilize two pre-trained autoencoders from Latent Diffusion Models (LDM) \citep{rombach2022high}, one with a KL regularization and the other one with a vector quantization layer, both with a scaling factor of $16$. We refer to them as \textit{KL-AE} and \textit{VQ-AE}, respectively. Additionally, we use pre-trained versions of \textit{DINOv2} \citep{oquab2023dinov2, darcet2023vitneedreg}, \textit{Masked Autoencoder} (\textit{MAE}) \citep{he2022masked}, and \textit{CLIP} \citep{radford2021learning}, all of which have achieved outstanding performance as a backbone in a diverse array of tasks.
Following previous works \citep{santoro2017simple, zambaldi2018relational,yoon2023investigation}, we also implement a simple CNN, and train it end-to-end with the downstream model. 
As a baseline providing a global representation, we follow \citet{Dittadi2021} and consider a variation of vanilla variational autoencoders (VAE) \citep{kingma2014vae,rezende2014} with a broadcast decoder \citep{watters2019spatial}.
Finally, to better understand the models' performances, we include, for each dataset, a baseline trained only on questions, without any information from the corresponding images.\looseness=-1

\paragraph{Training.} Regarding the training of upstream models, we have $3$ different types of models: pre-trained foundation models, pre-trained dataset-specific models, and end-to-end models. Pre-trained foundation models have been trained on large-scale datasets and tasks, serving as the basis for transfer learning in various applications. DINOv2, MAE, CLIP, VQ-AE, KL-AE, and ResNet50 belong to this category that we use off-the-shelf without fine-tuning, in all experiments. Dataset-specific pre-trained models are first trained with an autoencoding objective (only using images, disregarding the questions) on the same dataset that will be used for VQA with their original training procedures and hyperparameter choices. They are subsequently frozen, similarly to foundation models. MONet, SPACE, SA, ResNet SA, STEVE, DINOSAURv2, and VAE are in this category. Finally, end-to-end models are trained from scratch alongside the downstream model to solve the VQA task directly. CNN and MultiCNN belong to this category. For more information about the upstream models, see \cref{appendix:upstream_models}.\looseness=-1

\subsection{Datasets}
\paragraph{Synthetic.}
We utilize three popular multi-object datasets in our experiments: \textit{Multi-dSprites} \citep{dsprites17}, a variation of \textit{CLEVR} \citep{johnson2017clevr} with $6$ objects known as \textit{CLEVR6} \citep{greff2019multi, locatello2020object, Dittadi2021}, and \textit{CLEVRTex} \citep{karazija2021clevrtex} which is a variation of CLEVR featuring synthetic scenes with diverse shapes, textures and photo-mapped materials. This dataset is closer to real-world datasets in terms of visual complexity. To analyze the effect of training data size, we consider $4$ different training data sizes in Multi-dSprites with $40$k, $80$k, $160$k, and $320$k unique images, with the $320$k version as the default version. Each image in the multi-object datasets consists of a background with a fixed color and a set of objects with different properties. Originally, only CLEVR contains questions associated with each image. To make the other datasets applicable to the same VQA task, we augment them with several questions (roughly 40-50) for each image, by adapting the question generation mechanism of \citet{johnson2017clevr} to each dataset. We use this to generate different types of questions, with possible answers including ``yes'', ``no'', natural numbers up to the maximum number of objects, and all possible values of object properties. For more details about the datasets and question generation, see \cref{appendix:datasets}.

\paragraph{Real-World.} Additionally, we extend our results to real-world scenarios with the \textit{VQA-v2} \citep{goyal2017making, antol2015vqa} and \textit{GQA} \citep{hudson2019gqa} datasets. VQA-v2 consists of open-ended questions about images sourced from MS COCO 2014 \citep{lin2014microsoft}, a real-world multi-object dataset. Recently, COCO has been increasingly utilized in object-centric literature \citep{seitzer2023bridging, jiang2023object, wu2023slotdiffusion}, marking a significant advancement in complexity compared to datasets typically used to evaluate object-centric models. GQA is a large-scale dataset designed for visual question answering, focusing on compositional reasoning over real-world scenes. VQA-v2 features a diverse range of possible answers. To align with the same classification pipeline used for synthetic datasets, we limit the questions to yes/no and questions with numeric answers ranging from 0 to 14. This results in a total of 17 possible answers. For more details about the dataset and the preprocessing, see \cref{appendix:datasets}.

\subsection{Metrics}
Following previous works \citep{Ding2021, wu2022slotformer}, we measure performance in our VQA downstream task by average accuracy. 
As metrics for the upstream OC models, we use the Mean Squared Error (MSE) of the reconstructions, and 3 segmentation metrics: the Adjusted Rand Index (ARI) \citep{hubert1985comparing}, Segmentation Covering (SC) \citep{arbelaez2010contour}, and mean Segmentation Covering (mSC) \citep{engelcke2019genesis}. All of these metrics have been extensively used in previous studies \citep{locatello2020object, Dittadi2021, singh2022simple, biza2023invariant}. See \cref{appendix:metrics} for more details about the metrics.

\subsection{Framework Setup}
\label{subsection:framework_setup}

Our VQA framework, depicted in \cref{fig:framework}, closely follows \citet{Ding2021}. 
Given a pair $(x, q)$ where $x \in \R^{3 \times H \times W}$ denotes an image of height $H$ and width $W$, and $q$ denotes a question, the task is to select the correct answer from the set of all possible answers. Since the number of answers in each dataset is relatively small, it is not necessary to generate text tokens as the answers, and similarly to \citet{Ding2021}, we stick to the simpler case of predicting a probability vector over all possible answers in the dataset. Another key aspect to consider is that our primary focus is on evaluating representations while the method for generating questions and the format of the answers hold less significance in this context.

\paragraph{Image and Text Representations.} 
Given a data pair $(x, q)$, the upstream model computes the image representation $z$.
In global representations, $z$ is a vector of size $D_{glob}$. In OC models, $z$ is a $N_{slots} \times D_{oc}$ matrix where $N_{slots}$ is the number of slots in the OC model. In fixed-region representations, $z$ is a feature map of size $P_H \times P_W \times D_{fr}$ where the first two dimensions correspond to the feature map sizes and the third dimension is the size of the representation. For more details about obtaining image representations from the upstream models, see \cref{appendix:upstream_models}.

To embed the question $q$ from text format to word embeddings, we use the Text-to-Text Transfer Transformer \citep[T5;][]{raffel2020exploring} which outputs a matrix $t$ of size $N_{tokens} \times D_{emb}$ representing the embeddings of the tokens in the question where the dimensions correspond to the number of tokens and the embedding size, respectively. See \cref{appendix:text_embedding_module} for more details.

\paragraph{Unifying Image Representations.} In order to use different types of image representations in the downstream model which follows a transformer architecture and will be explained later on, it is necessary to unify the format of representations and convert them to a sequence. We use $z$ as it is for OC representations since each slot corresponds to an object, and can be separately used as an item in the sequence. We reshape fixed-region representations by flattening the spatial dimensions, obtaining a matrix of size $P_H P_W \times D_{fr}$. 

For global representations, we split the single vector $z$ into $K$ vectors of size $D_{glob} / K$. Here, $K$ roughly corresponds to the number of slots in an OC model. In other words, we treat $z$ as a sequence of length $K$ with a latent size of $D_{glob} / K$. 
While we observed this to be the most effective option in terms of downstream performance, we also considered three alternative approaches. 
The first applies a 2-layer MLP to $z$ and subsequently splits the output similarly to what described above; the second method treats the single vector $z$ as one token in a sequence of length 1; the third splits $z$ into $D_{glob}$ sequences of size $1$. 
All these approaches showed poorer downstream performance, and in addition, the last one is computationally expensive due to a large sequence length.

\paragraph{Downstream Model.}
Following previous works on VQA \citep{Ding2021, devlin2018bert, lu2019vilbert},
we use a transformer-based architecture \citep{vaswani2017attention}. Having $t$ and the reformatted $z$ as text and image representations, we apply a separate linear layer on each to make the latent size and the embedding size equal, and we get $t'$ and $z'$, respectively. Then, to inform the downstream model about the order of words, we apply a sinusoidal positional encoding layer to $t'$. Additionally, following \citet{Ding2021}, we augment each vector in $z'$ and $t'$ with a $2$-dimensional one-hot vector indicating whether the input is from the image representation or the text, and the latent size for both will become $D_{model}$. We introduce a trainable vector $\textit{CLS} \in \R^{D_{model}}$, akin to the \textit{CLS} token in \textit{BERT} \citep{devlin2018bert}, to generate classification results.
In the final step, we concatenate $z'$, $t'$, and the \textit{CLS} token and pass this sequence through a transformer with $N_t$ layers. An MLP classifier then takes the transformed \textit{CLS} token and outputs a probability vector over all possible answers.\looseness=-1

\begin{figure*}[t]
    \centering
    \vskip -0.15in
    \includegraphics[width=0.98\textwidth]{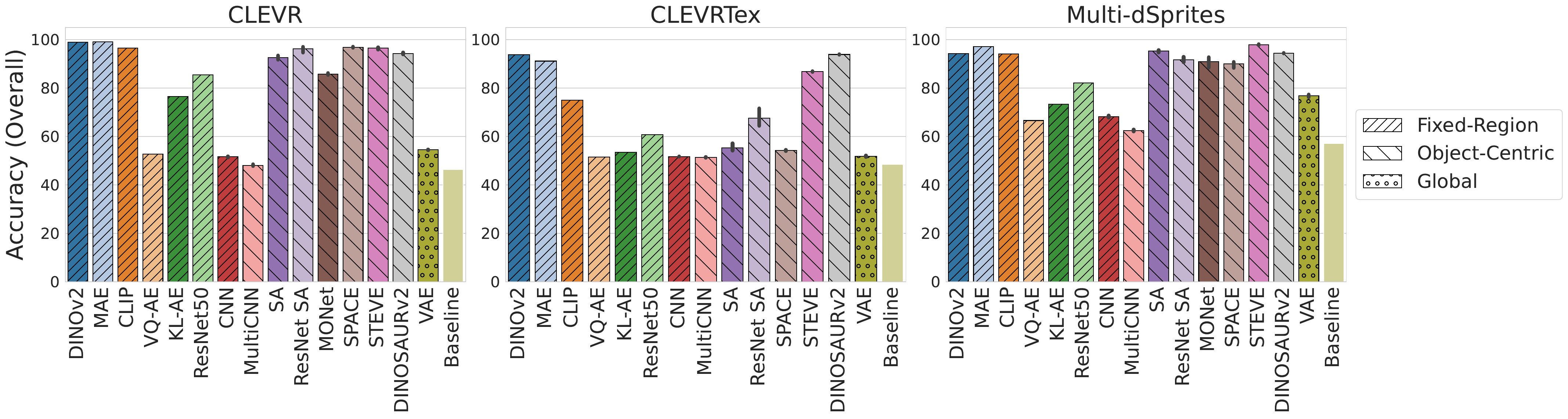}
    \caption{Average accuracies on the VQA downstream task for different upstream representation models on synthetic datasets, when using T-15 as the downstream model. The bars indicate means and 95\% confidence intervals with 3 random seeds, when available.}
    \label{fig:overall_bar_transformer15}
    \vskip -0.12in
\end{figure*}

\subsection{Limitations}
While our goal is to execute a robust and informative experimental study to address the research questions identified in \cref{sec:introduction}, it's important to acknowledge inherent limitations related to datasets, models, and evaluations. 
The foundation models in our study are trained with different objectives and on datasets that differ in size and characteristics, making direct comparisons with OC models more difficult. 
However, it is important to emphasize that this is first and foremost a pragmatic study aimed at deriving practical, actionable insights into representation learning for downstream reasoning tasks. 
To achieve this, we empirically investigate a diverse range of approaches directly available in the literature, without significant modifications, and evaluate their effectiveness for these tasks.

\section{Experimental Results}

Our key findings are presented in this section. In our main set of experiments, we assess how different model representations perform on the Visual Question Answering (VQA) downstream task defined in \cref{sec: experimental setup}. We primarily focus on results from synthetic datasets where we have a unified question-generation procedure and access to underlying ground-truth factors. Our downstream models are transformer encoders with $2$, $5$, and $15$ layers, which we refer to as T-$n$ with $n$ the number of layers. We train all combinations of upstream representation models and downstream classifiers, which amounts to 684 downstream models, with the cross-entropy loss.\footnote{Reproducing our experimental study requires approximately 13 GPU years on Nvidia A100 GPUs.}
We provide all implementation details in \cref{appendix: models and implementation details} and additional experimental results in \cref{appendix:results}.

We carried out an extensive set of experiments with numerous baselines. To improve clarity, we have organized the results into key points and summarized the main takeaways. 
In the following, we report average results and confidence intervals over 3 random seeds, except for foundation models, where only 1 seed is available. We omit MONet's results on CLEVRTex due to its suboptimal performance, consistent with similar experimental results by \citet{Papa2022a}. When extending to real-world datasets, we keep only the pre-trained foundation models and top-performing OC models, excluding other upstream representation models due to their poor performance. Additionally, we report the results on VQA-v2 only with T-2 as the downstream model due to a degradation in performance observed when increasing the number of transformer layers (see \cref{appendix:downstream_model} for more details).
Finally, unless explicitly mentioned, the Multi-dSprites version featured in the plots is the one comprising $320$k unique images.\looseness=-1

\subsection{Main Findings}

\paragraph{Performance of Large Foundation Models.}

\begin{figure*}[t]
    \centering
    \vskip -0.15in
    \includegraphics[width=0.98\textwidth]{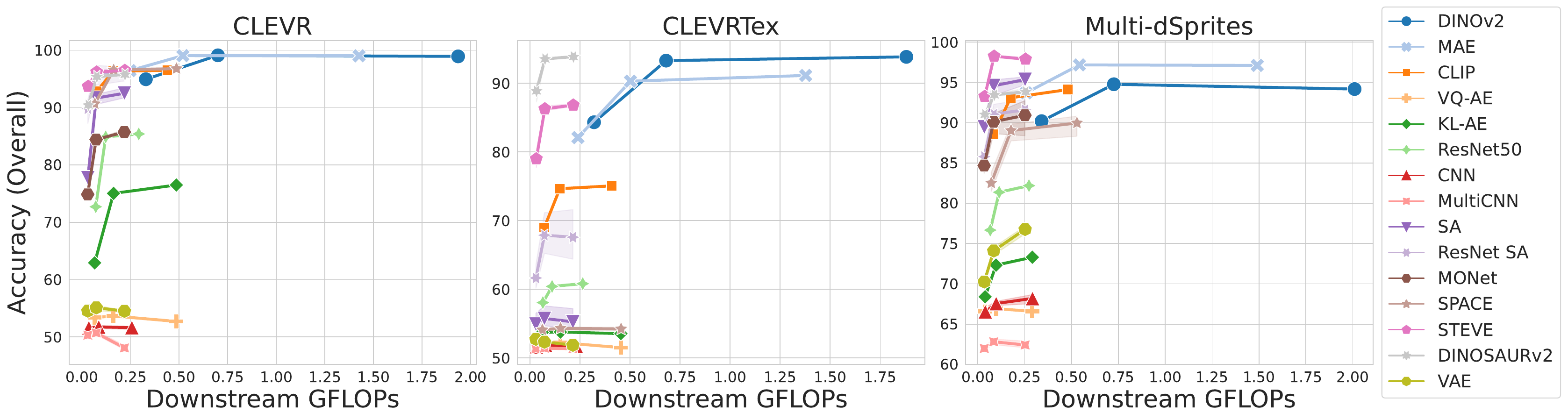}
    \caption{Average accuracies of different models vs. downstream GFLOPs across different datasets. Points along the x-axis represent T-2, T-5, and T-15, respectively. For pre-trained models, only 1 seed is available. For other models, the results are averaged over 3 random seeds and the shaded areas indicate 95\% confidence intervals.}
    \label{fig:downstream_flops}
    \vskip -0.12in
\end{figure*}

\cref{fig:overall_bar_transformer15} shows the overall accuracy for different upstream models across different synthetic datasets with T-15 as the downstream model, which generally achieves the best performance across synthetic datasets and upstream models. 
We observe that large foundation models, i.e., DINOv2, CLIP, and MAE, without any fine-tuning perform comparably well or the best on all datasets, although not by a large margin. 
However, when considering compute requirements, the picture appears more nuanced. In \cref{fig:downstream_flops}, which shows overall accuracy against the GFLOPs used for downstream training, we observe that some OC models achieve comparable performance to large foundation models with significantly less compute, making them more appealing under a limited compute budget.

It's important to emphasize the differences in model sizes and training data between foundation models and OC models. As shown in \cref{tab:pretrained_training_details} in \cref{appendix:upstream_models}, best-performing OC models like STEVE and ResNet SA are much smaller than their counterparts in the foundation model group and are specifically trained on the datasets studied in this work, which are significantly smaller than those used for training foundation models. Additionally, foundation models require substantial computational resources and significant engineering for pre-training, which are beyond our control. Therefore, carefully analyzing the effects of these factors in studies like ours is challenging.

\paragraph{Effect of Object-Centric Bias.} DINOSAURv2 \citep{seitzer2023bridging}, which consists of a pre-trained DINOv2 with Slot Attention applied downstream, allows us to explore the effect of applying the OC bias on a foundation model. Comparing the results of DINOv2 and DINOSAURv2 in \cref{fig:overall_bar_transformer15,fig:downstream_flops} on CLEVRTex and Multi-dSprites, we observe that DINOSAURv2 outperforms DINOv2 while requiring significantly less downstream compute. However, on CLEVR, we do not observe the same patterns, as DINOSAURv2 performs slightly worse than DINOv2. After experimenting with different hyperparameter choices on CLEVR, we found that DINOSAURv2 is highly sensitive to these choices, which likely explains this suboptimal performance.

Additionally, by looking at \cref{fig:downstream_model_size} (\cref{app: effect of downstream model size}) which shows the overall accuracies on different downstream model sizes, we observe that on T-2, DINOv2 exhibits inferior performance compared to DINOSAURv2 on CLEVRTex, Multi-dSprites, and GQA. However, as we scale up the downstream model, starting from T-5, DINOv2 almost matches DINOSAURv2. This indicates that DINOv2 representations do contain the relevant information for the downstream task, but they seem to be less explicit and less readily usable, necessitating a larger downstream model compared to DINOSAURv2 to extract the required information effectively \citep{eastwood2023dci}.\looseness=-1

\paragraph{Performance of Other Upstream Models.}
In \cref{fig:overall_bar_transformer15}, a discernible pattern emerges among upstream models. Generally, OC models consistently outperform other models except large foundation models. Smaller pre-trained models (VQ-AE, KL-AE, and ResNet50) tend to perform worse. Notably, on CLEVRTex, this trend is less pronounced, as most OC and pre-trained models struggle due to the dataset's complexity. Consistent with prior studies \citep{yoon2023investigation}, End-to-end CNN and MultiCNN models consistently score the lowest, and are followed by the global representation of VAEs. Additionally, on CLEVR and CLEVRTex, several models show only a slight improvement over the baseline, which relies solely on the question without any image-related information.

Within foundation models, DINOv2 and MAE consistently outperform others, with CLIP ranking as the third-best model probably due to its relatively smaller size. 
Looking at \cref{tab:pretrained_training_details}, we observe that while the good performance of DINOv2, MAE, and CLIP can likely be attributed to the size of their backbone, there appears to be no clear trend explaining the performance gap among smaller models.
% \vspace{10pt}

\paragraph{Real-world Data.}

\begin{figure*}[t]
    \centering
    \vskip -0.15in
    \includegraphics[width=0.96\textwidth]{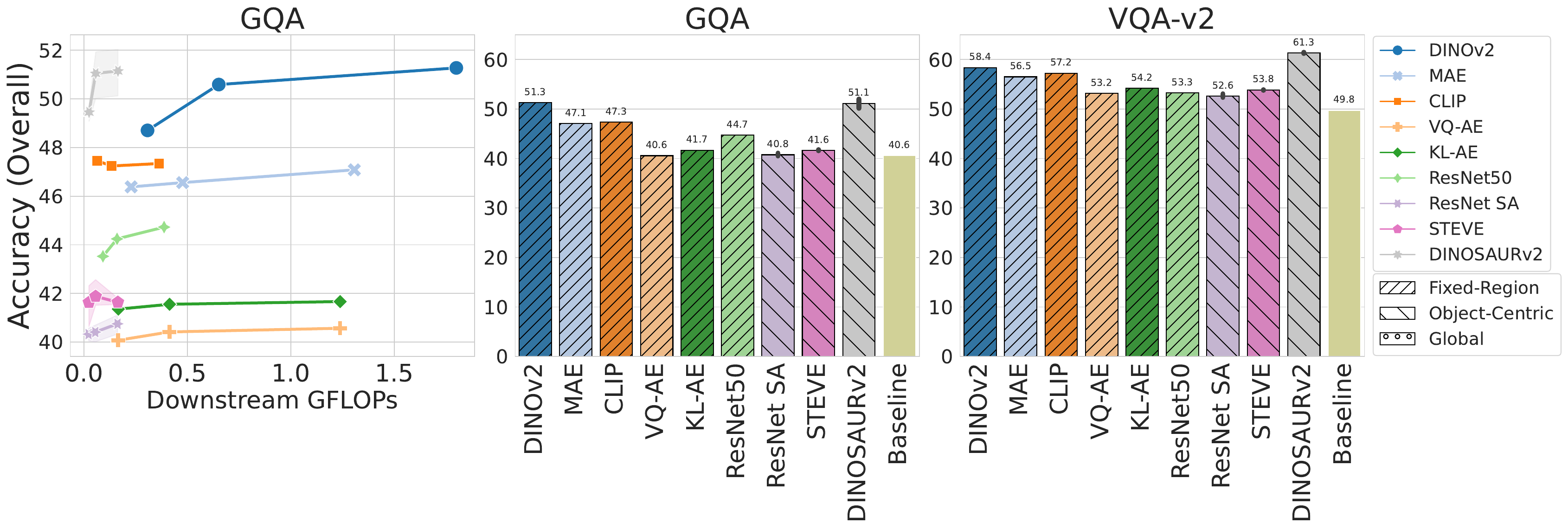}
    \caption{Left: Average accuracies of different models w.r.t. downstream GFLOPs on GQA. Points along the x-axis represent T-2, T-5, and T-15, respectively. For pre-trained models, only one seed is available.  For other models, the results are averaged over 3 random seeds and the shaded areas indicate 95\% confidence intervals.
    Middle \& Right: Average accuracies on GQA and VQA-v2 for different upstream models, with T-15 and T-2 as the downstream models, respectively. The bars indicate means and 95\% confidence intervals with 3 random seeds, when available.}
    \label{fig:custom_real_world}
    \vskip -0.12in
\end{figure*}

To investigate whether our findings hold in real-world scenarios, we conduct the same experiments on the GQA \citep{hudson2019gqa} and VQA-v2 datasets \citep{goyal2017making, antol2015vqa}, two well-established benchmarks for the Visual Question Answering task. \cref{fig:custom_real_world} left shows the overall accuracy of different upstream models on GQA vs. downstream GFLOPs, and the overall accuracy on GQA and VQA-v2 with T-15 and T-2 as the downstream models, respectively. Comparing DINOSAURv2 with DINOv2, we observe the same pattern as in synthetic datasets.
Additionally, the performance trends across different models are consistent with those observed in synthetic datasets, further validating our primary conclusions and suggesting that the findings are robust across both real-world and synthetic datasets.

\begin{takeaway}
    \textbf{Takeaway.}
    While large foundation models can perform comparably to the best-performing OC models without any fine-tuning or hyperparameter adjustments, they generally require larger downstream models and more compute, presumably because their representations are less explicit than OC representations.
    On the other hand, the performance of many OC models drops on more complex datasets.
    Learning OC representations on top of a foundation model (see, e.g., DINOSAURv2) can be a viable solution to get the best of both worlds.
\end{takeaway}

\subsection{Additional Insights}

\begin{figure}
    \centering
    \vskip -0.15in
    \begin{minipage}[b]{0.46\textwidth}
        \centering
        \vspace{0pt}
        \includegraphics[width=\textwidth]{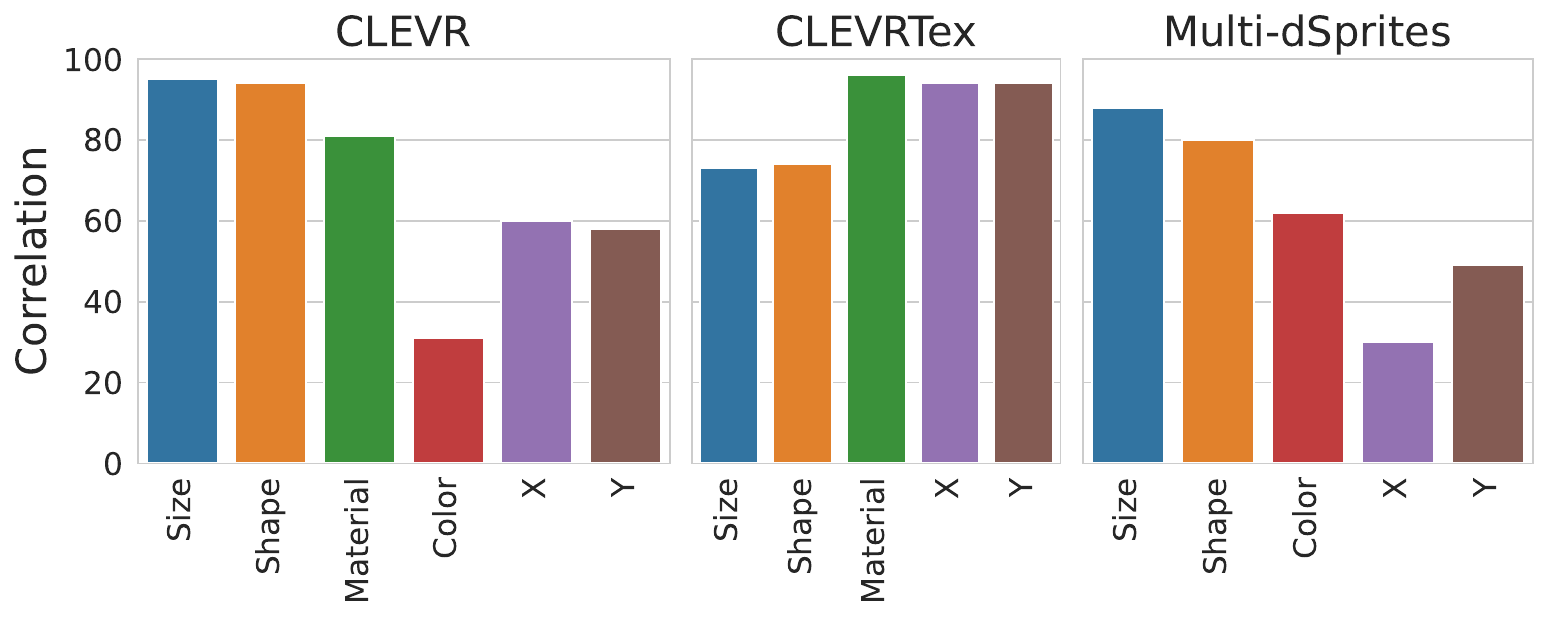}
        \caption{Correlation between property prediction accuracy (reported separately by object property) and overall VQA accuracy.}
        \label{fig:correlation_downstream_bar}
    \end{minipage}
    \hspace{0.02\textwidth}
    \begin{minipage}[b]{0.46\textwidth}
        \centering
        \vspace{0pt}
        \includegraphics[width=\textwidth]{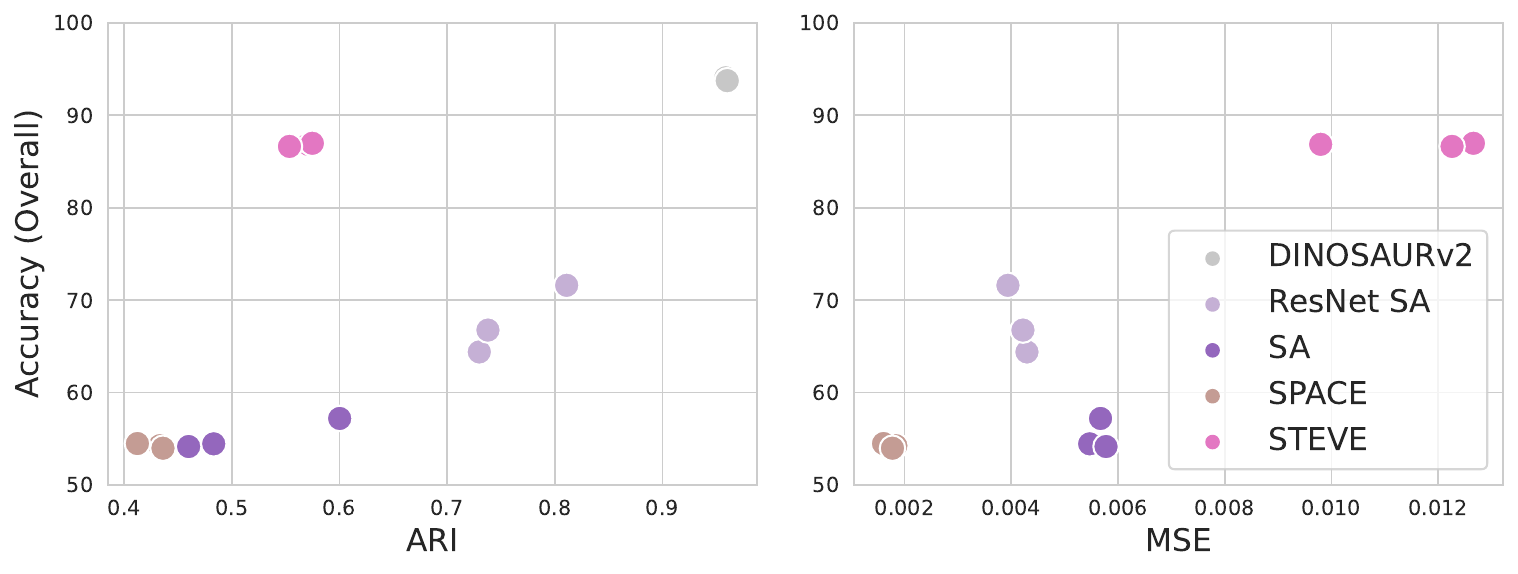}
        \caption{Overall VQA accuracy with T-15 as downstream model, plotted against ARI and MSE for different OC models on CLEVRTex.}
        \label{fig:ari_mse_clevrtex}
    \end{minipage}
    \vskip -0.12in
\end{figure}

\paragraph{Property Prediction vs VQA.}
We additionally evaluate the representations on \textit{property prediction}, a much simpler downstream task wherein the objective is to predict object properties from the representations.
We adopt the same setup as \citet{Dittadi2021} (see \cref{appendix:downstream_property_prediction} for further details).
In \cref{fig:correlation_downstream_bar}, we observe a strong correlation between accuracy of this simple task on most properties, and downstream VQA performance. 
This demonstrates that models capable of accurately predicting object properties excel on more challenging tasks like VQA.
Therefore, performance on simple tasks like property prediction can be a useful evaluation metric for model selection. For the complete correlation results, see \cref{appendix:property_prediction_vs_vqa}.

\begin{takeaway}
    \textbf{Takeaway.}
    Overall, performance on object property prediction strongly correlates with VQA performance, indicating that much simpler tasks can be used to guide model selection.
\end{takeaway}

\paragraph{Upstream vs. Downstream Performance.} 
\cref{fig:ari_mse_clevrtex} depicts the relationship between upstream performance metrics and downstream VQA accuracy of OC models when using T-15 as the downstream model on CLEVRTex.
Notably, STEVE exhibits the worst reconstruction MSE among OC models but achieves the second-best accuracy on VQA. This is not necessarily surprising: while \citet{Dittadi2021} observed a negative correlation between MSE and downstream performance, \citet{Papa2022a} later showed this to no longer hold in the presence of textured objects.
However, a higher ARI was shown to be predictive of better downstream performance. This appears not to hold in our case, as ResNet Slot Attention attains the second-best ARI but does not perform well in the VQA downstream task, while STEVE has a poor segmentation performance while achieving high accuracy. 
Further investigations are needed to shed more light on these trends, allowing for more robust upstream model selection strategies. 
For results on more upstream metrics and other datasets, see \cref{appendix:upstream_vs_downstream}.

\begin{takeaway}
    \textbf{Takeaway.}
    Upstream metrics such as ARI (segmentation) and MSE (reconstruction) are \textit{\textbf{not}} good predictors of downstream performance on our VQA task.
\end{takeaway}

\begin{figure}
    \centering
    \vskip -0.15in
    \begin{minipage}[b]{0.44\textwidth}
        \centering
        \includegraphics[width=\linewidth]{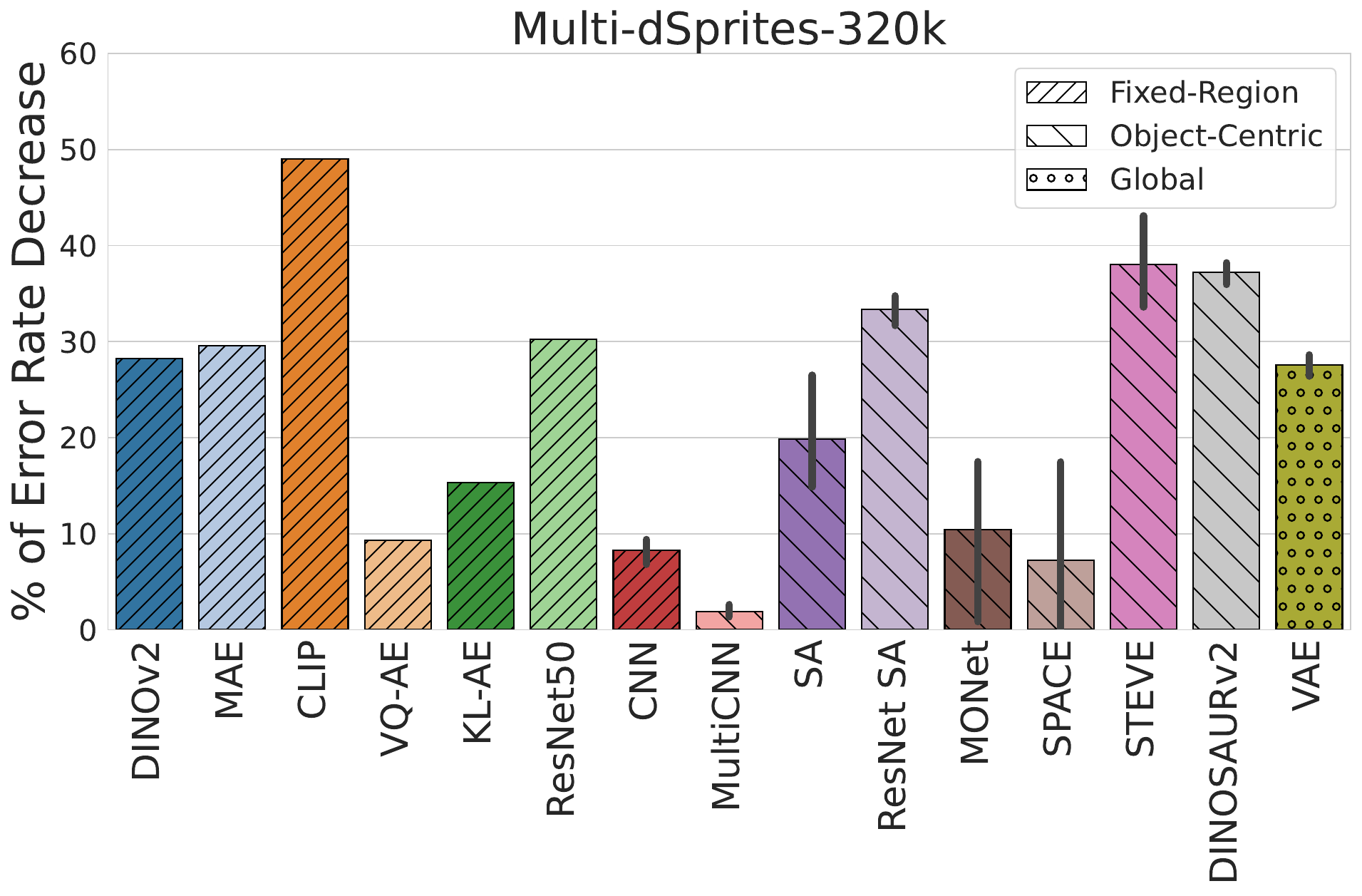}
        \caption{Average \% decrease in VQA error rate (means and 95\% CIs with 3 random seeds) for different upstream models on Multi-dSprites, when increasing the training set size from $40$k to $320$k, using T-15 as the downstream model.}
        \label{fig:single_delta_320k_40k_multi}
    \end{minipage}
    \hspace{0.06\textwidth}
    \begin{minipage}[b]{0.44\textwidth}
        \centering
        \vspace{0pt}
        \includegraphics[width=\linewidth]{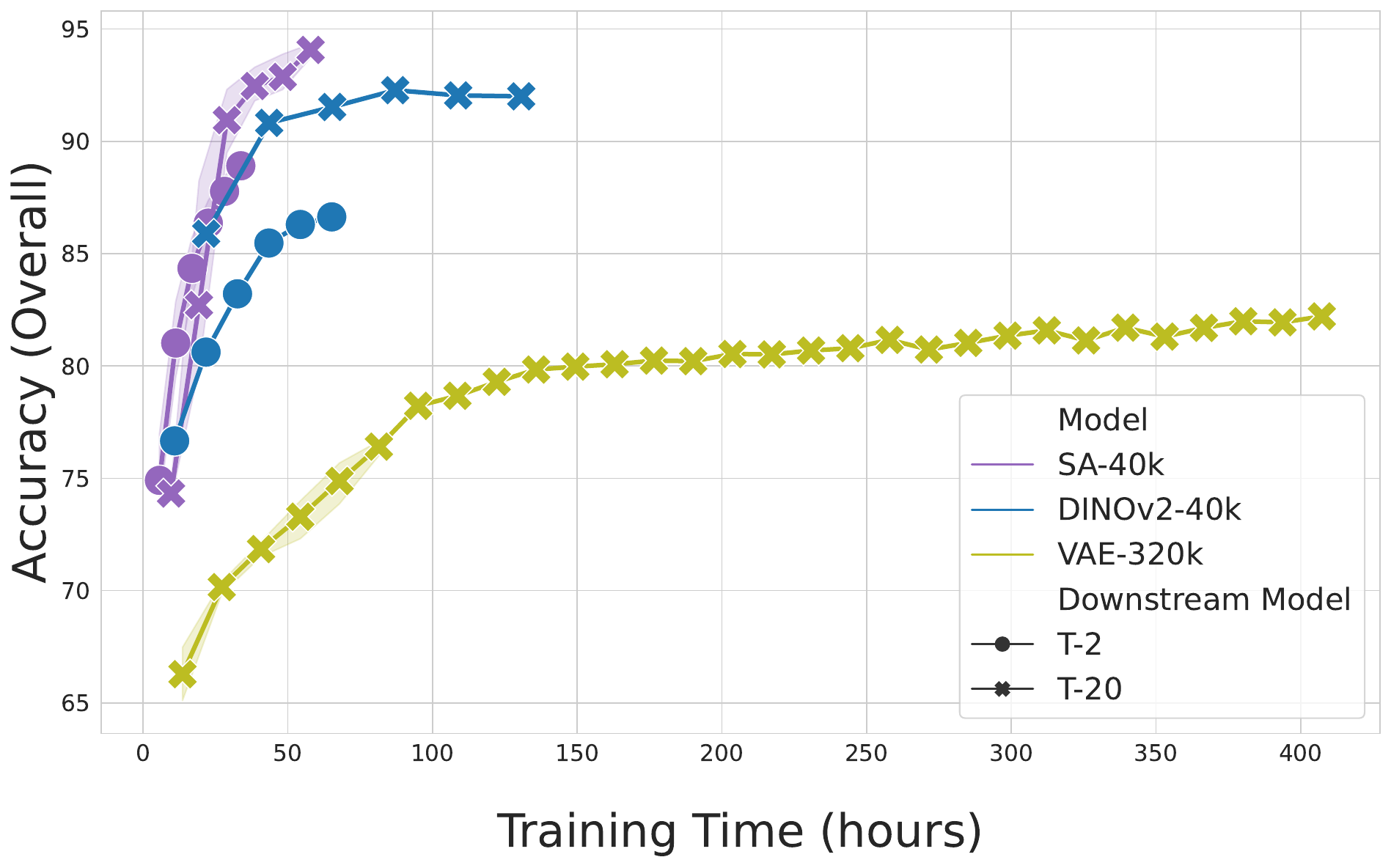}
        \caption{VQA accuracy vs. downstream training time on Multi-dSprites. Slot Attention and DINOv2 are trained on 40k images, while the VAE is trained on 320k images for 3 million steps (5x more than the other models).}
        \label{fig:vae_compute_training_time}
    \end{minipage}
    \vskip -0.12in
\end{figure}

\paragraph{Effect of Training Size.}
\cref{fig:single_delta_320k_40k_multi} shows the percentage decrease in the overall error rate of different models on Multi-dSprites when the training size increases from $40$k to $320$k unique images.
Notably, with approximately $8$x more data, most upstream models exhibit similar improvements, typically around 20--40\%, regardless of their initial performance. CLIP is a notable exception, showing an increase in overall error rate of up to 50\%.
Additionally, end-to-end models (CNN and MultiCNN) and VQ-AE show only minimal improvement compared to the other models. 
See \cref{appendix:training_size} for further results, including raw accuracies and additional dataset sizes between $40$k and $320$k.

\begin{takeaway}
    \textbf{Takeaway.}
    Except for end-to-end models which show minimal improvements, all other models generally exhibit a similar performance gain with larger downstream training sizes.
\end{takeaway}

\paragraph{Consistency of the Results Across Question Types.}
The average Spearman rank correlation between VQA accuracy on different question categories is 0.96 for CLEVR, 0.98 for CLEVRTex, 0.88 for Multi-dSprites, and 0.92 for VQA-v2. This suggests that the average VQA accuracy results shown in \cref{fig:overall_bar_transformer15,fig:custom_real_world} are consistent across question categories. In \cref{appendix:consistency different question types}, \cref{fig:correlation_question_type} shows that for these datasets, the rank correlations are consistently high for all pairs of question categories, and \cref{fig:Exist_bar_transformer15,fig:Count_bar_transformer15,fig:Compare Less_bar_transformer15,fig:Compare Greater_bar_transformer15,fig:Compare Equal_bar_transformer15,fig:Compare Shape_bar_transformer15,fig:Compare Size_bar_transformer15,fig:Query Shape_bar_transformer15,fig:Query Size_bar_transformer15,fig:cocovqav2_question_types_bar_transformer2,fig:gqa_question_types_bar_transformer15} illustrate the complete results separately by question category. In contrast, on GQA, the correlation between accuracies of different question types is notably weaker. In particular, \emph{Compare} and \emph{Logical} questions exhibit low correlation with other categories. This is likely due to their reduced dependency on visual representations and heavier reliance on linguistic cues \citep{liu2022delving}.

Delving deeper into the results for each category, it becomes apparent that on VQA-v2, \emph{Number} questions, which require recognizing quantities in the image, are harder for all the models compared to \emph{Yes/No} questions. On GQA, \emph{Query} questions which demand complex, open-ended reasoning, consistently prove to be the most difficult, whereas \emph{Verify} questions are the easiest. Moreover, models perform similarly on \emph{Logical} and \emph{Compare} questions, indicating that these question types depend much less on the visual information provided by the models.
On synthetic datasets, we observe that \emph{Count} questions, which necessitate an understanding of the existence of multiple objects with specific properties, are generally the most challenging for almost all models. In contrast, \emph{Exist} questions are the easiest, which is expected because they check for the existence of a single object with specific properties. Among \emph{Compare Integer} questions, \emph{Equal} questions appear to be the most challenging, requiring an exact count of two sets of objects. Finally, in \emph{Attribute} questions, \emph{Size} questions emerge as the easiest, while there is no specific discernible pattern among other object attributes.

\begin{takeaway}
    \textbf{Takeaway.}
    While some question categories are on average more difficult than others, we observe a strong correlation between accuracies across different question categories.
\end{takeaway}

\paragraph{Evaluation of Global Representations.}
Here we further investigate whether the global representations of a VAE can match the performance of OC representations when given a significant advantage in terms of data and training budget.
For this purpose, we train a downstream transformer encoder model with 20 layers (T-20) on top of the VAE on Multi-dSprites with $320$k unique images for $3$ million steps and compare the result with other models trained on Multi-dSprites with the smallest training size ($40$k), with the smallest downstream model (T-2) trained for the default number of training steps ($600$k).
From \cref{fig:vae_compute_training_time}, it is evident that the performance of T-20 trained on top of the VAE cannot match the performance of T-2 trained on top of Slot Attention and DINOv2. In conclusion, even with a larger downstream model, more training steps, and a larger training dataset size, global representations of VAEs cannot match the performance of OC models and therefore do not seem ideal for downstream tasks related to objects.

\begin{takeaway}
    \textbf{Takeaway.}
    Even with significantly more training data and compute, global representations such as those of standard VAEs are far from competitive. 
    This corroborates the common assumption that such representations are not suitable for object-related downstream tasks.
\end{takeaway}

\section{Conclusion and Discussion}
In this study, we systematically assess OC representations on downstream reasoning tasks by comparing them with foundation models and various benchmarks across three synthetic and two real-world multi-object datasets. Our primary focus is the VQA task, which requires a precise compositional understanding of images, objects, and their relationships.
Our findings indicate that foundation models perform comparably to OC models without requiring fine-tuning or hyperparameter adjustments. However, they are significantly larger and demand greater computational resources. Overall, this points to a complex trade-off between model classes. Still, one can benefit from both worlds by applying the OC bias to foundation models.
Beyond performance comparisons, we emphasize the importance of downstream evaluations, as they provide a more pragmatic assessment of representation quality and align well with the goals of OC learning---particularly in capturing the compositional properties of scenes. We encourage future research to adopt such evaluations to better understand the strengths and limitations of different approaches.

While our study covers commonly used OC datasets, they are limited to static images. Future work could extend these analyses to video data, where dynamic scene understanding presents additional challenges. Additionally, further research could explore the effects of fine-tuning foundation models, both with and without OC inductive biases, as well as systematically study the generalization capabilities of OC models in other tasks, such as causal inference.

\subsection*{Acknowledgments}
We would like to thank Thomas Kipf, Max Horn, and Sindy Löwe for the helpful discussions and comments. This work was partially supported by the Wallenberg AI, Autonomous Systems and Software Program (WASP), funded by the Knut and Alice Wallenberg Foundation, and by the Helmholtz Foundation Model Initiative, supported by the Helmholtz Association. A.D. acknowledges support from G-Research.
The computations were enabled by the Berzelius resource, provided by the Knut and Alice Wallenberg Foundation at the National Supercomputer Centre, and by the Gauss Centre for Supercomputing e.V. (www.gauss-centre.eu), which provided the required computing time through the John von Neumann Institute for Computing (NIC) on the GCS Supercomputer JUPITER | JUWELS \citep{JUWELS} at Jülich Supercomputing Centre (JSC).

\subsection*{Ethics Statement}
In this work, we are taking a step in the direction of systematically analyzing and understanding the reasoning capabilities of deep learning systems with a particular focus on object-centric models, which benefits both the research community and society. We do not see a negative societal impact of this work beyond what is brought about by general advances in machine learning.

\subsection*{Reproducibility Statement}
All reproduction details regarding the models, hyperparameters, implementation, and training and evaluation procedures are provided in the main text and appendix. Specifically, information about the upstream models, downstream VQA pipeline, and downstream property prediction pipeline is available in \cref{appendix: models and implementation details}. We use publicly available implementations of the models, which are properly referenced in the paper. All datasets are publicly available, and details about each dataset, as well as the question generation procedure, can be found in \cref{appendix:datasets}. Finally, the metrics are described and properly referenced in \cref{appendix:metrics}.

\bibliography{iclr2025_conference}

\begin{thebibliography}{101}
\providecommand{\natexlab}[1]{#1}
\providecommand{\url}[1]{\texttt{#1}}
\expandafter\ifx\csname urlstyle\endcsname\relax
  \providecommand{\doi}[1]{doi: #1}\else
  \providecommand{\doi}{doi: \begingroup \urlstyle{rm}\Url}\fi

\bibitem[Antol et~al.(2015)Antol, Agrawal, Lu, Mitchell, Batra, Zitnick, and Parikh]{antol2015vqa}
Stanislaw Antol, Aishwarya Agrawal, Jiasen Lu, Margaret Mitchell, Dhruv Batra, C.~Lawrence Zitnick, and Devi Parikh.
\newblock Vqa: Visual question answering.
\newblock In \emph{IEEE International Conference on Computer Vision}, pp.\  2425--2433, 2015.

\bibitem[Arbelaez et~al.(2010)Arbelaez, Maire, Fowlkes, and Malik]{arbelaez2010contour}
Pablo Arbelaez, Michael Maire, Charless Fowlkes, and Jitendra Malik.
\newblock Contour detection and hierarchical image segmentation.
\newblock \emph{IEEE Transactions on Pattern Analysis and Machine Intelligence}, 33\penalty0 (5):\penalty0 898--916, 2010.

\bibitem[Biza et~al.(2023)Biza, van Steenkiste, Sajjadi, Elsayed, Mahendran, and Kipf]{biza2023invariant}
Ondrej Biza, Sjoerd van Steenkiste, Mehdi S.~M. Sajjadi, Gamaleldin~Fathy Elsayed, Aravindh Mahendran, and Thomas Kipf.
\newblock Invariant slot attention: Object discovery with slot-centric reference frames.
\newblock In \emph{International Conference on Machine Learning}, 2023.

\bibitem[Borsos et~al.(2023)Borsos, Marinier, Vincent, Kharitonov, Pietquin, Sharifi, Roblek, Teboul, Grangier, Tagliasacchi, et~al.]{borsos2023audiolm}
Zal{\'a}n Borsos, Rapha{\"e}l Marinier, Damien Vincent, Eugene Kharitonov, Olivier Pietquin, Matt Sharifi, Dominik Roblek, Olivier Teboul, David Grangier, Marco Tagliasacchi, et~al.
\newblock Audiolm: A language modeling approach to audio generation.
\newblock \emph{IEEE/ACM Transactions on Audio, Speech, and Language Processing}, 2023.

\bibitem[Brady et~al.(2023)Brady, Zimmermann, Sharma, Schölkopf, von Kügelgen, and Brendel]{brady2023provably}
Jack Brady, Roland~S Zimmermann, Yash Sharma, Bernhard Schölkopf, Julius von Kügelgen, and Wieland Brendel.
\newblock Provably learning object-centric representations.
\newblock \emph{arXiv preprint arXiv:2305.14229}, 2023.

\bibitem[Brown et~al.(2020)Brown, Mann, Ryder, Subbiah, Kaplan, Dhariwal, Neelakantan, Shyam, Sastry, Askell, et~al.]{brown2020language}
Tom Brown, Benjamin Mann, Nick Ryder, Melanie Subbiah, Jared~D Kaplan, Prafulla Dhariwal, Arvind Neelakantan, Pranav Shyam, Girish Sastry, Amanda Askell, et~al.
\newblock Language models are few-shot learners.
\newblock In \emph{Advances in Neural Information Processing Systems}, pp.\  1877--1901, 2020.

\bibitem[Burgess et~al.(2019)Burgess, Matthey, Watters, Kabra, Higgins, Botvinick, and Lerchner]{burgess2019monet}
Christopher~P. Burgess, Loic Matthey, Nicholas Watters, Rishabh Kabra, Irina Higgins, Matt Botvinick, and Alexander Lerchner.
\newblock Monet: Unsupervised scene decomposition and representation.
\newblock \emph{arXiv preprint arXiv:1901.11390}, 2019.

\bibitem[Caron et~al.(2021)Caron, Touvron, Misra, J{\'e}gou, Mairal, Bojanowski, and Joulin]{caron2021emerging}
Mathilde Caron, Hugo Touvron, Ishan Misra, Herv{\'e} J{\'e}gou, Julien Mairal, Piotr Bojanowski, and Armand Joulin.
\newblock Emerging properties in self-supervised vision transformers.
\newblock In \emph{IEEE International Conference on Computer Vision}, pp.\  9650--9660, 2021.

\bibitem[Chen et~al.(2021{\natexlab{a}})Chen, Deng, and Ahn]{chen2021roots}
Chang Chen, Fei Deng, and Sungjin Ahn.
\newblock Roots: Object-centric representation and rendering of 3d scenes.
\newblock \emph{Journal of Machine Learning Research}, 22\penalty0 (1):\penalty0 11770--11805, 2021{\natexlab{a}}.

\bibitem[Chen et~al.(2021{\natexlab{b}})Chen, Mao, Wu, Wong, Tenenbaum, and Gan]{zfchen2021iclr}
Zhenfang Chen, Jiayuan Mao, Jiajun Wu, Kwan-Yee~K Wong, Joshua~B Tenenbaum, and Chuang Gan.
\newblock Grounding physical concepts of objects and events through dynamic visual reasoning.
\newblock In \emph{International Conference on Learning Representations}, 2021{\natexlab{b}}.

\bibitem[Chowdhery et~al.(2023)Chowdhery, Narang, Devlin, Bosma, Mishra, Roberts, Barham, Chung, Sutton, Gehrmann, et~al.]{chowdhery2023palm}
Aakanksha Chowdhery, Sharan Narang, Jacob Devlin, Maarten Bosma, Gaurav Mishra, Adam Roberts, Paul Barham, Hyung~Won Chung, Charles Sutton, Sebastian Gehrmann, et~al.
\newblock Palm: Scaling language modeling with pathways.
\newblock \emph{Journal of Machine Learning Research}, 24\penalty0 (240):\penalty0 1--113, 2023.

\bibitem[Crawford \& Pineau(2019)Crawford and Pineau]{crawford2019spatially}
Eric Crawford and Joelle Pineau.
\newblock Spatially invariant unsupervised object detection with convolutional neural networks.
\newblock In \emph{AAAI Conference on Artificial Intelligence}, pp.\  3412--3420, 2019.

\bibitem[Daniel \& Tamar(2022)Daniel and Tamar]{daniel2022unsupervised}
Tal Daniel and Aviv Tamar.
\newblock Unsupervised image representation learning with deep latent particles.
\newblock \emph{arXiv preprint arXiv:2205.15821}, 2022.

\bibitem[Darcet et~al.(2023)Darcet, Oquab, Mairal, and Bojanowski]{darcet2023vitneedreg}
Timothée Darcet, Maxime Oquab, Julien Mairal, and Piotr Bojanowski.
\newblock Vision transformers need registers.
\newblock \emph{arXiv preprint arXiv:2309.16588}, 2023.

\bibitem[Deng et~al.(2009)Deng, Dong, Socher, Li, Li, and Fei-Fei]{deng2009imagenet}
Jia Deng, Wei Dong, Richard Socher, Li-Jia Li, Kai Li, and Li~Fei-Fei.
\newblock Imagenet: A large-scale hierarchical image database.
\newblock In \emph{IEEE Conference on Computer Vision and Pattern Recognition}, pp.\  248--255, 2009.

\bibitem[Devlin et~al.(2018)Devlin, Chang, Lee, and Toutanova]{devlin2018bert}
Jacob Devlin, Ming-Wei Chang, Kenton Lee, and Kristina Toutanova.
\newblock Bert: Pre-training of deep bidirectional transformers for language understanding.
\newblock \emph{arXiv preprint arXiv:1810.04805}, 2018.

\bibitem[Ding et~al.(2021{\natexlab{a}})Ding, Hill, Santoro, Reynolds, and Botvinick]{Ding2021}
David Ding, Felix Hill, Adam Santoro, Malcolm Reynolds, and Matt Botvinick.
\newblock Attention over learned object embeddings enables complex visual reasoning.
\newblock In \emph{Advances in Neural Information Processing Systems}, pp.\  9112--9124, 2021{\natexlab{a}}.

\bibitem[Ding et~al.(2021{\natexlab{b}})Ding, Chen, Du, Luo, Tenenbaum, and Gan]{ding2021dynamic}
Mingyu Ding, Zhenfang Chen, Tao Du, Ping Luo, Josh Tenenbaum, and Chuang Gan.
\newblock Dynamic visual reasoning by learning differentiable physics models from video and language.
\newblock In \emph{Advances in Neural Information Processing Systems}, pp.\  887--899, 2021{\natexlab{b}}.

\bibitem[Dittadi \& Winther(2019)Dittadi and Winther]{dittadi2019lavae}
Andrea Dittadi and Ole Winther.
\newblock Lavae: Disentangling location and appearance.
\newblock \emph{arXiv preprint arXiv:1909.11813}, 2019.

\bibitem[Dittadi et~al.(2022)Dittadi, Papa, De~Vita, Sch{\"o}lkopf, Winther, and Locatello]{Dittadi2021}
Andrea Dittadi, Samuele~S Papa, Michele De~Vita, Bernhard Sch{\"o}lkopf, Ole Winther, and Francesco Locatello.
\newblock Generalization and robustness implications in object-centric learning.
\newblock In \emph{International Conference on Machine Learning}, pp.\  5221--5285, 2022.

\bibitem[Dosovitskiy et~al.(2020)Dosovitskiy, Beyer, Kolesnikov, Weissenborn, Zhai, Unterthiner, Dehghani, Minderer, Heigold, Gelly, et~al.]{dosovitskiy2020image}
Alexey Dosovitskiy, Lucas Beyer, Alexander Kolesnikov, Dirk Weissenborn, Xiaohua Zhai, Thomas Unterthiner, Mostafa Dehghani, Matthias Minderer, Georg Heigold, Sylvain Gelly, et~al.
\newblock An image is worth 16x16 words: Transformers for image recognition at scale.
\newblock \emph{arXiv preprint arXiv:2010.11929}, 2020.

\bibitem[Driess et~al.(2023)Driess, Xia, Sajjadi, Lynch, Chowdhery, Ichter, Wahid, Tompson, Vuong, Yu, et~al.]{driess2023palm}
Danny Driess, Fei Xia, Mehdi~SM Sajjadi, Corey Lynch, Aakanksha Chowdhery, Brian Ichter, Ayzaan Wahid, Jonathan Tompson, Quan Vuong, Tianhe Yu, et~al.
\newblock Palm-e: An embodied multimodal language model.
\newblock \emph{arXiv preprint arXiv:2303.03378}, 2023.

\bibitem[Eastwood et~al.(2023)Eastwood, Nicolicioiu, Von~K{\"u}gelgen, Keki{\'c}, Tr{\"a}uble, Dittadi, and Sch{\"o}lkopf]{eastwood2023dci}
Cian Eastwood, Andrei~Liviu Nicolicioiu, Julius Von~K{\"u}gelgen, Armin Keki{\'c}, Frederik Tr{\"a}uble, Andrea Dittadi, and Bernhard Sch{\"o}lkopf.
\newblock Dci-es: An extended disentanglement framework with connections to identifiability.
\newblock In \emph{International Conference on Learning Representations}, 2023.

\bibitem[Elsayed et~al.(2022)Elsayed, Mahendran, van Steenkiste, Greff, Mozer, and Kipf]{elsayed2022savi++}
Gamaleldin Elsayed, Aravindh Mahendran, Sjoerd van Steenkiste, Klaus Greff, Michael~C Mozer, and Thomas Kipf.
\newblock Savi++: Towards end-to-end object-centric learning from real-world videos.
\newblock In \emph{Advances in Neural Information Processing Systems}, pp.\  28940--28954, 2022.

\bibitem[Engelcke et~al.(2019)Engelcke, Kosiorek, Jones, and Posner]{engelcke2019genesis}
Martin Engelcke, Adam~R Kosiorek, Oiwi~Parker Jones, and Ingmar Posner.
\newblock Genesis: Generative scene inference and sampling with object-centric latent representations.
\newblock \emph{arXiv preprint arXiv:1907.13052}, 2019.

\bibitem[Engelcke et~al.(2021)Engelcke, Parker~Jones, and Posner]{engelcke2021genesis}
Martin Engelcke, Oiwi Parker~Jones, and Ingmar Posner.
\newblock Genesis-v2: Inferring unordered object representations without iterative refinement.
\newblock In \emph{Advances in Neural Information Processing Systems}, pp.\  8085--8094, 2021.

\bibitem[Eslami et~al.(2016)Eslami, Heess, Weber, Tassa, Szepesvari, Hinton, et~al.]{eslami2016attend}
S.~M. Eslami, Nicolas Heess, Theophane Weber, Yuval Tassa, David Szepesvari, Geoffrey~E Hinton, et~al.
\newblock Attend, infer, repeat: Fast scene understanding with generative models.
\newblock In \emph{Advances in Neural Information Processing Systems}, 2016.

\bibitem[Goyal et~al.(2019)Goyal, Lamb, Hoffmann, Sodhani, Levine, Bengio, and Sch{\"o}lkopf]{goyal2019recurrent}
Anirudh Goyal, Alex Lamb, Jordan Hoffmann, Shagun Sodhani, Sergey Levine, Yoshua Bengio, and Bernhard Sch{\"o}lkopf.
\newblock Recurrent independent mechanisms.
\newblock \emph{arXiv preprint arXiv:1909.10893}, 2019.

\bibitem[Goyal et~al.(2017)Goyal, Khot, Summers-Stay, Batra, and Parikh]{goyal2017making}
Yash Goyal, Tejas Khot, Douglas Summers-Stay, Dhruv Batra, and Devi Parikh.
\newblock Making the v in vqa matter: Elevating the role of image understanding in visual question answering.
\newblock In \emph{IEEE Conference on Computer Vision and Pattern Recognition}, pp.\  6904--6913, 2017.

\bibitem[Greff et~al.(2017)Greff, Van~Steenkiste, and Schmidhuber]{greff2017neural}
Klaus Greff, Sjoerd Van~Steenkiste, and Jürgen Schmidhuber.
\newblock Neural expectation maximization.
\newblock In \emph{Advances in Neural Information Processing Systems}, 2017.

\bibitem[Greff et~al.(2019)Greff, Kaufman, Kabra, Watters, Burgess, Zoran, Matthey, Botvinick, and Lerchner]{greff2019multi}
Klaus Greff, Rapha{\"e}l~Lopez Kaufman, Rishabh Kabra, Nick Watters, Christopher Burgess, Daniel Zoran, Loic Matthey, Matthew Botvinick, and Alexander Lerchner.
\newblock Multi-object representation learning with iterative variational inference.
\newblock In \emph{International Conference on Machine Learning}, pp.\  2424--2433, 2019.

\bibitem[Gregor et~al.(2015)Gregor, Danihelka, Graves, Rezende, and Wierstra]{gregor2015draw}
Karol Gregor, Ivo Danihelka, Alex Graves, Danilo Rezende, and Daan Wierstra.
\newblock Draw: A recurrent neural network for image generation.
\newblock In \emph{International Conference on Machine Learning}, pp.\  1462--1471, 2015.

\bibitem[He et~al.(2016)He, Zhang, Ren, and Sun]{he2016deep}
Kaiming He, Xiangyu Zhang, Shaoqing Ren, and Jian Sun.
\newblock Deep residual learning for image recognition.
\newblock In \emph{IEEE Conference on Computer Vision and Pattern Recognition}, pp.\  770--778, 2016.

\bibitem[He et~al.(2022)He, Chen, Xie, Li, Doll{\'a}r, and Girshick]{he2022masked}
Kaiming He, Xinlei Chen, Saining Xie, Yanghao Li, Piotr Doll{\'a}r, and Ross Girshick.
\newblock Masked autoencoders are scalable vision learners.
\newblock In \emph{IEEE Conference on Computer Vision and Pattern Recognition}, pp.\  16000--16009, 2022.

\bibitem[Huang et~al.(2023)Huang, Wang, Deng, Ye, Su, Sun, He, Gu, Gu, Zhang, et~al.]{huang2023stu}
Ziyan Huang, Haoyu Wang, Zhongying Deng, Jin Ye, Yanzhou Su, Hui Sun, Junjun He, Yun Gu, Lixu Gu, Shaoting Zhang, et~al.
\newblock Stu-net: Scalable and transferable medical image segmentation models empowered by large-scale supervised pre-training.
\newblock \emph{arXiv preprint arXiv:2304.06716}, 2023.

\bibitem[Hubert \& Arabie(1985)Hubert and Arabie]{hubert1985comparing}
Lawrence Hubert and Phipps Arabie.
\newblock Comparing partitions.
\newblock \emph{Journal of classification}, 2:\penalty0 193--218, 1985.

\bibitem[Hudson \& Manning(2019)Hudson and Manning]{hudson2019gqa}
Drew~A Hudson and Christopher~D Manning.
\newblock Gqa: A new dataset for real-world visual reasoning and compositional question answering.
\newblock In \emph{IEEE Conference on Computer Vision and Pattern Recognition}, pp.\  6700--6709, 2019.

\bibitem[Jabri et~al.(2023)Jabri, van Steenkiste, Hoogeboom, Sajjadi, and Kipf]{jabri2023dorsal}
Allan Jabri, Sjoerd van Steenkiste, Emiel Hoogeboom, Mehdi S.~M. Sajjadi, and Thomas Kipf.
\newblock Dorsal: Diffusion for object-centric representations of scenes et al.
\newblock \emph{arXiv preprint arXiv:2306.08068}, 2023.

\bibitem[Jia et~al.(2023)Jia, Liu, and Huang]{jia2023improving}
Baoxiong Jia, Yu~Liu, and Siyuan Huang.
\newblock Improving object-centric learning with query optimization.
\newblock In \emph{International Conference on Learning Representations}, 2023.

\bibitem[Jiang et~al.(2019)Jiang, Janghorbani, De~Melo, and Ahn]{JiangJanghorbaniDeMeloAhn2020SCALOR}
Jindong Jiang, Sepehr Janghorbani, Gerard De~Melo, and Sungjin Ahn.
\newblock Scalor: Generative world models with scalable object representations.
\newblock In \emph{International Conference on Learning Representations}, 2019.

\bibitem[Jiang et~al.(2023)Jiang, Deng, Singh, and Ahn]{jiang2023object}
Jindong Jiang, Fei Deng, Gautam Singh, and Sungjin Ahn.
\newblock Object-centric slot diffusion.
\newblock In \emph{Advances in Neural Information Processing Systems}, 2023.

\bibitem[Johnson et~al.(2017)Johnson, Hariharan, Van Der~Maaten, Fei-Fei, Zitnick, and Girshick]{johnson2017clevr}
Justin Johnson, Bharath Hariharan, Laurens Van Der~Maaten, Li~Fei-Fei, C.~Lawrence Zitnick, and Ross Girshick.
\newblock Clevr: A diagnostic dataset for compositional language and elementary visual reasoning.
\newblock In \emph{IEEE Conference on Computer Vision and Pattern Recognition}, pp.\  2901--2910, 2017.

\bibitem[{J\"{u}lich Supercomputing Centre}(2021)]{JUWELS}
{J\"{u}lich Supercomputing Centre}.
\newblock Juwels cluster and booster: Exascale pathfinder with modular supercomputing architecture at juelich supercomputing centre.
\newblock \emph{JLSRF}, 7\penalty0 (A138), 2021.
\newblock \doi{10.17815/jlsrf-7-183}.

\bibitem[Jumper et~al.(2021)Jumper, Evans, Pritzel, Green, Figurnov, Ronneberger, Tunyasuvunakool, Bates, {\v{Z}}{\'\i}dek, Potapenko, et~al.]{jumper2021highly}
John Jumper, Richard Evans, Alexander Pritzel, Tim Green, Michael Figurnov, Olaf Ronneberger, Kathryn Tunyasuvunakool, Russ Bates, Augustin {\v{Z}}{\'\i}dek, Anna Potapenko, et~al.
\newblock Highly accurate protein structure prediction with alphafold.
\newblock \emph{Nature}, 596\penalty0 (7873):\penalty0 583--589, 2021.

\bibitem[Jung et~al.(2024)Jung, Yoo, Ahn, and Hong]{jung2024learning}
Whie Jung, Jaehoon Yoo, Sungjin Ahn, and Seunghoon Hong.
\newblock Learning to compose: Improving object-centric learning by injecting compositionality.
\newblock \emph{arXiv preprint arXiv:2405.00646}, 2024.

\bibitem[Karazija et~al.(2021)Karazija, Laina, and Rupprecht]{karazija2021clevrtex}
Laurynas Karazija, Iro Laina, and Christian Rupprecht.
\newblock Clevrtex: A texture-rich benchmark for unsupervised multi-object segmentation.
\newblock \emph{arXiv preprint arXiv:2111.10265}, 2021.

\bibitem[Kim et~al.(2023{\natexlab{a}})Kim, Choi, Choi, and Kim]{kim2023shepherding}
Jinwoo Kim, Janghyuk Choi, Ho-Jin Choi, and Seon~Joo Kim.
\newblock Shepherding slots to objects: Towards stable and robust object-centric learning.
\newblock In \emph{IEEE Conference on Computer Vision and Pattern Recognition}, pp.\  19198--19207, 2023{\natexlab{a}}.

\bibitem[Kim et~al.(2023{\natexlab{b}})Kim, Singh, Park, Gulcehre, and Ahn]{kim2023imagine}
Yeongbin Kim, Gautam Singh, Junyeong Park, Caglar Gulcehre, and Sungjin Ahn.
\newblock Imagine the unseen world: A benchmark for systematic generalization in visual world models.
\newblock \emph{arXiv preprint arXiv:2311.09064}, 2023{\natexlab{b}}.

\bibitem[Kingma et~al.(2014)Kingma, Welling, et~al.]{kingma2014vae}
Diederik~P Kingma, Max Welling, et~al.
\newblock Auto-encoding variational bayes.
\newblock In \emph{International Conference on Learning Representations}, 2014.

\bibitem[Kipf et~al.(2019)Kipf, Van~der Pol, and Welling]{kipf2019contrastive}
Thomas Kipf, Elise Van~der Pol, and Max Welling.
\newblock Contrastive learning of structured world models.
\newblock \emph{arXiv preprint arXiv:1911.12247}, 2019.

\bibitem[Kipf et~al.(2022)Kipf, Elsayed, Mahendran, Stone, Sabour, Heigold, Jonschkowski, Dosovitskiy, and Greff]{kipf2022conditional}
Thomas Kipf, Gamaleldin~F. Elsayed, Aravindh Mahendran, Austin Stone, Sara Sabour, Georg Heigold, Rico Jonschkowski, Alexey Dosovitskiy, and Klaus Greff.
\newblock Conditional object-centric learning from video.
\newblock In \emph{International Conference on Learning Representations}, 2022.

\bibitem[Kirillov et~al.(2023)Kirillov, Mintun, Ravi, Mao, Rolland, Gustafson, Xiao, Whitehead, Berg, Lo, Doll{\'a}r, and Girshick]{kirillov2023segany}
Alexander Kirillov, Eric Mintun, Nikhila Ravi, Hanzi Mao, Chloe Rolland, Laura Gustafson, Tete Xiao, Spencer Whitehead, Alexander~C. Berg, Wan-Yen Lo, Piotr Doll{\'a}r, and Ross Girshick.
\newblock Segment anything.
\newblock \emph{arXiv preprint arXiv:2304.02643}, 2023.

\bibitem[Kori et~al.(2023)Kori, Locatello, Ribeiro, Toni, and Glocker]{kori2023grounded}
Avinash Kori, Francesco Locatello, Fabio De~Sousa Ribeiro, Francesca Toni, and Ben Glocker.
\newblock Grounded object-centric learning.
\newblock In \emph{International Conference on Learning Representations}, 2023.

\bibitem[Kosiorek et~al.(2018)Kosiorek, Kim, Teh, and Posner]{kosiorek2018sequential}
Adam Kosiorek, Hyunjik Kim, Yee~Whye Teh, and Ingmar Posner.
\newblock Sequential attend, infer, repeat: Generative modelling of moving objects.
\newblock In \emph{Advances in Neural Information Processing Systems}, 2018.

\bibitem[Kuznetsova et~al.(2018)Kuznetsova, Rom, Alldrin, Uijlings, Krasin, Pont-Tuset, Kamali, Popov, Malloci, Duerig, and Ferrari]{OpenImages}
Alina Kuznetsova, Hassan Rom, Neil Alldrin, Jasper Uijlings, Ivan Krasin, Jordi Pont-Tuset, Shahab Kamali, Stefan Popov, Matteo Malloci, Tom Duerig, and Vittorio Ferrari.
\newblock The open images dataset v4: Unified image classification, object detection, and visual relationship detection at scale.
\newblock \emph{arXiv preprint arXiv:1811.00982}, 2018.

\bibitem[Lake et~al.(2017)Lake, Ullman, Tenenbaum, and Gershman]{lake2017building}
Brenden~M Lake, Tomer~D Ullman, Joshua~B Tenenbaum, and Samuel~J Gershman.
\newblock Building machines that learn and think like people.
\newblock \emph{Behavioral and brain sciences}, 40:\penalty0 e253, 2017.

\bibitem[Lin et~al.(2014)Lin, Maire, Belongie, Hays, Perona, Ramanan, Doll{\'a}r, and Zitnick]{lin2014microsoft}
Tsung-Yi Lin, Michael Maire, Serge Belongie, James Hays, Pietro Perona, Deva Ramanan, Piotr Doll{\'a}r, and C.~Lawrence Zitnick.
\newblock Microsoft coco: Common objects in context.
\newblock In \emph{European Conference on Computer Vision}, pp.\  740--755, 2014.

\bibitem[Lin et~al.(2020{\natexlab{a}})Lin, Wu, Peri, Fu, Jiang, and Ahn]{lin2020improving}
Zhixuan Lin, Yi-Fu Wu, Skand Peri, Bofeng Fu, Jindong Jiang, and Sungjin Ahn.
\newblock Improving generative imagination in object-centric world models.
\newblock In \emph{International Conference on Machine Learning}, pp.\  6140--6149, 2020{\natexlab{a}}.

\bibitem[Lin et~al.(2020{\natexlab{b}})Lin, Wu, Peri, Sun, Singh, Deng, Jiang, and Ahn]{lin2020space}
Zhixuan Lin, Yi-Fu Wu, Skand~Vishwanath Peri, Weihao Sun, Gautam Singh, Fei Deng, Jindong Jiang, and Sungjin Ahn.
\newblock Space: Unsupervised object-oriented scene representation via spatial attention and decomposition.
\newblock In \emph{International Conference on Learning Representations}, 2020{\natexlab{b}}.

\bibitem[Liu et~al.(2022)Liu, Pfeiffer, Korhonen, Vuli{\'c}, and Gurevych]{liu2022delving}
Chen Liu, Jonas Pfeiffer, Anna Korhonen, Ivan Vuli{\'c}, and Iryna Gurevych.
\newblock Delving deeper into cross-lingual visual question answering.
\newblock \emph{arXiv preprint arXiv:2202.07630}, 2022.

\bibitem[Liu et~al.(2023)Liu, Alahi, Russell, Horn, Zietlow, Schölkopf, and Locatello]{liu2023causal}
Yuejiang Liu, Alexandre Alahi, Chris Russell, Max Horn, Dominik Zietlow, Bernhard Schölkopf, and Francesco Locatello.
\newblock Causal triplet: An open challenge for intervention-centric causal representation learning.
\newblock \emph{arXiv preprint arXiv:2301.05169}, 2023.

\bibitem[Locatello et~al.(2020)Locatello, Weissenborn, Unterthiner, Mahendran, Heigold, Uszkoreit, Dosovitskiy, and Kipf]{locatello2020object}
Francesco Locatello, Dirk Weissenborn, Thomas Unterthiner, Aravindh Mahendran, Georg Heigold, Jakob Uszkoreit, Alexey Dosovitskiy, and Thomas Kipf.
\newblock Object-centric learning with slot attention.
\newblock In \emph{Advances in Neural Information Processing Systems}, pp.\  11525--11538, 2020.

\bibitem[Lu et~al.(2019)Lu, Batra, Parikh, and Lee]{lu2019vilbert}
Jiasen Lu, Dhruv Batra, Devi Parikh, and Stefan Lee.
\newblock Vilbert: Pretraining task-agnostic visiolinguistic representations for vision-and-language tasks.
\newblock In \emph{Advances in Neural Information Processing Systems}, 2019.

\bibitem[Löwe et~al.(2023)Löwe, Lippe, Locatello, and Welling]{lowe2023rotating}
Sindy Löwe, Phillip Lippe, Francesco Locatello, and Max Welling.
\newblock Rotating features for object discovery.
\newblock In \emph{Advances in Neural Information Processing Systems}, 2023.

\bibitem[Majellaro \& Collu(2024)Majellaro and Collu]{Majellaro2024explicitly}
Riccardo Majellaro and Jonathan Collu.
\newblock Explicitly disentangled representations in object-centric learning.
\newblock \emph{arXiv preprint arXiv:2401.10148}, 2024.

\bibitem[Mansouri et~al.(2023)Mansouri, Hartford, Zhang, and Bengio]{mansouri2023object}
Amin Mansouri, Jason Hartford, Yan Zhang, and Yoshua Bengio.
\newblock Object-centric architectures enable efficient causal representation learning.
\newblock \emph{arXiv preprint arXiv:2310.19054}, 2023.

\bibitem[Matthey et~al.(2017)Matthey, Higgins, Hassabis, and Lerchner]{dsprites17}
Loic Matthey, Irina Higgins, Demis Hassabis, and Alexander Lerchner.
\newblock dsprites: Disentanglement testing sprites dataset, 2017.

\bibitem[Mondal et~al.(2023)Mondal, Webb, and Cohen]{mondal2023learning}
Shanka~Subhra Mondal, Taylor~Whittington Webb, and Jonathan Cohen.
\newblock Learning to reason over visual objects.
\newblock In \emph{International Conference on Learning Representations}, 2023.

\bibitem[Oquab et~al.(2023)Oquab, Darcet, Moutakanni, Vo, Szafraniec, Khalidov, Fernandez, Haziza, Massa, El-Nouby, Howes, Huang, Xu, Sharma, Li, Galuba, Rabbat, Assran, Ballas, Synnaeve, Misra, Jegou, Mairal, Labatut, Joulin, and Bojanowski]{oquab2023dinov2}
Maxime Oquab, Timothée Darcet, Theo Moutakanni, Huy~V. Vo, Marc Szafraniec, Vasil Khalidov, Pierre Fernandez, Daniel Haziza, Francisco Massa, Alaaeldin El-Nouby, Russell Howes, Po-Yao Huang, Hu~Xu, Vasu Sharma, Shang-Wen Li, Wojciech Galuba, Mike Rabbat, Mido Assran, Nicolas Ballas, Gabriel Synnaeve, Ishan Misra, Herve Jegou, Julien Mairal, Patrick Labatut, Armand Joulin, and Piotr Bojanowski.
\newblock Dinov2: Learning robust visual features without supervision.
\newblock \emph{arXiv preprint arXiv:2304.07193}, 2023.

\bibitem[Papa et~al.(2022)Papa, Winther, and Dittadi]{Papa2022a}
Samuele Papa, Ole Winther, and Andrea Dittadi.
\newblock Inductive biases for object-centric representations in the presence of complex textures.
\newblock \emph{arXiv preprint arXiv:2204.08479}, 2022.

\bibitem[Pearl(2009)]{pearl2009causality}
Judea Pearl.
\newblock \emph{Causality}.
\newblock Cambridge University Press, 2009.

\bibitem[Peters et~al.(2017)Peters, Janzing, and Schölkopf]{peters2017elements}
Jonas Peters, Dominik Janzing, and Bernhard Schölkopf.
\newblock \emph{Elements of Causal Inference: Foundations and Learning Algorithms}.
\newblock The MIT Press, 2017.

\bibitem[Radford et~al.(2021)Radford, Kim, Hallacy, Ramesh, Goh, Agarwal, Sastry, Askell, Mishkin, Clark, et~al.]{radford2021learning}
Alec Radford, Jong~Wook Kim, Chris Hallacy, Aditya Ramesh, Gabriel Goh, Sandhini Agarwal, Girish Sastry, Amanda Askell, Pamela Mishkin, Jack Clark, et~al.
\newblock Learning transferable visual models from natural language supervision.
\newblock In \emph{International Conference on Machine Learning}, pp.\  8748--8763, 2021.

\bibitem[Raffel et~al.(2020)Raffel, Shazeer, Roberts, Lee, Narang, Matena, Zhou, Li, and Liu]{raffel2020exploring}
Colin Raffel, Noam Shazeer, Adam Roberts, Katherine Lee, Sharan Narang, Michael Matena, Yanqi Zhou, Wei Li, and Peter~J Liu.
\newblock Exploring the limits of transfer learning with a unified text-to-text transformer.
\newblock \emph{Journal of Machine Learning Research}, 21\penalty0 (1):\penalty0 5485--5551, 2020.

\bibitem[Ren et~al.(2015)Ren, Kiros, and Zemel]{ren2015exploring}
Mengye Ren, Ryan Kiros, and Richard Zemel.
\newblock Exploring models and data for image question answering.
\newblock In \emph{Advances in Neural Information Processing Systems}, 2015.

\bibitem[Rezende et~al.(2014)Rezende, Mohamed, and Wierstra]{rezende2014}
Danilo~Jimenez Rezende, Shakir Mohamed, and Daan Wierstra.
\newblock Stochastic backpropagation and approximate inference in deep generative models.
\newblock In \emph{International Conference on Machine Learning}, pp.\  1278--1286, 2014.

\bibitem[Rombach et~al.(2022)Rombach, Blattmann, Lorenz, Esser, and Ommer]{rombach2022high}
Robin Rombach, Andreas Blattmann, Dominik Lorenz, Patrick Esser, and Bj{\"o}rn Ommer.
\newblock High-resolution image synthesis with latent diffusion models.
\newblock In \emph{IEEE Conference on Computer Vision and Pattern Recognition}, pp.\  10684--10695, 2022.

\bibitem[Sajjadi et~al.(2022)Sajjadi, Duckworth, Mahendran, van Steenkiste, Pavetic, Lucic, Guibas, Greff, and Kipf]{sajjadi2022object}
Mehdi~SM Sajjadi, Daniel Duckworth, Aravindh Mahendran, Sjoerd van Steenkiste, Filip Pavetic, Mario Lucic, Leonidas~J Guibas, Klaus Greff, and Thomas Kipf.
\newblock Object scene representation transformer.
\newblock \emph{Advances in Neural Information Processing Systems}, pp.\  9512--9524, 2022.

\bibitem[Santoro et~al.(2017)Santoro, Raposo, Barrett, Malinowski, Pascanu, Battaglia, and Lillicrap]{santoro2017simple}
Adam Santoro, David Raposo, David~G Barrett, Mateusz Malinowski, Razvan Pascanu, Peter Battaglia, and Timothy Lillicrap.
\newblock A simple neural network module for relational reasoning.
\newblock In \emph{Advances in Neural Information Processing Systems}, 2017.

\bibitem[Schölkopf et~al.(2021)Schölkopf, Locatello, Bauer, Ke, Kalchbrenner, Goyal, and Bengio]{scholkopf2021toward}
Bernhard Schölkopf, Francesco Locatello, Stefan Bauer, Nan~Rosemary Ke, Nal Kalchbrenner, Anirudh Goyal, and Yoshua Bengio.
\newblock Toward causal representation learning.
\newblock \emph{Proceedings of the IEEE}, 109\penalty0 (5):\penalty0 612--634, 2021.

\bibitem[Seitzer et~al.(2023)Seitzer, Horn, Zadaianchuk, Zietlow, Xiao, Simon-Gabriel, He, Zhang, Schölkopf, Brox, and Locatello]{seitzer2023bridging}
Maximilian Seitzer, Max Horn, Andrii Zadaianchuk, Dominik Zietlow, Tianjun Xiao, Carl-Johann Simon-Gabriel, Tong He, Zheng Zhang, Bernhard Schölkopf, Thomas Brox, and Francesco Locatello.
\newblock Bridging the gap to real-world object-centric learning.
\newblock \emph{arXiv preprint arXiv:2209.14860}, 2023.

\bibitem[Singh et~al.(2021)Singh, Peri, Kim, Kim, and Ahn]{singh2021structured}
Gautam Singh, Skand Peri, Junghyun Kim, Hyunseok Kim, and Sungjin Ahn.
\newblock Structured world belief for reinforcement learning in pomdp.
\newblock In \emph{International Conference on Machine Learning}, pp.\  9744--9755, 2021.

\bibitem[Singh et~al.(2022{\natexlab{a}})Singh, Deng, and Ahn]{singh2022illiterate}
Gautam Singh, Fei Deng, and Sungjin Ahn.
\newblock Illiterate dall-e learns to compose.
\newblock In \emph{International Conference on Learning Representations}, 2022{\natexlab{a}}.

\bibitem[Singh et~al.(2022{\natexlab{b}})Singh, Wu, and Ahn]{singh2022simple}
Gautam Singh, Yi-Fu Wu, and Sungjin Ahn.
\newblock Simple unsupervised object-centric learning for complex and naturalistic videos.
\newblock In \emph{Advances in Neural Information Processing Systems}, 2022{\natexlab{b}}.

\bibitem[Touvron et~al.(2023)Touvron, Martin, Stone, Albert, Almahairi, Babaei, Bashlykov, Batra, Bhargava, Bhosale, et~al.]{touvron2023llama}
Hugo Touvron, Louis Martin, Kevin Stone, Peter Albert, Amjad Almahairi, Yasmine Babaei, Nikolay Bashlykov, Soumya Batra, Prajjwal Bhargava, Shruti Bhosale, et~al.
\newblock Llama 2: Open foundation and fine-tuned chat models.
\newblock \emph{arXiv preprint arXiv:2307.09288}, 2023.

\bibitem[Van Den~Oord et~al.(2017)Van Den~Oord, Vinyals, et~al.]{van2017neural}
Aaron Van Den~Oord, Oriol Vinyals, et~al.
\newblock Neural discrete representation learning.
\newblock In \emph{Advances in Neural Information Processing Systems}, 2017.

\bibitem[Vaswani et~al.(2017)Vaswani, Shazeer, Parmar, Uszkoreit, Jones, Gomez, Kaiser, and Polosukhin]{vaswani2017attention}
Ashish Vaswani, Noam Shazeer, Niki Parmar, Jakob Uszkoreit, Llion Jones, Aidan~N Gomez, {\\L}ukasz Kaiser, and Illia Polosukhin.
\newblock Attention is all you need.
\newblock In \emph{Advances in Neural Information Processing Systems}, 2017.

\bibitem[Watters et~al.(2017)Watters, Zoran, Weber, Battaglia, Pascanu, and Tacchetti]{watters2017visual}
Nicholas Watters, Daniel Zoran, Theophane Weber, Peter Battaglia, Razvan Pascanu, and Andrea Tacchetti.
\newblock Visual interaction networks: Learning a physics simulator from video.
\newblock In \emph{Advances in Neural Information Processing Systems}, 2017.

\bibitem[Watters et~al.(2019)Watters, Matthey, Burgess, and Lerchner]{watters2019spatial}
Nicholas Watters, Loic Matthey, Christopher~P Burgess, and Alexander Lerchner.
\newblock Spatial broadcast decoder: A simple architecture for learning disentangled representations in vaes.
\newblock \emph{arXiv preprint arXiv:1901.07017}, 2019.

\bibitem[Webb et~al.(2023)Webb, Mondal, and Cohen]{webb2023systematic}
Taylor~W Webb, Shanka~Subhra Mondal, and Jonathan~D Cohen.
\newblock Systematic visual reasoning through object-centric relational abstraction.
\newblock \emph{arXiv preprint arXiv:2306.02500}, 2023.

\bibitem[Weis et~al.(2021)Weis, Chitta, Sharma, Brendel, Bethge, Geiger, and Ecker]{weis2021benchmarking}
Marissa~A Weis, Kashyap Chitta, Yash Sharma, Wieland Brendel, Matthias Bethge, Andreas Geiger, and Alexander~S Ecker.
\newblock Benchmarking unsupervised object representations for video sequences.
\newblock \emph{Journal of Machine Learning Research}, 22\penalty0 (1):\penalty0 8253--8313, 2021.

\bibitem[Wen et~al.(2023)Wen, Yang, Kautz, and Birchfield]{wen2023foundationpose}
Bowen Wen, Wei Yang, Jan Kautz, and Stan Birchfield.
\newblock Foundationpose: Unified 6d pose estimation and tracking of novel objects.
\newblock \emph{arXiv preprint arXiv:2312.08344}, 2023.

\bibitem[Wu et~al.(2021)Wu, Yoon, and Ahn]{wu2021generative}
Yi-Fu Wu, Jaesik Yoon, and Sungjin Ahn.
\newblock Generative video transformer: Can objects be the words?
\newblock In \emph{International Conference on Machine Learning}, pp.\  11307--11318, 2021.

\bibitem[Wu et~al.(2022)Wu, Dvornik, Greff, Kipf, and Garg]{wu2022slotformer}
Ziyi Wu, Nikita Dvornik, Klaus Greff, Thomas Kipf, and Animesh Garg.
\newblock Slotformer: Unsupervised visual dynamics simulation with object-centric models.
\newblock \emph{arXiv preprint arXiv:2210.05861}, 2022.

\bibitem[Wu et~al.(2023)Wu, Hu, Lu, Gilitschenski, and Garg]{wu2023slotdiffusion}
Ziyi Wu, Jingyu Hu, Wuyue Lu, Igor Gilitschenski, and Animesh Garg.
\newblock Slotdiffusion: Object-centric generative modeling with diffusion models.
\newblock In \emph{Advances in Neural Information Processing Systems}, pp.\  50932--50958, 2023.

\bibitem[Yang \& Yang(2024)Yang and Yang]{yang2024benchmarking}
Yafei Yang and Bo~Yang.
\newblock Benchmarking and analysis of unsupervised object segmentation from real-world single images.
\newblock \emph{International Journal of Computer Vision}, pp.\  1--37, 2024.

\bibitem[Yang et~al.(2022)Yang, Gan, Wang, Hu, Lu, Liu, and Wang]{yang2022empirical}
Zhengyuan Yang, Zhe Gan, Jianfeng Wang, Xiaowei Hu, Yumao Lu, Zicheng Liu, and Lijuan Wang.
\newblock An empirical study of gpt-3 for few-shot knowledge-based vqa.
\newblock In \emph{AAAI Conference on Artificial Intelligence}, pp.\  3081--3089, 2022.

\bibitem[Yang et~al.(2023)Yang, Li, Lin, Wang, Lin, Liu, and Wang]{yang2023dawn}
Zhengyuan Yang, Linjie Li, Kevin Lin, Jianfeng Wang, Chung-Ching Lin, Zicheng Liu, and Lijuan Wang.
\newblock The dawn of lmms: Preliminary explorations with gpt-4v (ision).
\newblock \emph{arXiv preprint arXiv:2309.17421}, 2023.

\bibitem[Yoon et~al.(2023)Yoon, Wu, Bae, and Ahn]{yoon2023investigation}
Jaesik Yoon, Yi-Fu Wu, Heechul Bae, and Sungjin Ahn.
\newblock An investigation into pre-training object-centric representations for reinforcement learning.
\newblock \emph{arXiv preprint arXiv:2302.04419}, 2023.

\bibitem[Yuan et~al.(2019)Yuan, Li, and Xue]{yuan2019generative}
Jinyang Yuan, Bin Li, and Xiangyang Xue.
\newblock Generative modeling of infinite occluded objects for compositional scene representation.
\newblock In \emph{International Conference on Machine Learning}, pp.\  7222--7231, 2019.

\bibitem[Zambaldi et~al.(2018)Zambaldi, Raposo, Santoro, Bapst, Li, Babuschkin, Tuyls, Reichert, Lillicrap, Lockhart, Shanahan, Langston, Pascanu, Botvinick, Vinyals, and Battaglia]{zambaldi2018relational}
Vinicius Zambaldi, David Raposo, Adam Santoro, Victor Bapst, Yujia Li, Igor Babuschkin, Karl Tuyls, David Reichert, Timothy Lillicrap, Edward Lockhart, Murray Shanahan, Victoria Langston, Razvan Pascanu, Matthew Botvinick, Oriol Vinyals, and Peter Battaglia.
\newblock Relational deep reinforcement learning.
\newblock \emph{arXiv preprint arXiv:1806.01830}, 2018.

\end{thebibliography}
\bibliographystyle{iclr2025_conference}

\appendix
\newpage
\section{Models and Implementation Details}
\label{appendix: models and implementation details}
Here we elaborate on the upstream and downstream models included in this study along with details on the training, the implementation, and hyperparameter choices.

\subsection{Upstream Models}
\label{appendix:upstream_models}
Here we elaborate on all the upstream models we use in our experiments and provide details on the implementation, training, and hyperparameter choices.

\paragraph{Implementation \& Training Details.}
Our code is based on the implementations of object-centric models of \citet{Dittadi2021} and we use their implementation of Slot Attention, MONet, SPACE, and VAE. For these models, we use the same set of recommended hyperparameters on CLEVR and Multi-dSprites, and apply the same hyperparameters for CLEVRTex as used in CLEVR. All other models are either re-implemented or adapted from available code. 
Unless explicitly stated otherwise, except for pre-trained foundation models, we train all the other models with a default batch size of $32$ for $3$ random seeds with Adam optimizer. The training finishes after $500$k steps on synthetic datasets and $250$k steps on real-world datasets. The reported metrics are then averaged over the seeds to provide a comprehensive assessment. Additionally, more information about pre-trained models and the size of the models are shown in \cref{tab:pretrained_training_details}. The details of each model are explained below.

\paragraph{DINOv2.} 
DINOv2 \citep{oquab2023dinov2}, an enhanced version of the DINO \citep{caron2021emerging}, stands out as a self-supervised ViT-based model designed for training high-performance computer vision models without the need for extensive labeled data. It serves as a versatile backbone for diverse tasks, including image classification, video action recognition, semantic segmentation, and depth estimation. Trained on a carefully curated dataset comprising $142$ million images with a discriminative self-supervised method, DINOv2 excels in producing versatile visual features that transcend specific image distributions and tasks without the necessity for fine-tuning. 

Similar to DINO, DINOv2 follows a transformer architecture, with a patch size of $14$, and is trained with $1$B parameters with a self-supervised learning objective, and distilled into a series of smaller models that generally surpass the other best available all-purpose features. There are $4$ distinct backbone versions, each varying in the number of transformer layers. After experimenting with all $4$, considering a balance between the performance and downstream training time, we selected the second-largest variant, denoted as \textit{ViT-L/14}. We employ this backbone without fine-tuning, and uniformly resize images from all datasets to dimensions of $224 \times 224$ and pass them to the model, generating fixed-region $16 \times 16$ representations with a channel size of $1024$ for each patch. We flatten the spatial dimensions and pass a matrix of size $256 \times 1024$ to the downstream model.

\paragraph{MAE.} The Masked Autoencoder (MAE) \citep{he2022masked} is a simple approach for reconstructing an original signal from a partial observation. In this approach, the image is divided into non-overlapping patches, with $75$\% of them randomly masked. These patches are then fed into a ViT-based encoder, which converts the partially observed input into a latent representation. Next, a lightweight decoder, which uses the representation and mask tokens, reconstructs the original image. The model is trained on ImageNet-1K \citep{deng2009imagenet} by minimizing the Mean Squared Error (MSE) between the reconstructed and original input in the pixel space.

There are various pre-trained MAEs available, differing in model sizes. In our framework, we utilize the pre-trained \textit{ViT-L/16}. We resize the input images to $224 \times 224$ and pass them to the encoder without masking. This generates a fixed-region representation sequence of length $197$ ($196$ corresponding to different regions in the image and one corresponding to the output of the \textit{CLS} token) with a latent size of $1024$ for each sequence. We pass the obtained representation matrix without any modifications to the downstream model.

\paragraph{CLIP.} The Contrastive Language-Image Pretraining (CLIP) \citep{radford2021learning} is a multimodal model that learns to associate images and corresponding text descriptions. CLIP uses a ViT to extract a feature vector that represents the visual content of the image. Similarly, for the text, CLIP uses a transformer-based model to generate a feature vector that represents the semantic content of the text. These two feature vectors are then projected into a shared embedding space and the cosine similarity between the two vectors is calculated. The model is trained on a dataset of $400$ million (image, text) pairs collected from a variety of publicly available sources on the Internet, using a contrastive loss function which is a symmetric cross-entropy loss over the similarity scores, that encourages the similarity between the image and text feature vectors to be high when they are a matching pair and low when they are not.

In our experiments, we utilize a pre-trained CLIP image encoder with a ViT architecture denoted as \textit{ViT-B/32}. Similar to DINOv2 and MAE, we resize the input images for all datasets to $224 \times 224$ and pass them to the encoder. The encoder produces a representation of size $50 \times 768$ (one corresponding to the output of the \textit{CLS} token and the rest corresponding to regions in the image) which we utilize directly in the downstream model.

\begin{table}[t]
  \centering
  \caption{Model sizes and training information of the models.}
  \begin{tabular}{lcccc}
    \toprule
    Model & Model Architecture & \# of Params$^\dagger$ & Pre-training Dataset & Dataset Size \\
    \midrule
    DINOv2 & ViT-L/14 & 304M & LVD-142M & 142M \\
    MAE & ViT-L/16 & 303M & INet-1k & 1.2M \\
    CLIP & ViT-B/32 & 87M & WIT-400M & 400M \\
    VQ-AE & - & 34.5M & OpenImages v4 & 9M \\
    KL-AE & - & 34.5M & OpenImages v4 & 9M \\
    ResNet50 & - & 23M & INet-21k & 14M \\
    CNN & - & 0.1M & - & - \\
    MultiCNN & - & 1.4M & - & - \\
    SA & - & 0.9M & - & - \\
    ResNet SA & - & 22.4M & - & - \\
    MONet & - & 1.6M & - & - \\
    SPACE & - & 5.3M & - & - \\
    STEVE & - & 17.1M & - & - \\
    DINOSAURv2$^*$ & - & 4M + 304M & - & - \\
    VAE & - & 19.2M & - & - \\
    \midrule
    \multicolumn{5}{l}{\footnotesize{$^*$\# of parameters of pre-trained DINOv2 is shown separately.}} \\
    \multicolumn{5}{l}{\footnotesize{$^\dagger$ Except for foundation models, \# of parameters are reported on CLEVR.}} \\
    \bottomrule
  \end{tabular}
  \label{tab:pretrained_training_details}
\end{table}

\paragraph{VQ-AE \& KL-AE.} 
VQ-AE and KL-AE are two pre-trained autoencoders of the latent diffusion model (LDM) \citep{rombach2022high}. In LDM, they don't directly use a diffusion model in pixel space. Instead, to facilitate training on constrained computational resources without compromising quality and flexibility, they apply the models in the latent space of a powerful autoencoder pre-trained on OpenImages \citep{OpenImages} with $9$M images in an adversarial manner. The autoencoder consists of an encoder that downsamples the images by a factor $f$, and a decoder that reconstructs the original image. In order to avoid arbitrarily high-variance latent spaces, they apply two different regularizations: one imposes a KL penalty towards a standard normal on the learned latent (KL-AE), similar to a VAE, and the other one uses a vector quantization layer \citep{van2017neural} within the decoder (VQ-AE). 

Various pre-trained autoencoders with different downsampling factors are available. In our experiments, we explore multiple models with varying downsampling factors and find that a factor of $16$ is the balancing point between performance and training speed. Therefore, we adopt models with this factor for further analysis. In our experiments, we utilize the encoder of these two autoencoders off-the-shelf. By applying the encoders, we get a vector of size $W/16 \times H/16 \times D_{fr}$ where $D_{fr}$ is $16$ and $8$ for KL-AE and VQ-AE, respectively. Similar to DINOv2, we flatten $2$d feature maps and feed them into the downstream model.

\paragraph{ResNet50.} ResNet50 \cite{he2016deep} is a deep neural network with 50 layers that has been pre-trained on ImageNet-21k \citep{deng2009imagenet}. It is well-known for its residual learning blocks and serves as a baseline in our study. In our experiments, we employ the off-the-shelf ResNet50 model and remove the pooling and fully connected layers at the end. We apply the default ResNet50 transformations on the input image, pass it to the model, and obtain a vector of size $W/32 \times H/32 \times 2048$ which we flatten the spatial dimensions and pass to the downstream model.

\paragraph{CNN.} CNN \citep{zambaldi2018relational} is a small convolutional neural network that is commonly used in the literature. It consists of a few convolutional layers with ReLU activation functions in between, and is trained end-to-end with the downstream model on the downstream task. It produces a fixed-size fixed-region representation of size $4 \times 4 \times 64$ which we flatten the spatial dimensions and produce a vector of size $16 \times 64$ and pass it to the downstream model. The hyperparameters of CNN are shown in \cref{table:hp_CNN}.

\paragraph{MultiCNN} MultiCNN \citep{kipf2019contrastive} is an object-centric model consisting of $N_{slots}$ CNNs, each dedicated to detecting a single object within an image. These CNNs operate with non-shared parameters, processing each input image independently. MultiCNN shares the same dataset-specific hyperparameters as the CNN baseline, and similar to CNN, it is trained end-to-end using the same training hyperparameters and loss function as the downstream model. The output of each CNN undergoes complete flattening, followed by a shared linear layer of size $64$, resulting in a representation of size $N_{slots} \times 64$, which is then passed to the downstream model.

\paragraph{Slot Attention.} Slot Attention \citep{locatello2020object} has become the primary representative for object-centric (OC) learning in recent years. It follows an autoencoder setup and begins with a Convolutional Neural Network (CNN) and is followed by the Slot Attention module. This module refines the initial image features through multiple iterations, turning them into distinct slots representing objects. Each slot is updated using a Gated Recurrent Unit (GRU) that takes the current slot and attention information as inputs. After refining, these slots are used to reconstruct the appearance and mask of each object, which are then combined to reconstruct the original image. The model is trained by minimizing the MSE reconstruction loss.

We also employ an improved version of Slot Attention introduced in \citet{biza2023invariant} which we refer to as \textit{ResNet SA}. In this improved version, the CNN backbone is replaced by a ResNet34 \citep{he2016deep} without pre-training, and a larger feature map resolution of $16$ on synthetic datasets and $7$ on VQA-v2 and GQA is used in both the encoder and the decoder of the model. Furthermore, the initial slots are changed into learnable slots. Both the original and improved models are trained on each dataset with a batch size of $64$, a learning rate of $0.0004$, a learning rate warmup of $10$k steps, and an exponential learning rate decay with a half-life of $100$k steps. For ResNet SA, we additionally clip the gradient norm at $0.05$ to stabilize training. We follow the same architecture for ResNet SA as in \citet{biza2023invariant} on synthetic datasets. On VQA-v2 and GQA, we modify the architecture by replacing the initial convolutional layer of ResNet34 with a convolutional layer featuring a kernel size of $7 \times 7$ and a stride of $4$. Additionally, in the decoder, we incorporate two additional transpose convolutional layers at the beginning, each configured with the same hyperparameters as the existing transpose convolutional layers. After training, the learned slot vectors of size $N_{slots} \times 64$ are used as representations in the downstream model for both versions.

\begin{table}[t]
    \caption{Hyperparameters of CNN.}
    \centering
    \begin{tabular}{lcccc}
        \toprule
        \multicolumn{5}{l}{\textbf{CNN}} \\
        \toprule
        \textbf{Dataset} & \textbf{Kernel Size} & \textbf{Stride} & \textbf{Output Channels} & \textbf{Activation Function} \\
        \midrule
        CLEVR \& CLEVRTex & 8 & 4 & 32 & ReLU \\
              & 4 & 2 & 64 & ReLU \\
              & 4 & 2 & 64 & ReLU \\
              & 3 & 1 & 64 & ReLU \\
        \midrule
        Multi-dSprites & 8 & 4 & 32 & ReLU \\
                      & 4 & 2 & 64 & ReLU \\
                      & 3 & 1 & 64 & ReLU \\
        \bottomrule
    \end{tabular}
    \label{table:hp_CNN}
\end{table}

\paragraph{MONet.} The Multi-Object Network (MONet) \citep{burgess2019monet} consists of a recurrent segmentation network that generates attention masks that represent the probability of each pixel belonging to each object. For each slot, a VAE (the component VAE) encodes the image and the current attention mask, and decodes the latent representation to an image reconstruction of the slot and the slot mask. To create the final reconstructed image, the reconstructed images are combined using the attention masks obtained from the segmentation network. The model is trained by an objective function comprising a reconstruction loss defined as the negative log-likelihood of a spatial Gaussian mixture model (GMM) with one component per slot, where each pixel is modeled independently, and a KL divergence of the component VAE, and an additional mask reconstruction loss for the component VAE. The mean of the GMM for each component is used as the representation of each object. The learned representation is a vector of size $N_{slots} \times 16$ and is directly used for the downstream task.

\paragraph{SPACE.} Spatially Parallel Attention and Component Extraction (SPACE) \citep{lin2020space} provides a unified probabilistic modeling framework that combines the best of spatial attention and scene-mixture approaches. Foreground objects are identified using bounding boxes computed in a parallel spatial attention process, and background elements are modeled using a mixture of components. The model is trained by optimizing the Evidence Lower Bound (ELBO) of the probabilistic model. An additional boundary loss is introduced to penalize the splitting of objects across bounding boxes, addressing potential under- or over-segmentation issues. SPACE representations are vectors of size $69 \times 38$ where $69$ is the number of slots that is determined by the grid size, and the latent representation of each slot is obtained by concatenating all the latent variables which will have a dimension of $38$. We utilize this representation directly in the downstream model.

\begin{table}[t]
    \caption{Hyperparameters of STEVE.}
    \centering
    \begin{tabular}{llc}
        \toprule
        \multicolumn{3}{l}{\textbf{STEVE}} \\
        \toprule
        \textbf{Module} & \textbf{Hyperparameter} & \textbf{Hyperparameter Value} \\
        \midrule
        Encoder & Corrector Iterations & 2 \\
        & Slot Size & 192 \\
        & MLP Hidden Size & 192 \\
        & \# Predictor Blocks & 1 \\
        & \# Predictor Heads & 4 \\
        & Learning Rate & 0.0001 \\
        \midrule
        Transformer Decoder & \# Decoder Blocks & 8 \\
        & \# Decoder Heads & 4 \\
        & Hidden Size & 192 \\
        & Dropout & 0.1 \\
        & Learning Rate & 0.0003 \\
        \midrule
        DVAE & Learning Rate & 0.0003 \\
        & Patch Size & 4 $\times$ 4 pixels \\
        & Vocabulary Size & 4096 \\
        & Temperature Start & 1.0 \\
        & Temperature End & 0.1 \\
        & Temperature Decay Steps & 30k \\
        \bottomrule
    \end{tabular}
    \label{table:hp_steve}
\end{table}

\paragraph{STEVE.} STEVE \citep{singh2022simple} is a simple object-centric video model achieving remarkable performance over various complex and naturalistic videos. It is a more robust version of SLATE \citep{singh2022illiterate}, a state-of-the-art object-centric model. The model contains two reconstruction paths. The first path uses a discrete VAE encoder to convert the input image into discrete tokens, and then a discrete VAE decoder to reconstruct the original image. This path is trained using MSE reconstruction loss. The second path uses a CNN-based image encoder on the input, and the output is fed into a recurrent slot encoder that updates slots over time using recurrent neural networks. Finally, a slot-transformer decoder, similar to SLATE, is applied to the produced slots to predict and reconstruct the discrete tokens of the input image. This path is trained by minimizing the cross-entropy loss between the original tokens produced by the discrete VAE encoder, and
the predicted tokens of the slot-transformer decoder.

We incorporate the original implementation of STEVE into our framework. STEVE works on videos, considering the input to be a sequence of $t$ image frames. Following the authors' recommendation, to utilize it on images, we treat images as $1$-frame videos and pass them to the model. On synthetic datasets, we train STEVE for $500$k steps with an exponential learning rate decay with a half-life of $250$k steps and with $30$k warm-up steps. For , we employ the same training hyperparameters but reduce the number of steps to $250$k. After the training, we use slots of size $N_{slots} \times 192$ obtained from the recurrent slot encoder in the downstream model. 

A summary of the model's hyperparameters on all datasets is shown in \cref{table:hp_steve}. We maintain the original architecture for the CNN backbone and discrete VAE encoder/decoder for synthetic datasets. However, for VQA-v2 and GQA with image sizes of $224 \times 224$, we reduce the feature map dimensions of both the discrete VAE encoder and CNN backbone by half. This adjustment includes changing the stride of the first convolutional layer of the CNN backbone to $4$. Additionally, in the discrete VAE encoder, we insert a convolutional layer with a $2 \times 2$ kernel and stride $2$, followed by ReLU activation, after the initial convolutional layer. To accommodate these feature map changes in the discrete VAE decoder, we duplicate the four convolutional blocks and the pixel shuffling layer before the final convolutional block.

It is noteworthy that the training of STEVE is not entirely stable, as we observed that upstream performance metrics begin to degrade after some training time (after roughly 20-40k steps). However, when training the model for longer, we observe that it performs well on the downstream task. Nevertheless, in some seeds, the training fails as the upstream metrics are significantly worse than in other seeds. Therefore, we only consider the seeds that are stable and perform the best in terms of reconstruction and segmentation quality.

Additionally, we experimented with a modified version of STEVE using a pre-trained DINOv2 as the CNN encoder backbone. However, it performed similarly or worse than the original STEVE on the downstream VQA. We suspect this may be due to the chosen hyperparameters or an architectural bottleneck in the discrete VAE. As a result, we decided not to include it in our reported results.

\begin{table}[t]
    \caption{Hyperparameters of DINOSAURv2.}
    \centering
    \setlength{\tabcolsep}{2pt}
    \begin{tabular}{llccc}
        \toprule
        \multicolumn{4}{l}{\textbf{DINOSAURv2}} \\
        \toprule
        \multirow{2}{*}{\textbf{Hyperparameter}} & & \textbf{CLEVR \&} & \textbf{CLEVRTex} & \textbf{VQA-v2 (COCO) \&} \\
        & & \textbf{Multi-dSprites} & & \textbf{GQA} \\
        \midrule
        Training Steps & & 250k & 500k & 250k \\
        Batch Size & & 64 & 64 & 64 \\
        LR Warmup Steps & & 10k & 10k & 10k \\
        Peak LR & & 0.0004 & 0.0004 & 0.0004 \\
        Exp. Decay Half-Life & & 100k & 100k & 100k \\
        ViT Architecture & & ViT-L & ViT-L & ViT-L \\
        Patch Size & & 14 & 14 & 14 \\
        Feature Dim. & & 1024 & 1024 & 1024 \\
        Gradient Norm Clipping & & 1.0 & 1.0 & 1.0 \\
        \midrule
        Image Size & & 224 & 224 & 224 \\
        Cropping Strategy & & Full & Full & Full \\
        Image Tokens & & 256 & 256 & 256 \\
        \midrule
        \multirow{3}{*}{Decoder} & Type & MLP & MLP & MLP \\
        & Layers & 4 & 4 & 4 \\
        & MLP Hidden Dim. & 1024 & 512 & 2048\\
        \midrule
        \multirow{3}{*}{Slot Attention} & Iterations & 3 & 3 & 3 \\
        & Slot Dim. & 256 & 256 & 256 \\
        & MLP Hidden Dim. & 1024 & 1024 & 1024 \\
        \bottomrule
    \end{tabular}
    \label{table:hp_dinosaur}
\end{table}

\paragraph{DINOSAUR.} DINO and Slot Attention Using Real-world data (DINOSAUR) \citep{seitzer2023bridging} is an object-centric model designed to bridge the gap between object-centric models and real-world data. It consists of an encoder that extracts features from the input data, a slot attention module that groups the extracted features into slots, and a decoder that reconstructs the extracted features. Their approach can be considered similar to SLATE \citep{singh2022illiterate} and STEVE \citep{singh2022simple}, but with the difference of reconstructing global features from a pre-trained Vision Transformer \citep{dosovitskiy2020image} instead of local features from a VQ-VAE \citep{van2017neural}.

We adapt the original implementation of DINOSAUR into our framework and replace the pre-trained DINO backbone with pre-trained DINOv2 \citep{oquab2023dinov2}, which we refer to as DINOSAURv2. Similar to the original training procedure of DINOSAUR, the input images are resized to $224 \times 224$ and we train DINOSAURv2 for $500$k steps on CLEVRTex, and $250$k steps on CLEVR, Multi-dSprites, COCO images of VQA-v2, and GQA images. The training uses a learning rate of $0.0004$, a learning rate warm-up of $10$k optimization steps, and an exponentially decaying learning rate schedule. Furthermore, we clip the gradient norm at $1$ to stabilize training. After the training, we use the slots of size $N_{slots} \times 256$ obtained from the Slot Attention module in the downstream model. The full hyperparameters of the model are provided in \cref{table:hp_dinosaur}.

\paragraph{VAE.} We train a variational autoencoder (VAEs) \citep{kingma2014vae} with a broadcast decoder \citep{watters2019spatial} as a baseline that learns global representations. We use the broadcast VAE implementation of \citet{Dittadi2021} with the same architecture and hyperparameter choices. For CLEVRTex, we use the same hyperparameters as in CLEVR. The latent size is selected to be $64$ times the number of slots used when training an object-centric model on the same dataset. As explained in \cref{subsection:framework_setup}, we divide the flat representation vector into $N_{slots}$ vectors of size $64$, and pass them to the downstream model.

\subsection{Text Embedding Module}
\label{appendix:text_embedding_module}
As our text embedding module, we use Text-to-Text Transfer Transformer (T5) \citep{raffel2020exploring}, a transformer-based language model developed by Google AI Language. It is capable of performing a wide range of natural language processing tasks such as text classification, question answering, summarization, and translation. The model is trained on the colossal, cleaned version of Common Crawl's web crawl corpus (C4), an 806-gigabyte corpus of text data using a pretext task called Text-to-Text Transfer Transformer (T5), which involves converting a given input text to a target output text. We utilize the available implementation in Hugging Face’s Transformers library. T5 comes in different sizes and we use the T5-base tokenizer and encoder which produces the representations of size $768$ for each token.

\subsection{Downstream VQA Setup}
\label{appendix:downstream_model}

\paragraph{Architecture and Hyperparameters.}
We use a transformer-based architecture \citep{vaswani2017attention} which is the standard downstream architecture for the VQA task \citep{Ding2021, devlin2018bert, lu2019vilbert}. We utilize the original PyTorch implementation of the transformer. We first project the image and the text representations with two separate linear layers with a $126$ size and a dropout of $0.1$. Then, we augment the image and text vectors with a $2$-dimensional one-hot vector indicating whether the input is from the image representation or the text embeddings. A sinusoidal positional encoding is then added to the text embeddings. We introduce a trainable vector $\textit{CLS} \in \R^{128}$, akin to the \textit{CLS} token in \textit{BERT} \citep{devlin2018bert}, to generate classification results. The image and text representations, along with the \textit{CLS} token, are concatenated and passed through a transformer encoder with a d\_model of $128$ and a hidden dimension of $128$. The transformed value of the \textit{CLS} token is passed through a classifier MLP that generates a probability vector over all possible answers in each dataset. The MLP consists of $2$ linear layers of size $128$. A normalization layer, a dropout of $0.1$, and a ReLU activation function are applied in between the layers.

\paragraph{Training.}
On synthetic datasets, all downstream models are trained with a batch size of $128$ and a learning rate of $0.0001$ for $600$k training steps, with the cross-entropy loss. However, when using DINOv2 and MAE as upstream models, it is infeasible to keep the current batch size with only one GPU due to the substantial sequence length of the representations. To ensure a fair comparison, gradients are accumulated, and the optimizer is applied every $4$ training step with a reduced batch size of $32$. Consequently, the downstream model is trained for $2.4$ million steps, $4$ times the default number of steps. 

On GQA, the downstream models are trained for $900$k steps with a batch size of $32$ and a learning rate of $0.0001$, using the same loss function as for synthetic datasets. On VQA-v2, we train all downstream models with the same batch size and loss function for $300$k steps. Furthermore, on VQA-v2, we use a learning rate of $0.0001$ for T-2 and T-5, and $0.00005$ for T-15. However, we find that T-2 outperforms the other downstream models, with performance degrading significantly as the number of layers increases which is due to overfitting caused by the much smaller training size of VQA-v2 compared to other datasets in our study. As a result, we only report results for T-2 on VQA-v2.

\subsection{Downstream Property Prediction Setup}
\label{appendix:downstream_property_prediction}
Here we assess representations by training downstream models to predict ground-truth object properties from these representations. Following the approach outlined by \citet{Dittadi2021} for object property prediction on synthetic datasets. in summary, we employ a single downstream model $f$ to predict the properties of each object independently. To be more specific, for OC models, we apply $f$ on each slot representation, i.e. $\hat{y}_k = f(z_k)$, where $z_k$ is the $k$th slot representation and $\hat{y}_k$ is the predicted properties of the $k$th slot. Similarly, for models with fixed-region representations, we treat each region as a slot and apply the same approach as with OC models. We find this approach to be effective for these models. Lastly, for models with global representations, since the representations of individual objects are not readily available, we adopt the same strategy as demonstrated in \citet{Dittadi2021} which has proven to be working. We predict the properties of all objects using the downstream model $f$ and split it into $K$ vectors $\{\hat{y}_k\}_{k=1}^K$ where $K$ roughly corresponds to the number of slots in an OC model.

Since the predicted properties might not correspond to the objects in the same order as the ground-truth objects, and the number of slots can exceed the number of objects in the image, we follow \citet{Dittadi2021, locatello2020object} in using the same loss-matching algorithm to match $\hat{y}_k$ with its corresponding ground-truth vector. We define $f$ as an MLP with $1$ hidden layer of size $256$, and we utilize cross-entropy loss for categorical properties and MSE for numerical properties. We train $f$ for one seed on $10000$ images of each dataset using Adam optimizer with a learning rate of $0.001$ and a batch size of $64$ for $6000$ steps. For a randomly selected test set of $2000$ images, we calculate the accuracy for categorical properties and the adjusted $R^2$ for numerical properties and report the results.

\section{Datasets}
\label{appendix:datasets}
We work with $3$ existing multi-object datasets: \textit{Multi-dSprites}, \textit{CLEVR}, and \textit{CLEVRTex}. We use the common format for these datasets as outlined in \citet{Dittadi2021}. We primarily focus on these datasets due to their unified question-generation procedure and complete access to all underlying ground-truth object properties. Additionally, we utilize \textit{VQA-v2} and \textit{GQA} to extend our study to real-world settings. All datasets are summarized in \cref{table:datasets}. More details about each dataset are provided in the next subsections.

\subsection{Overview of Datasets}

\paragraph{CLEVR}
\textit{CLEVR} \citep{johnson2017clevr} comprises $128 \times 128$ images depicting $3$D scenes with a plain gray background featuring up to $10$ objects that may partially occlude one another. Objects vary in colors ($8$ options in total), materials (rubber or metal), shapes (sphere, cylinder, cube), sizes (small or large), and positions (x and y) as well as rotations. Following previous works \citep{locatello2020object, Dittadi2021, greff2019multi}, we utilize the CLEVR6 variant to learn object-centric representations in which the number of objects is limited to $6$. The dataset has been cropped and resized according to the procedure detailed originally by \citet{burgess2019monet}.

\paragraph{CLEVRTex}
The \textit{CLEVRTex} dataset extends the original \textit{CLEVR} \citep{johnson2017clevr} by incorporating textures on object surfaces, introducing a more visually complex environment. Each scene in \textit{CLEVRTex} consists of $3$-$10$ objects with distinct shapes (cube, cylinder, sphere, monkey head), sizes (small, medium, large), and textures ($60$ in total), contributing to the diversity of visual features. The backgrounds also present complex textures compared to the plain gray ones in CLEVR. The dataset is designed to facilitate the exploration of models' abilities in handling textured objects, providing a valuable resource for evaluating the performance of vision-related tasks in the context of rich visual scenes.

\paragraph{Multi-dSprites}
This dataset is derived from the dSprites dataset \citep{dsprites17} of $64 \times 64$ synthetic images. Following prior research \citep{Dittadi2021, locatello2020object, greff2019multi}, we utilize the Multi-dSprites variant, featuring colored sprites set against a grayscale background where the intensity of the uniform grayscale background is randomly determined for each image. Each scene consists of $2$–$5$ objects with randomized attributes, including shapes (ellipse, square, heart), sizes (selected from $6$ discrete values in $[0.5, 1]$, and converted to small and large with a threshold of $0.8$), x and y positions, orientation, and color (randomly sampled in HSV space). Objects might occlude one another, with certain objects being nearly entirely concealed by others in specific images. Consequently, we eliminate images where an object is significantly obscured by another object, leaving only those with clearer visibility of individual objects. We use $4$ different training sizes of this dataset which is demonstrated in \cref{table:datasets}.

\paragraph{VQA-v2} \textit{VQA-v2} \citep{goyal2017making, antol2015vqa} pairs open-ended questions with images from the MS COCO 2014 dataset \citep{lin2014microsoft}. It serves as a benchmark for evaluating how well models can comprehend and reason about visual information in real-world scenarios. The questions cover a wide range of topics and require a detailed understanding of the image content, spanning from basic inquiries about object presence to more complex queries about relationships and attributes within the scene. Answers are categorized into \textit{Yes/No}, \textit{Number}, and \textit{Other} types. Each question has 10 ground-truth answers, with the most frequent ground-truth answer considered correct in our evaluation framework. We filter out specific questions in the dataset based on their answer type, as explained in detail in \cref{appendix:question_generation}. We train downstream models using the filtered \textit{train} split of the dataset. However, since answer types for the \textit{test} split are not available, we cannot apply the same filtering process to evaluate on this split properly. Therefore, we utilize the \textit{val} split as our test set for evaluation, and additionally select a subset of $10$k questions from it as our validation set during training.

The \textit{MS COCO} (Microsoft Common Objects in Context) dataset \citep{lin2014microsoft} is a widely used collection of images designed for object detection, segmentation, and captioning tasks. It contains images sourced from everyday scenes that are diverse in content, encompassing various objects, activities, and environments. The dataset includes over $200$k labeled images across $80$ object categories, such as people, animals, vehicles, and indoor objects, with a significant emphasis on diversity in scenes and object appearances. Following \citet{seitzer2023bridging}, we resize all the images to $224 \times 224$ without any cropping, while ignoring the aspect ratio. We use the original split of the dataset when training and evaluating the upstream models. Furthermore, during upstream training, we augment the dataset by horizontally flipping images with a probability of $0.5$.

\paragraph{GQA} \textit{GQA} \citep{hudson2019gqa} is a benchmark for visual reasoning, featuring open-ended questions about images that require a detailed understanding of object relationships and spatial reasoning. it provides both questions and corresponding scene graphs, making it particularly suited for evaluating relational and compositional reasoning in multi-object environments. This makes GQA a challenging and insightful benchmark for assessing object-centric and foundation models alike. The dataset is available in both balanced and unbalanced versions. The balanced version ensures a uniform distribution of answers across question categories, focusing on reasoning over memorization, while the unbalanced version reflects more natural distributions of answers. In our work, we utilize the balanced version, with the \textit{train} split used for training and the \textit{testdev} split used for evaluation. Furthermore, we use the same image preprocessing and augmentations as for VQA-v2.

\begin{table*}[t]
    \caption{Dataset splits for upstream and downstream training, number of slots used in training slot-based object-centric models, and count of unique answers to questions in each dataset.}
    \label{table:datasets}
    % \vskip -0.15in
    \begin{center}
        \setlength{\tabcolsep}{4pt}
        \begin{tabularx}{0.98\linewidth}{lcclcc}
            \toprule
            \multirow{2}{*}{\textbf{Dataset Name}} & \multirow{2}{*}{\textbf{Slots}$^*$} & \multirow{2}{*}{\textbf{Answers}} & \multicolumn{3}{c}{\textbf{Dataset Splits}$^\dagger$} \\\cmidrule(l{0.6em}r{1.7em}){4-6}
            & & & \textbf{Train Size} & \textbf{Validation Size} & \textbf{Test Size} \\
            \midrule
            CLEVR6 & $7$ & $24$ & $2314980$ ($49000$) & $70702$ ($1500$) & $70997$ ($1500$) \\
            CLEVRTex & $11$ & $20$ & $1489005$ ($40000$) & $55977$ ($1500$) & $55646$ ($1500$) \\
            \midrule
            \multirow{4}{*}{Multi-dSprites$^\ddagger$} & \multirow{4}{*}{$6$} & \multirow{4}{*}{$22$} & $1545444$ ($320000$) & $57816$ ($1500$) & $57557$ ($1500$) \\
            & & & $3089425$ ($320000$) & $57540$ ($1500$) & $57557$ ($1500$) \\
            & & & $6181799$ ($320000$) & $58289$ ($1500$) & $57557$ ($1500$) \\
            & & & $12365042$ ($320000$) & $57792$ ($1500$) & $57557$ ($1500$) \\
            \midrule
            VQA-v2 (COCO) & 7 & 17 & 215553 (82753) & 10000 (10000) & 103717 (40504) \\
            GQA & 7 & 1834 & 943000 (72140) & 132044 (10234) & 12576 (398) \\
            \hline
            \multicolumn{6}{l}{\footnotesize{$^*$ We define it as the maximum number of objects plus one additional slot for the background.}}\\
            \multicolumn{6}{l}{\footnotesize{$^\dagger$ The values in parentheses denote the size of the corresponding image split used for the upstream model.}}\\
            \multicolumn{6}{l}{\footnotesize{$^\ddagger$ For Multi-dSprites, We use $4$ different training sizes.}}\\
            \bottomrule
        \end{tabularx}
    \end{center}
\end{table*}

\subsection{Question Generation \& Preprocessing}
\label{appendix:question_generation}
\paragraph{Synthetic.}
Originally, excluding CLEVR, the synthetic datasets exclusively comprise images without associated questions. Consequently, for their transformation into Visual Question Answering (VQA) datasets, we employ a question generation mechanism based on \citet{johnson2017clevr}, adapted for all datasets. The process involves utilizing a script that takes as input a JSON file that contains scene information and object features in the dataset. The script outputs a JSON file containing questions for each image. To generate questions for each image, there are $9$ different question templates that take the features of the objects in each image and generate a question based on the template. These templates are adjusted for each dataset, producing a maximum of $50$ questions for each image in each dataset. However, due to some images featuring only a few objects, the average number of questions per image tends to be less than $50$ in each dataset. Additionally, we refrain from utilizing pre-existing questions for CLEVR. Instead, we generate questions for it to establish a standardized question-generation process for all datasets.

Question templates can be categorized into $5$ categories:
\begin{enumerate}
    \item \textbf{Counting}: Counting questions ask for the number of objects meeting specific criteria (\textit{e.g. ``How many other things are there of the same size as the tiny metal block?''}).
    \item \textbf{Existence}: Existence questions ask if an object with certain properties is present in the image (\textit{e.g. ``Are there any other things that are the same color as the cylinder?''}).
    \item \textbf{Integer Comparison}: Integer comparison questions ask about the relative sizes of two sets of objects (\textit{e.g. ``Are there fewer spheres than large cylinders?''}).
    \item \textbf{Comparing Attributes}: Attribute comparison questions check if two objects have the same value for an attribute (\textit{e.g. ``Do the red cube and the green cylinder have the same size?''}).
    \item \textbf{Querying Attributes}: Query questions ask about an attribute of a particular object (\textit{e.g. ``What shape is the object at the right of the green cube?''}).
\end{enumerate}

\begin{figure}[t]
    \centering
    \includegraphics[width=\textwidth]{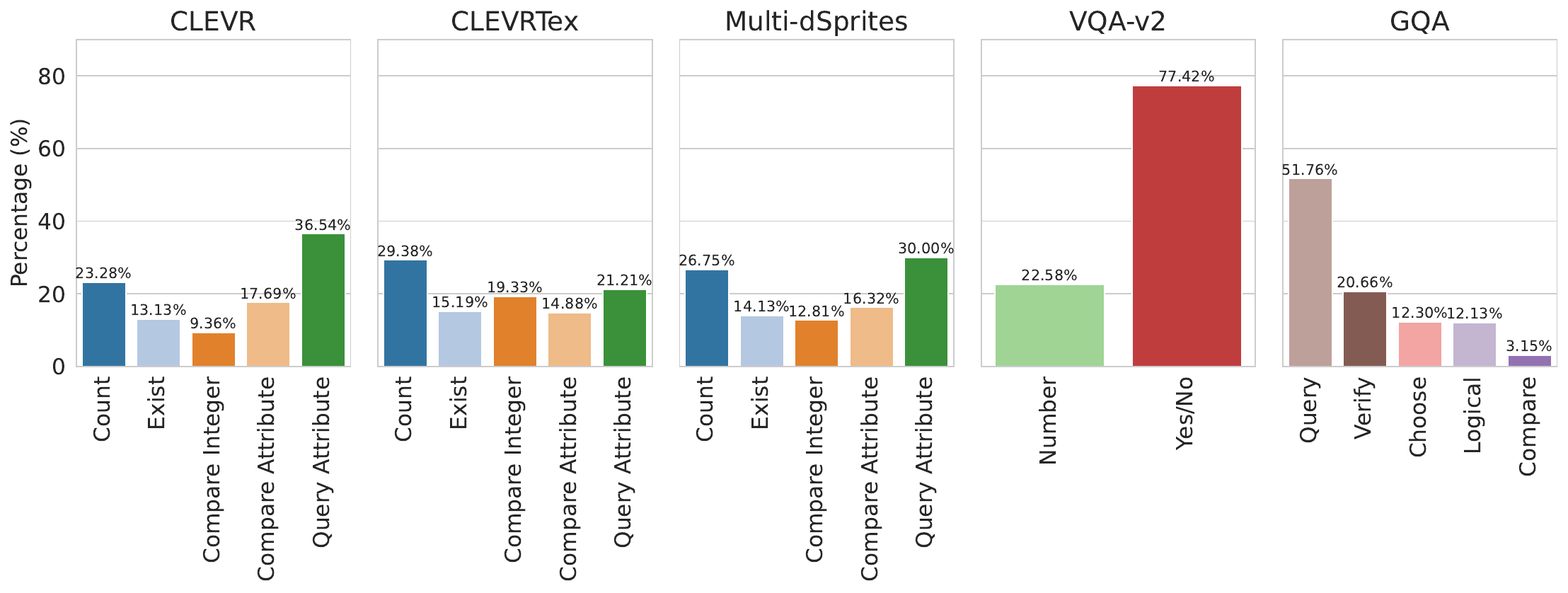}
    \caption{Distribution of question categories per dataset.}
    \label{fig:question_type_distribution}
\end{figure}

The set of possible answers includes yes, no, numerical values between $0$ and the maximum number of objects in the dataset, and all possible values of object properties. We have a different number of templates per each question category in our question-generation process. The proportion of each question category in each dataset is outlined in \cref{fig:question_type_distribution}.

For CLEVR, questions are generated based on the attributes of each object, including shape, color, size, material, and their relative position in comparison to other objects. In the case of Multi-dSprites, where objects lack a material attribute, questions focus on the shape, color, size, and relative position of the objects. Regarding CLEVRTex, questions are generated considering only shape, size, and relative position, as color is absent in the dataset and is replaced with $60$ materials, each having a specific color. The material is not included as a feature for the questions because the names of the materials are not suitable for accurate processing by a text embedding module.

\paragraph{Real-World.} The VQA-v2 dataset includes questions from three categories: \textit{Yes/No}, \textit{Number}, and \textit{Other}, with open-ended answers that are not limited to specific words or numbers. Therefore, in order to use the same framework employed for synthetic datasets, we only keep a subset of questions with specific answers. We omit questions with "Other" answer types and keep "Yes/No" questions that have "yes" or "no" answers. Additionally, we retain questions with numerical answers ranging from $0$ to $14$, resulting in a total of $17$ possible answers in the "Yes/No" and "Number" categories. Before filtering, the train and test sets contain 443,757 and 214,354 questions, respectively. After filtering, they are reduced to 215,553 and 103,717. 

The GQA dataset contains questions from five structural categories: \textit{Verify}, \textit{Logical}, \textit{Compare}, \textit{Query}, \textit{Choose}. \textit{Verify} questions are yes/no queries that check for the presence or attributes of objects. \textit{Logical} questions demand inference and multi-step reasoning. \textit{Compare} questions require evaluating two or more objects against each other. \textit{Query} questions are open-ended, asking for descriptive details about the scene. Finally, \textit{Choose} questions present two alternatives from which the answer must be selected. Furthermore, unlike VQA-v2, on GQA, the number of unique answers is not significantly large, making it possible to use the pipeline as-is without any deletion.

\section{Metrics}
\label{appendix:metrics}
\paragraph{Upstream.} To evaluate the performance of upstream models trained on the datasets used in our study, we use MSE reconstruction error and Adjusted Rand Index (ARI) \citep{hubert1985comparing}, which are two commonly used metrics in the literature \citep{locatello2020object, Dittadi2021, singh2022simple, biza2023invariant}. Consistent with prior work, we calculate the ARI considering only foreground objects. Additionally, following \citet{Dittadi2021}, we include Segmentation Covering \citep{arbelaez2010contour} and mean Segmentation Covering \citep{engelcke2019genesis} as evaluation metrics. For more information on the upstream metrics, we refer to \citet{Dittadi2021}.

\paragraph{Downstream VQA.} To assess the performance of representations in the VQA downstream task, following previous works on applying the VQA task to object-centric models \citep{Ding2021, wu2022slotformer, johnson2017clevr} and other works on VQA in general \citep{antol2015vqa, goyal2017making, ren2015exploring, yang2022empirical}, we use accuracy as our main metric. We also analyzed balanced accuracy which takes into account the class imbalances and is defined as the average of recall obtained in each class, and F1 score as alternative metrics but our results revealed consistent trends across all metrics. Consequently, we focus on presenting results based on the accuracy metric.

% \newpage

\section{Additional Results}
\label{appendix:results}

In this section, we report additional results that did not fit into the main part.

\subsection{Property Prediction vs VQA}
\label{appendix:property_prediction_vs_vqa}
\begin{figure}[h]
    \centering
    \includegraphics[width=\textwidth]{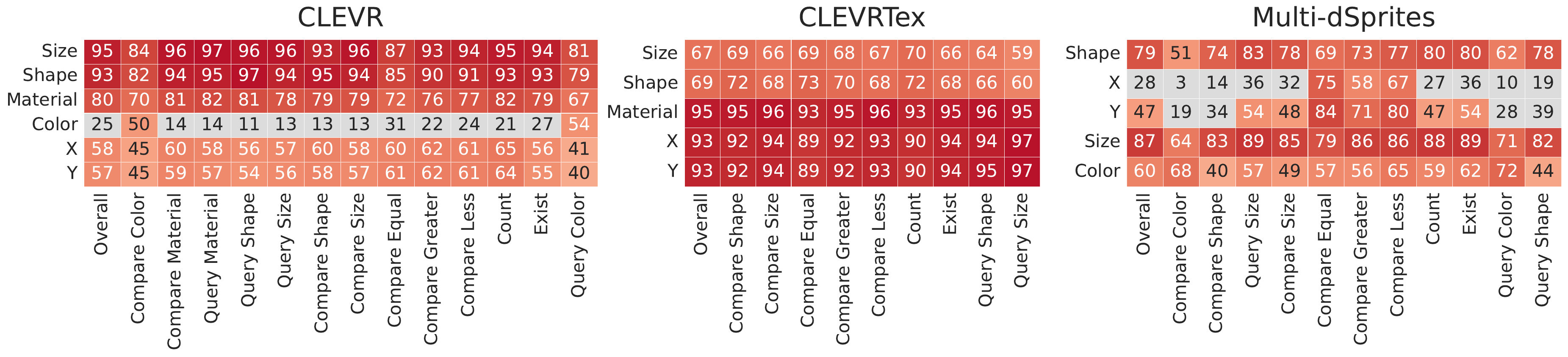}
    \caption{Spearman's rank correlation between downstream property prediction performance and downstream VQA performance of the models. The correlations are color-coded only when $p < 0.05$.}
    \label{fig:correlation_downstream}
\end{figure}

\cref{fig:correlation_downstream} shows Spearman's rank correlation between downstream property prediction performance of the models for each property, and downstream VQA performance for each question category. Overall, we observe a strong correlation between the performance of the models in two downstream tasks. It's important to note that training for property prediction takes significantly less time than VQA, by around $2$ orders of magnitude. Hence, property prediction can be a helpful guide when selecting a model for a downstream task.

\subsection{Upstream vs Downstream Performance}
\label{appendix:upstream_vs_downstream}
\begin{figure}[h]
    \centering
    \includegraphics[width=\textwidth]{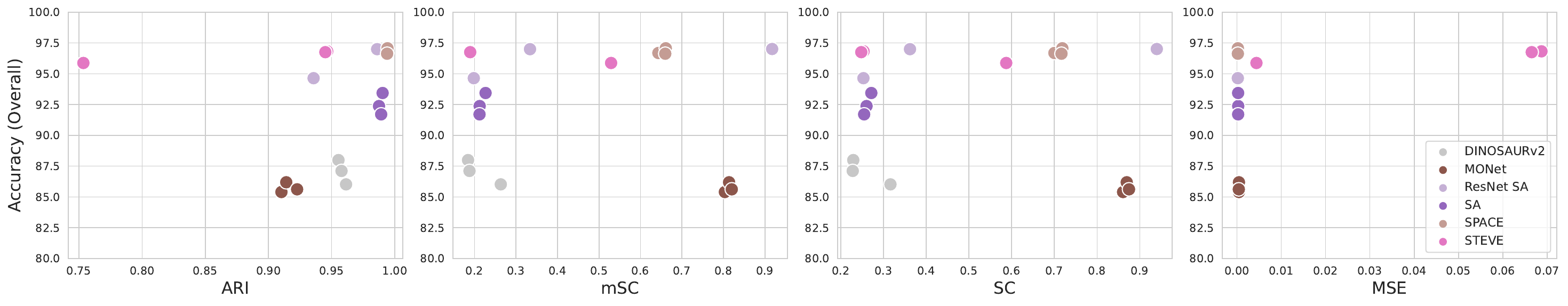}
    \caption{Upstream performance of object-centric models against the overall VQA accuracy when using T-15 as the downstream model on CLEVR.}
    \label{fig:correlation_scatter_clevr}
\end{figure}

\begin{figure}[h]
    \centering
    \includegraphics[width=\textwidth]{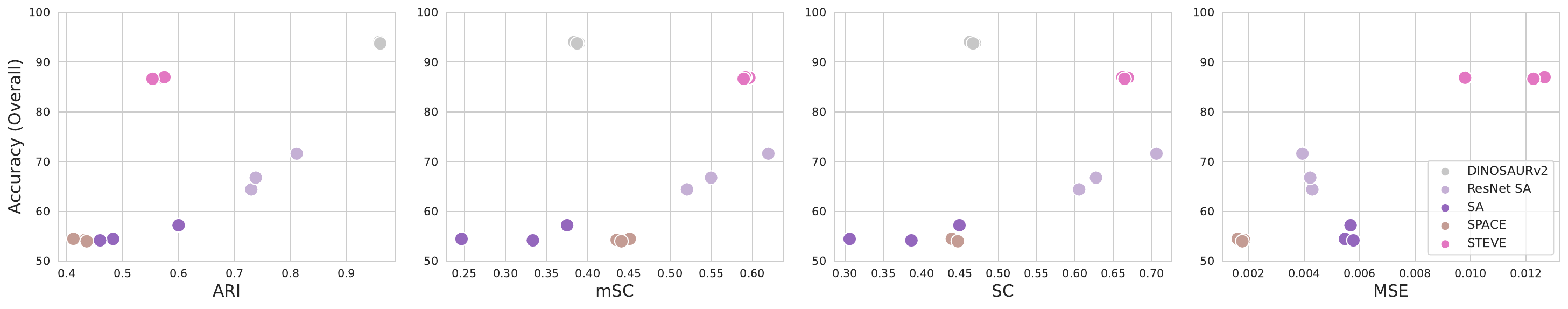}
    \caption{Upstream performance of object-centric models against the overall VQA accuracy when using T-15 as the downstream model on CLEVRTex.}
    \label{fig:correlation_scatter_clevrtex}
\end{figure}

\begin{figure}[h]
    \centering
    \includegraphics[width=\textwidth]{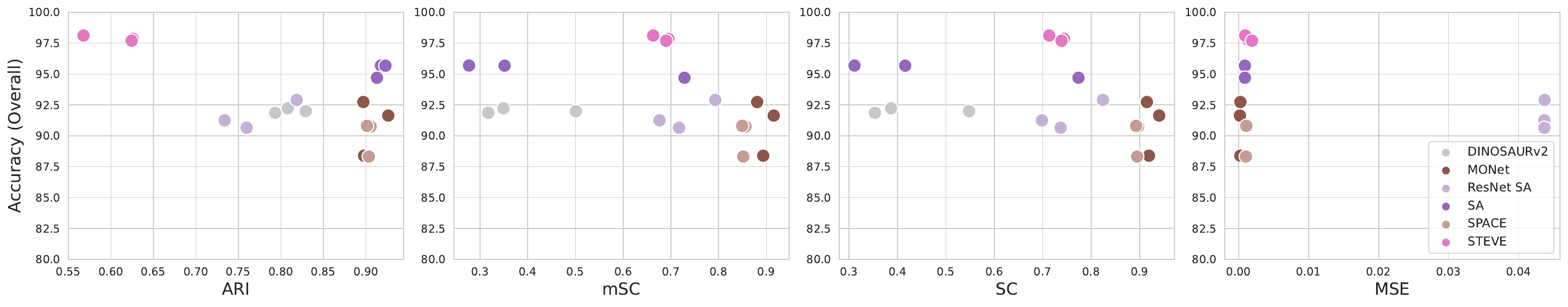}
    \caption{Upstream performance of object-centric models against the overall VQA accuracy when using T-15 as the downstream model on Multi-dSprites.}
    \label{fig:correlation_scatter_multi}
\end{figure}

Here we analyze the relationship between the upstream and downstream VQA performance of OC models. We exclude VQA-v2 and GQA results due to only having three well-performing OC models for this dataset.
\cref{fig:correlation_scatter_clevr,fig:correlation_scatter_clevrtex,fig:correlation_scatter_multi} depict the upstream performance of object-centric models in comparison to their downstream performance when using T-15 as the downstream model. Generally, there is no strong correlation between the two performances, and several outliers are observed in each plot.

More specifically, on CLEVR and Multi-dSprites, only ARI on CLEVR shows a slight correlation with VQA accuracy. All other upstream metrics do not correlate with the downstream performance. On CLEVRTex, however, we observe a weak correlation between ARI, mSC, and SC with VQA accuracy. At the same time, contrary to expectations, we observe a positive correlation between MSE and overall VQA accuracy which suggests that models with higher MSE values tend to perform better on the downstream task. These require further investigation which is beyond the scope of this work.

Additionally, when we look at the outliers, STEVE is the main one among OC models in all datasets, showing higher MSE and lower ARI, but better downstream performance. Also, one seed of STEVE tends to perform poorly in upstream metrics but achieves a downstream performance comparable to other seeds. Moreover, DINOSAURv2 consistently performs poorly on mSC and SC across all datasets despite performing well downstream. Additionally, SPACE typically has the best MSE but the worst downstream performance.

In conclusion, we observe that upstream metrics are not a good indicator of the downstream performance of different models and thus, are not reliable for upstream model selection.

It is noteworthy that DINOSAURv2 does not reconstruct the input but instead reconstructs the latent features of the input. Thus, it is not included in the plots showing reconstruction MSE. Furthermore, we also calculated the Spearman rank correlations between upstream and downstream metrics but due to the high p-value of most of the correlations, we chose not to report them.

\clearpage

\subsection{Effect of Downstream Model Size}
\label{app: effect of downstream model size}
\begin{figure}[h]
    \centering
    \includegraphics[width=\textwidth]{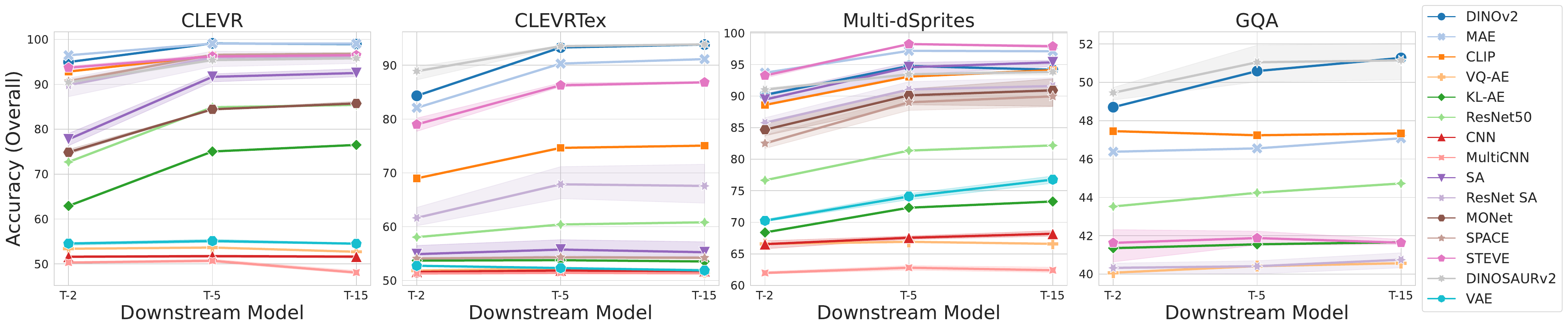}
    \caption{Average accuracies of different models w.r.t. downstream model size across different datasets. For pre-trained models, only one seed is available. For other models, the results are averaged over 3 random seeds and the shaded areas indicate 95\% confidence intervals.}
    \label{fig:downstream_model_size}
    % \vskip -0.1in
\end{figure}

\cref{fig:downstream_model_size}
depicts the overall accuracy of the models across different datasets with a downstream model having a varying number of layers. As expected, the downstream performance increases with the downstream model size. Furthermore, we observe that DINOSAURv2 performs better than DINOv2 on CLEVRTex, Multi-dSprites, and GQA when using T-2 as the downstream model. However, as the downstream model size increases, DINOv2's performance matches DINOSAURv2. This indicates that while DINOv2 representations do have the necessary information for object-related tasks, the information is less obvious and needs a larger downstream model to be effectively extracted. However, on CLEVR, we don’t see the same pattern, as DINOSAURv2 performs slightly worse than DINOv2. After experimenting with various hyperparameter sets, we found the results of DINOSAURv2 on CLEVR to be highly sensitive to the selected hyperparameters, which likely explains its suboptimal performance.

\subsection{Effect of Training Size}
\label{appendix:training_size}
\begin{figure}[h]
    \centering
    \includegraphics[width=\textwidth]{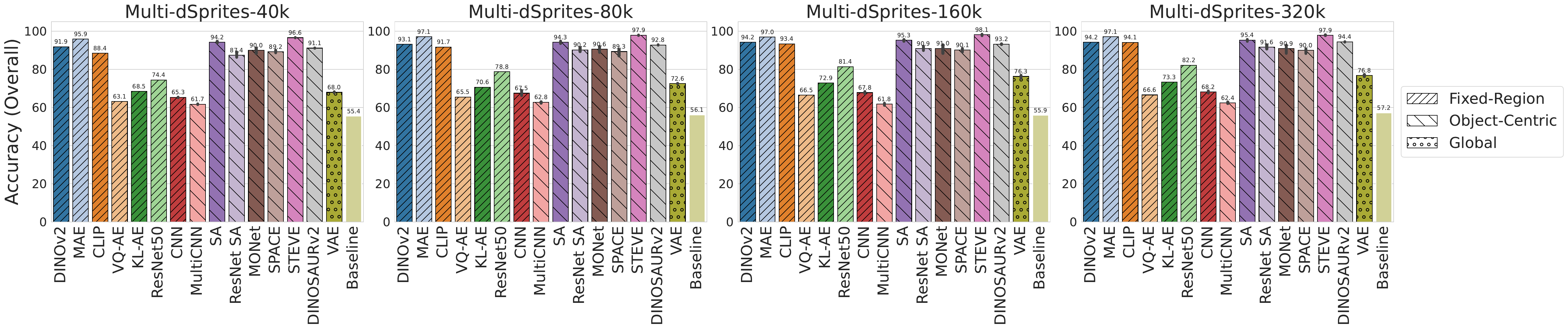}
    \caption{Average accuracies on the VQA downstream task for different models on Multi-dSprites with different training sizes when using T-15 as the downstream model. The bars indicate means and 95\% confidence intervals with 3 random seeds.}
    \label{fig:multi_overall_bar_transformer15}
\end{figure}

\begin{figure}[h]
    \centering
    \includegraphics[width=\textwidth]{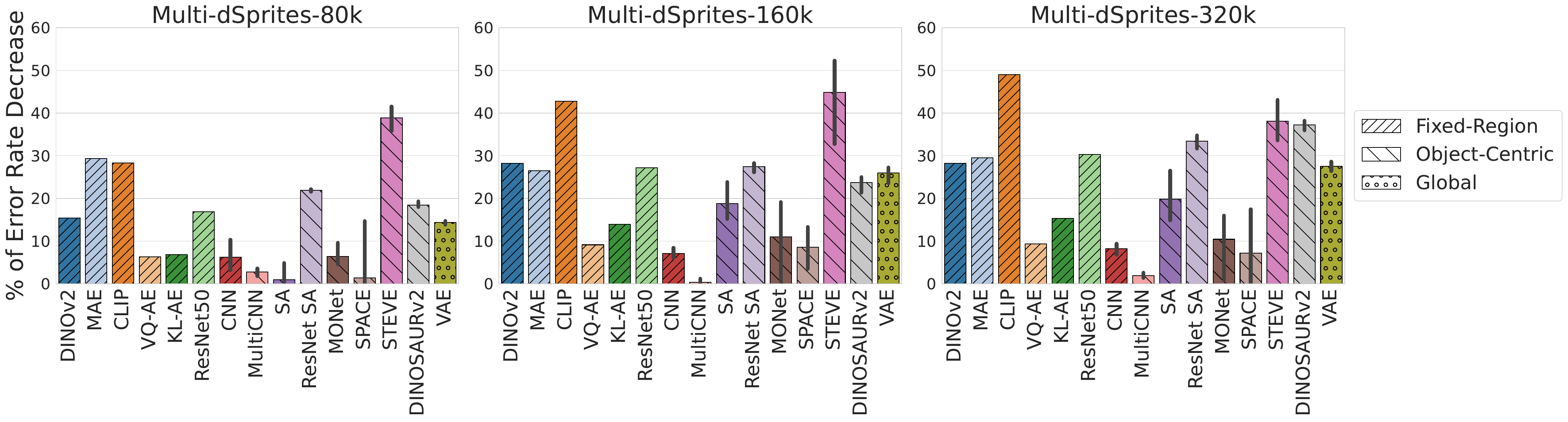}
    \caption{Average \% decrease in VQA error rate for different upstream models on Multi-dSprites, when increasing the training size from $40$k to larger sizes, using T-15 as the downstream model. The bars indicate means and 95\% confidence intervals with 3 random seeds.}
    \label{fig:multi_delta_overall_bar_transformer}
\end{figure}

\cref{fig:multi_overall_bar_transformer15} depicts the overall accuracy of different models on Multi-dSprites with varying training sizes of $40$k, $80$k, $160$k, and $320$k unique images, and \cref{fig:multi_delta_overall_bar_transformer} shows the average percentage of decrease in error rate when increasing the dataset size from $40$k to the respective size. With increased data, most upstream models show similar performance improvement across different training sizes regardless of their initial performance. CLIP is the only exception, showing a decrease in overall error rate of up to 50\%. Furthermore, the performance of the end-to-end CNN and MultiCNN models show minimal improvement compared to the other models.

% \newpage

\subsection{Consistency of the Results Across Different Question Types}
\label{appendix:consistency different question types}
\begin{figure}[h]
    \centering
    \includegraphics[width=\textwidth]{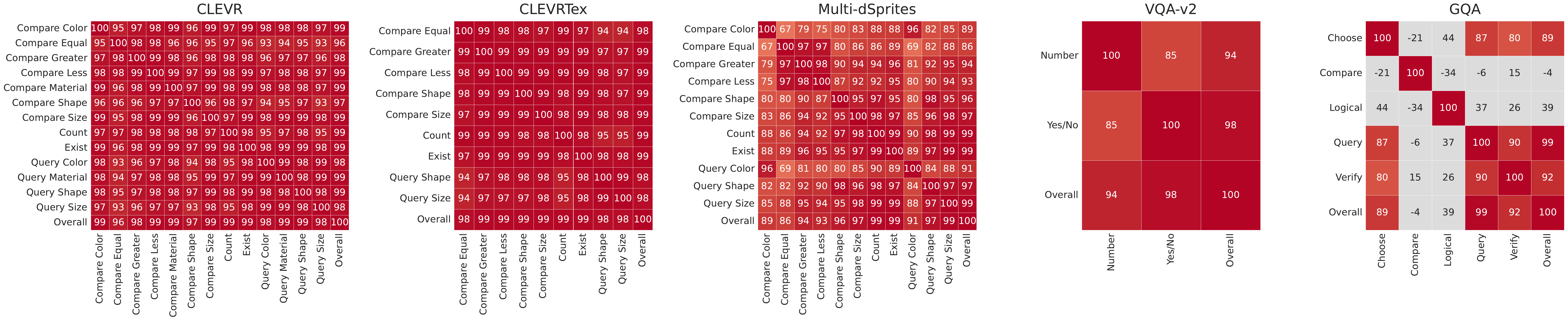}
    \caption{Spearman's rank correlation of model performances for each question category using T-15 as the downstream model.}
    \label{fig:correlation_question_type}
\end{figure}

\cref{fig:correlation_question_type} shows the Spearman's rank correlation between the performances of models on each question category when using T-15 for synthetic datasets and GQA, and T-2 for VQA-v2. Additionally, \cref{fig:Exist_bar_transformer15,fig:Count_bar_transformer15,fig:Compare Less_bar_transformer15,fig:Compare Greater_bar_transformer15,fig:Compare Equal_bar_transformer15,fig:Compare Shape_bar_transformer15,fig:Compare Size_bar_transformer15,fig:Query Shape_bar_transformer15,fig:Query Size_bar_transformer15,fig:cocovqav2_question_types_bar_transformer2,fig:gqa_question_types_bar_transformer15} illustrate the accuracy of different question types of all upstream model representations across different datasets. Except for GQA, we observe strong correlations between the performance of different question categories which indicates that the trend in the overall accuracy in \cref{fig:overall_bar_transformer15} matches the trend in the accuracy of each question category and the results are consistent across different question categories.

%%%% 

\begin{figure}[h]
    \centering
    \includegraphics[width=\textwidth]{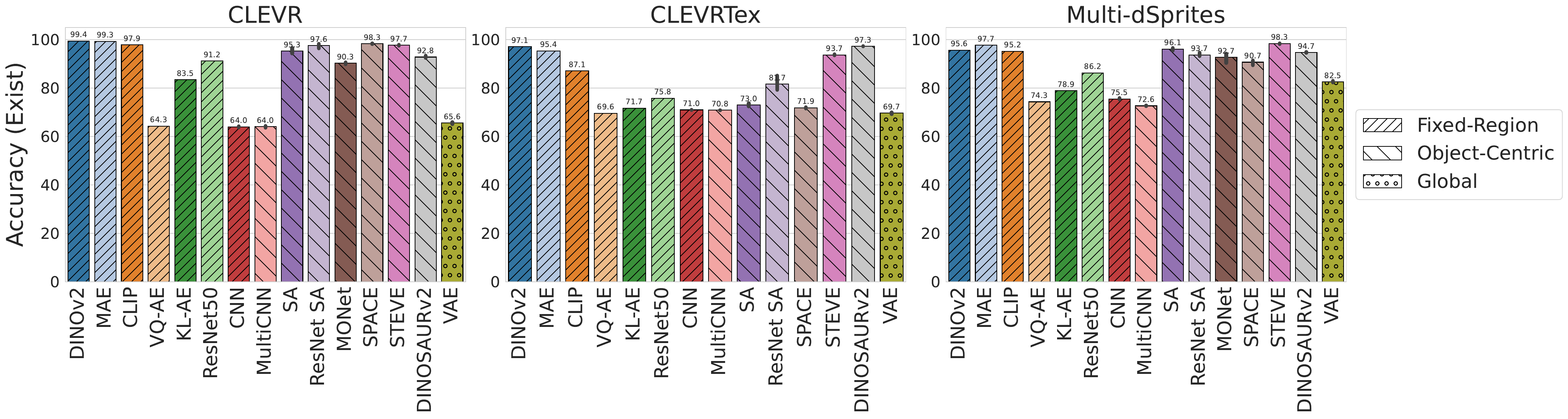}
    \caption{Average accuracies of \textit{Exist} questions for different upstream representation models when using T-15 as the downstream model. The bars indicate means and 95\% confidence intervals with 3 random seeds.}
    \label{fig:Exist_bar_transformer15}
\end{figure}

\begin{figure}[h]
    \centering
    \includegraphics[width=\textwidth]{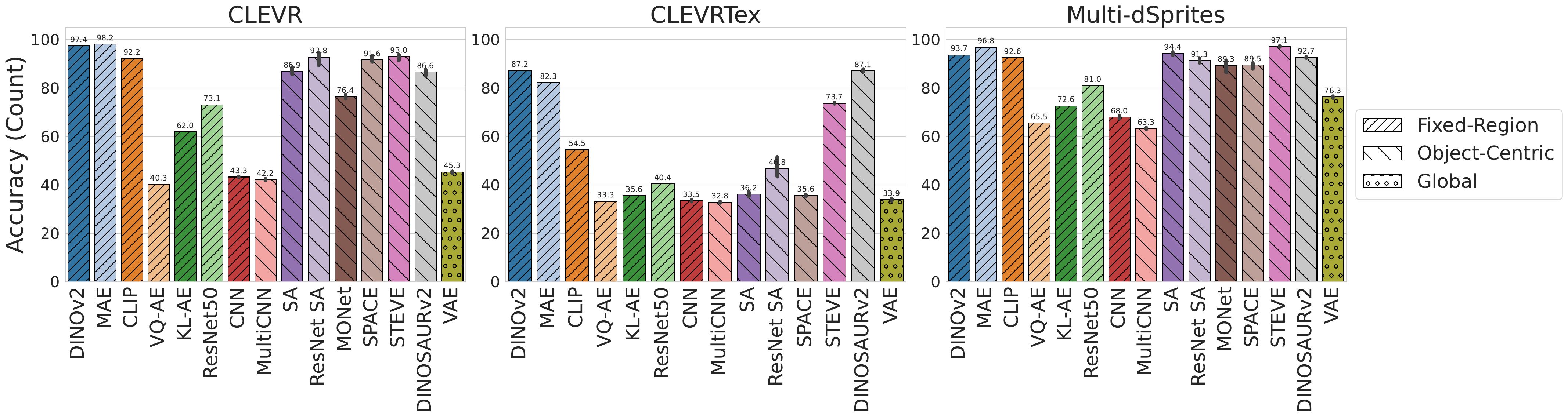}
    \caption{Average accuracies of \textit{Count} questions for different upstream representation models when using T-15 as the downstream model. The bars indicate means and 95\% confidence intervals with 3 random seeds.}
    \label{fig:Count_bar_transformer15}
\end{figure}

\begin{figure}[h]
    \centering
    \includegraphics[width=\textwidth]{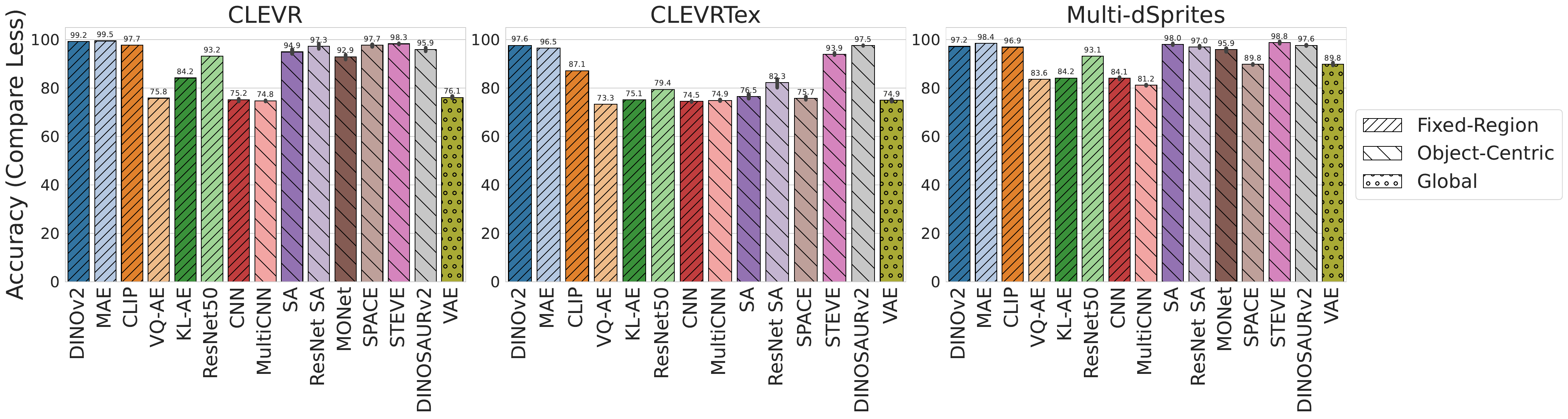}
    \caption{Average accuracies of \textit{Compare Integer (Less)} questions for different upstream representation models when using T-15 as the downstream model. The bars indicate means and 95\% confidence intervals with 3 random seeds.}
    \label{fig:Compare Less_bar_transformer15}
\end{figure}

\begin{figure}[h]
    \centering
    \includegraphics[width=\textwidth]{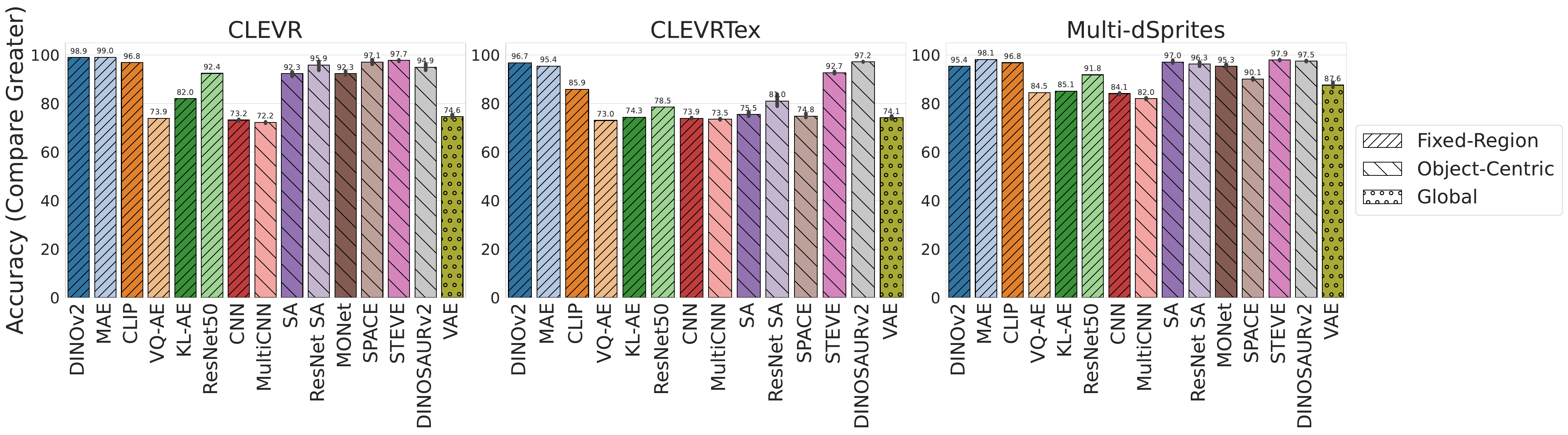}
    \caption{Average accuracies of \textit{Compare Integer (Greater)} questions for different upstream representation models when using T-15 as the downstream model. The bars indicate means and 95\% confidence intervals with 3 random seeds.}
    \label{fig:Compare Greater_bar_transformer15}
\end{figure}

\begin{figure}[h]
    \centering
    \includegraphics[width=\textwidth]{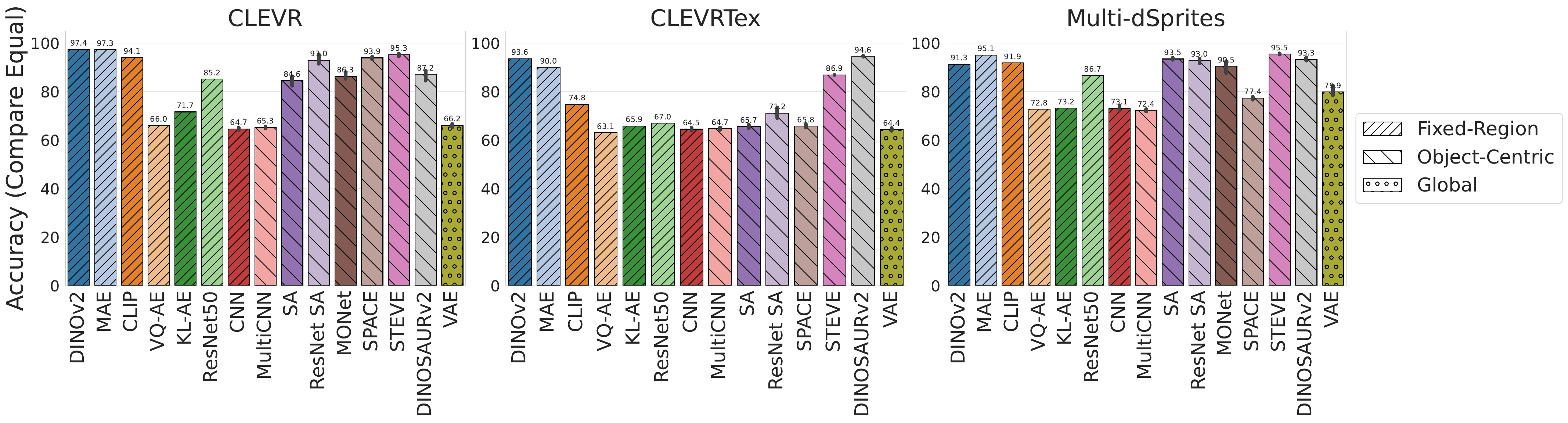}
    \caption{Average accuracies of \textit{Compare Integer (Equal)} questions for different upstream representation models when using T-15 as the downstream model. The bars indicate means and 95\% confidence intervals with 3 random seeds.}
    \label{fig:Compare Equal_bar_transformer15}
\end{figure}

\begin{figure}[h]
    \centering
    \includegraphics[width=\textwidth]{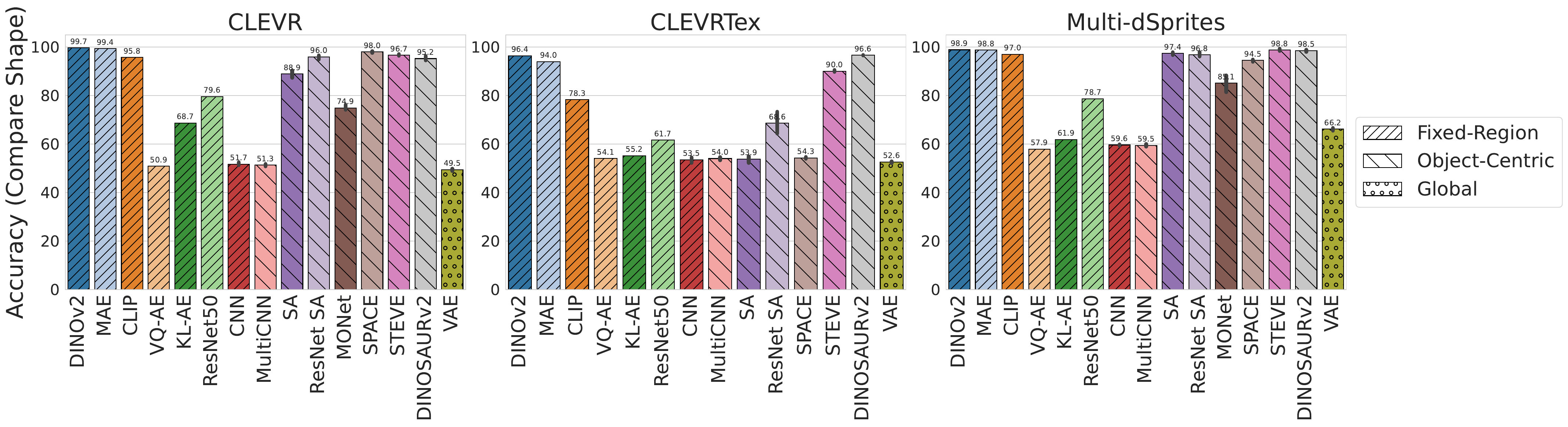}
    \caption{Average accuracies of \textit{Compare Attribute (Shape)} questions for different upstream representation models when using T-15 as the downstream model. The bars indicate means and 95\% confidence intervals with 3 random seeds.}
    \label{fig:Compare Shape_bar_transformer15}
\end{figure}

\begin{figure}[h]
    \centering
    \includegraphics[width=\textwidth]{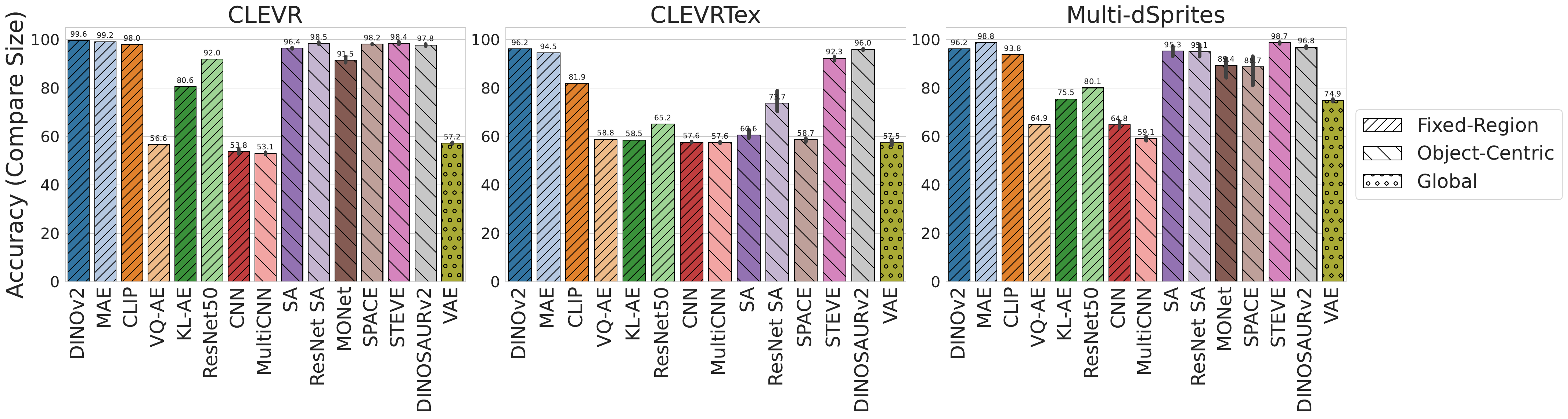}
    \caption{Average accuracies of \textit{Compare Attribute (Size)} questions for different upstream representation models when using T-15 as the downstream model. The bars indicate means and 95\% confidence intervals with 3 random seeds.}
    \label{fig:Compare Size_bar_transformer15}
\end{figure}

\begin{figure}[h]
    \centering
    \includegraphics[width=\textwidth]{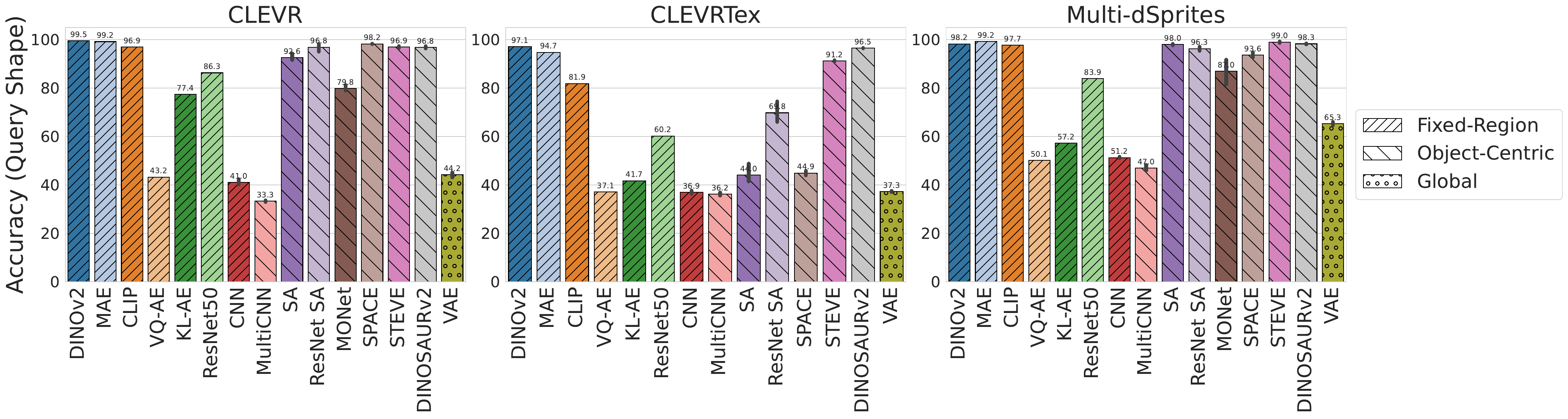}
    \caption{Average accuracies of \textit{Query Attribute (Shape)} questions for different upstream representation models when using T-15 as the downstream model. The bars indicate means and 95\% confidence intervals with 3 random seeds.}
    \label{fig:Query Shape_bar_transformer15}
\end{figure}

\begin{figure}[h]
    \centering
    \includegraphics[width=\textwidth]{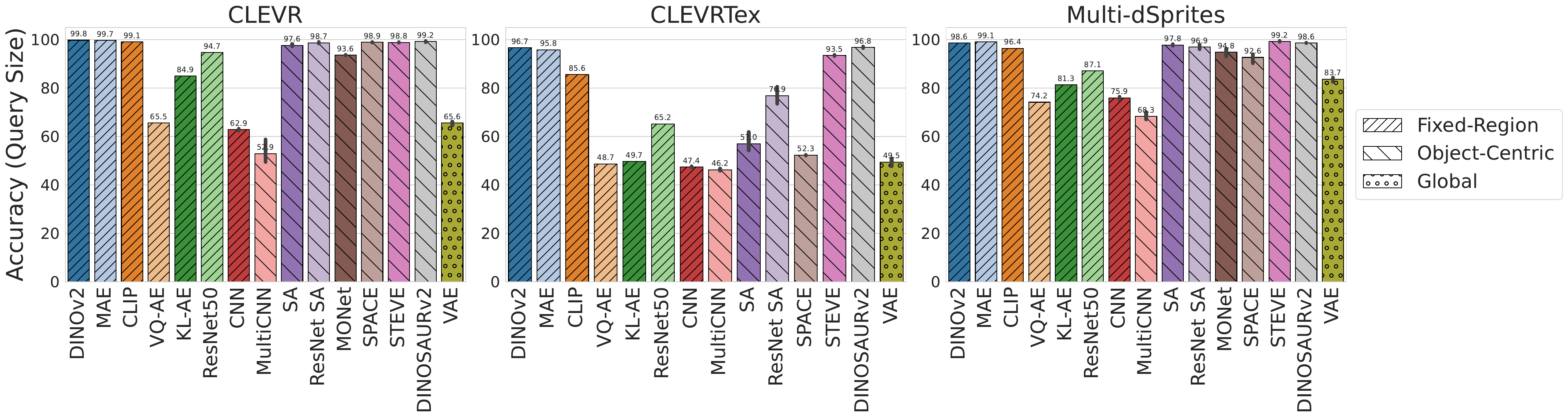}
    \caption{Average accuracies of \textit{Query Attribute (Size)} questions for different upstream representation models when using T-15 as the downstream model. The bars indicate means and 95\% confidence intervals with 3 random seeds.}
    \label{fig:Query Size_bar_transformer15}
\end{figure}

\begin{figure}[h]
    \centering
    \includegraphics[width=\textwidth]{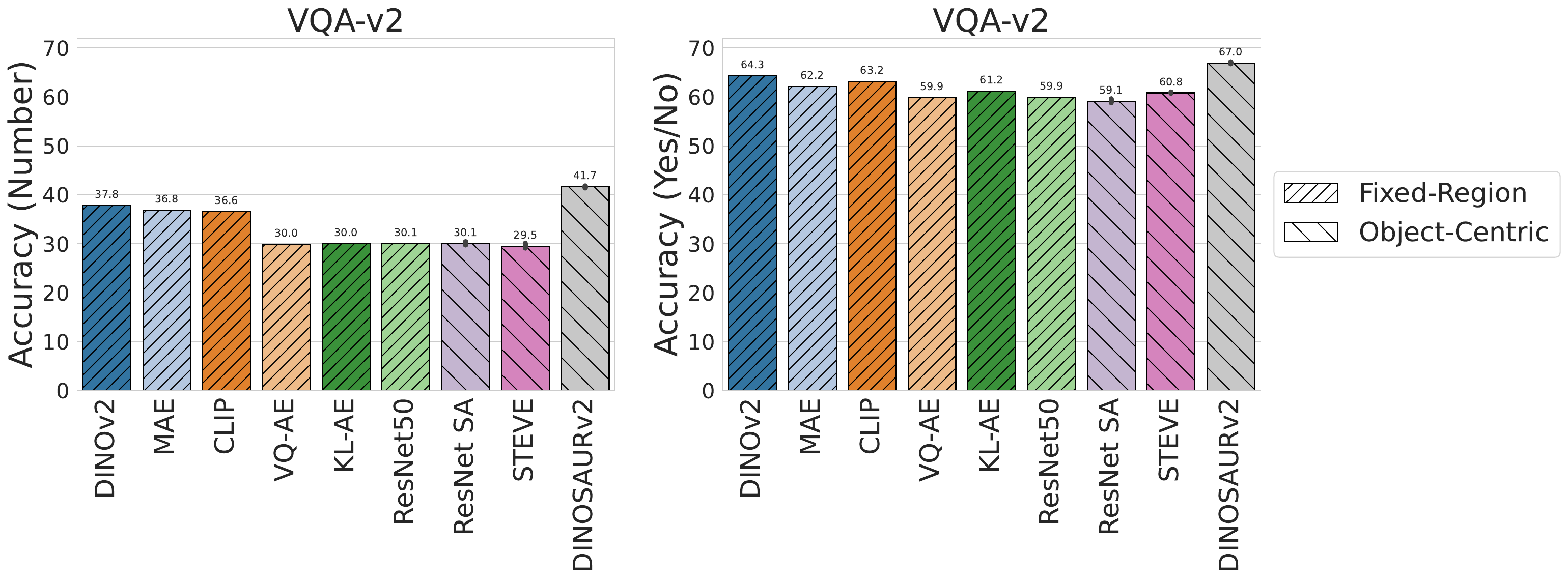}
    \caption{Average accuracies of different question types for different upstream representation models when using T-2 as the downstream model on VQA-v2. The bars indicate means and 95\% confidence intervals with 3 random seeds.}
    \label{fig:cocovqav2_question_types_bar_transformer2}
\end{figure}

\begin{figure}[h]
    \centering
    \includegraphics[width=\textwidth]{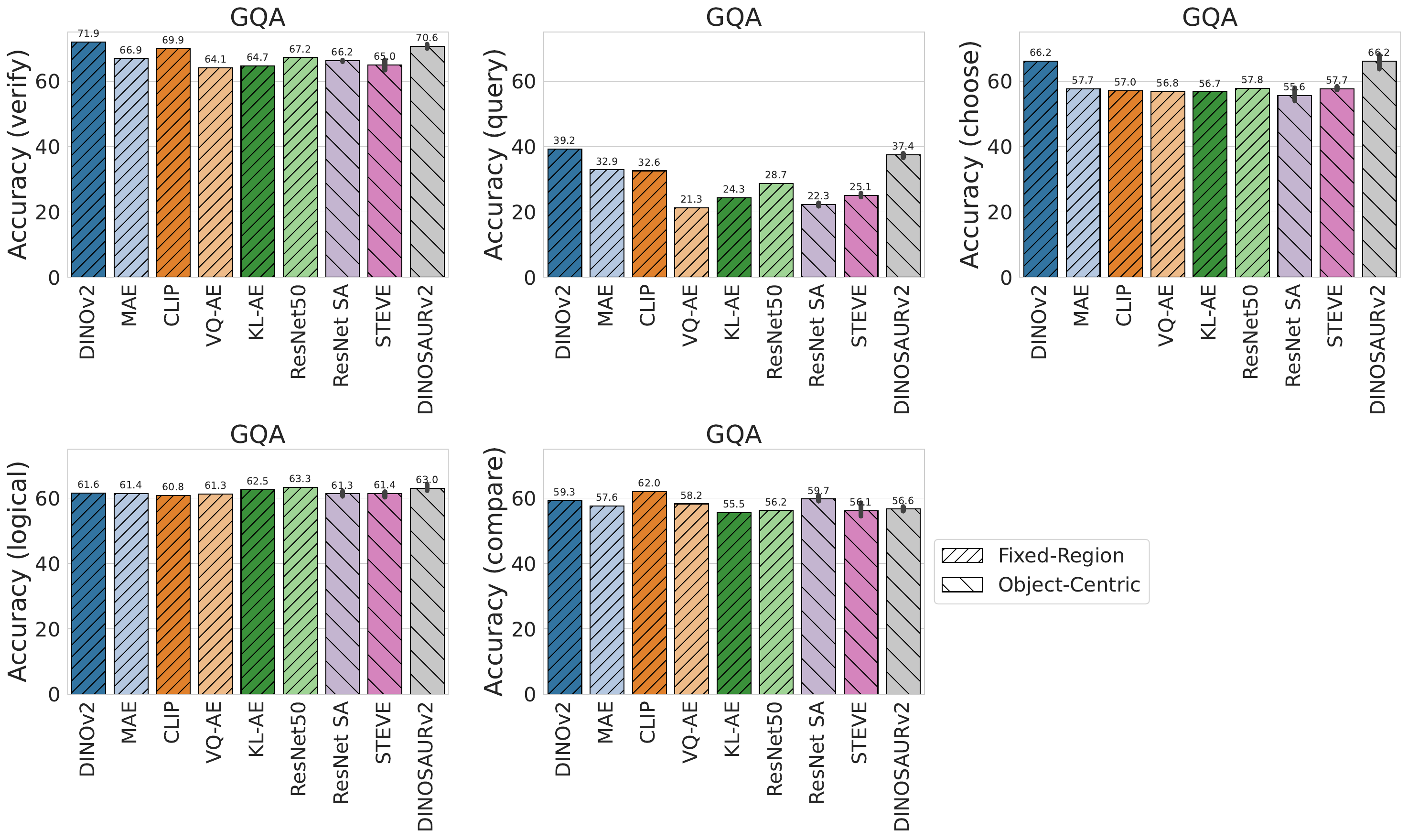}
    \caption{Average accuracies of different question types for different upstream representation models when using T-15 as the downstream model on GQA. The bars indicate means and 95\% confidence intervals with 3 random seeds.}
    \label{fig:gqa_question_types_bar_transformer15}
\end{figure}

%%%%

\clearpage

\subsection{Full Results}
The complete VQA accuracies of different upstream models across all datasets are presented in \cref{tab:acc_clevr,tab:acc_clevrtex,tab:acc_multi40,tab:acc_multi80,tab:acc_multi160,tab:acc_multi320,tab:acc_vqav2,tab:acc_gqa}.

\begin{table*}[h]
  \centering
  \caption{Average accuracies on CLEVR when using T-15 as the downstream model. For pre-trained models, only one seed is available. For other models, the results are aggregated over 3 random seeds.}
  \label{tab:acc_clevr}
  \begin{scriptsize}
  \setlength{\tabcolsep}{3pt}
  \begin{tabular}{lc|ccccccccccccc}
    \toprule
    \textbf{Model} & \textbf{Overall} & \textbf{Exist} & \textbf{Count} & \multicolumn{3}{c}{\textbf{Compare Integer}} & \multicolumn{4}{c}{\textbf{Compare Attribute}} & \multicolumn{4}{c}{\textbf{Query Attribute}} \\
    \cmidrule(lr){5-7} \cmidrule(lr){8-11} \cmidrule(lr){12-15}
    & & & & \textbf{Less} & \textbf{Greater} & \textbf{Equal} & \textbf{Shape} & \textbf{Color} & \textbf{Material} & \textbf{Size} & \textbf{Shape} & \textbf{Color} & \textbf{Material} & \textbf{Size} \\
    \midrule
    DINOv2 & 98.9 & 99.4 & 97.4 & 99.2 & 98.9 & 97.4 & 99.7 & 100.0 & 97.4 & 99.6 & 99.5 & 99.7 & 99.3 & 99.8 \\
    MAE & 99.0 & 99.3 & 98.2 & 99.5 & 99.0 & 97.3 & 99.4 & 99.9 & 98.2 & 99.2 & 99.2 & 99.6 & 99.5 & 99.7 \\
    CLIP & 96.5 & 97.9 & 92.2 & 97.7 & 96.8 & 94.1 & 95.8 & 98.6 & 96.2 & 98.0 & 96.9 & 98.2 & 98.1 & 99.1 \\
    VQ-AE & 52.7 & 64.3 & 40.3 & 75.8 & 73.9 & 66.0 & 50.9 & 52.1 & 49.9 & 56.6 & 43.2 & 33.2 & 58.5 & 65.5 \\
    KL-AE & 76.5 & 83.5 & 62.0 & 84.2 & 82.0 & 71.7 & 68.7 & 80.0 & 75.6 & 80.6 & 77.4 & 77.4 & 84.6 & 84.9 \\
    ResNet50 & 85.4 & 91.2 & 73.1 & 93.2 & 92.4 & 85.2 & 79.6 & 85.1 & 82.3 & 92.0 & 86.3 & 84.9 & 89.6 & 94.7 \\
    CNN & 51.6 & 64.0 & 43.3 & 75.2 & 73.2 & 64.7 & 51.7 & 53.2 & 51.3 & 53.8 & 41.0 & 21.1 & 59.0 & 62.9 \\
    \midrule
    MultiCNN & 48.1 & 64.0 & 42.2 & 74.8 & 72.2 & 65.3 & 51.3 & 52.7 & 51.4 & 53.1 & 33.3 & 12.6 & 50.3 & 52.9 \\
    SA & 92.5 & 95.3 & 86.9 & 94.9 & 92.3 & 84.6 & 88.9 & 93.7 & 92.7 & 96.4 & 92.6 & 93.5 & 95.8 & 97.6 \\
    ResNet SA & 96.2 & 97.6 & 92.8 & 97.3 & 95.9 & 93.0 & 96.0 & 96.4 & 96.3 & 98.5 & 96.8 & 96.4 & 98.0 & 98.7 \\
    MONet & 85.7 & 90.3 & 76.4 & 92.9 & 92.3 & 86.3 & 74.9 & 88.0 & 87.9 & 91.5 & 79.8 & 86.1 & 92.2 & 93.6 \\
    SPACE & 96.8 & 98.3 & 91.6 & 97.7 & 97.1 & 93.9 & 98.0 & 98.0 & 97.5 & 98.2 & 98.2 & 98.5 & 98.8 & 98.9 \\
    STEVE & 96.5 & 97.7 & 93.0 & 98.3 & 97.7 & 95.3 & 96.7 & 96.9 & 96.2 & 98.4 & 96.9 & 96.5 & 97.7 & 98.8 \\
    DINOSAURv2 & 94.3 & 92.8 & 86.6 & 95.9 & 94.9 & 87.2 & 95.2 & 76.3 & 95.1 & 97.8 & 96.8 & 63.0 & 97.7 & 99.2 \\
    \midrule
    VAE & 54.5 & 65.6 & 45.3 & 76.1 & 74.6 & 66.2 & 49.5 & 52.2 & 52.6 & 57.2 & 44.2 & 32.6 & 63.2 & 65.6 \\
    \bottomrule
  \end{tabular}
  \end{scriptsize}
\end{table*}

\begin{table*}[h]
  \centering
  \caption{Average accuracies on CLEVRTex when using T-15 as the downstream model. For pre-trained models, only one seed is available. For other models, the results are aggregated over 3 random seeds.}
  \label{tab:acc_clevrtex}
  \begin{scriptsize}
  \begin{tabular}{lc|ccccccrcr}
    \toprule
    \textbf{Model} & \textbf{Overall} & \textbf{Exist} & \textbf{Count} & \multicolumn{3}{c}{\textbf{Compare Integer}} & \multicolumn{2}{c}{\textbf{Compare Attribute}} & \multicolumn{2}{c}{\textbf{Query Attribute}} \\
    \cmidrule(lr){5-7} \cmidrule(lr){8-9} \cmidrule(lr){10-11}
    & & & & \textbf{Less} & \textbf{Greater} & \textbf{Equal} & \textbf{Shape} & \textbf{Size} & \textbf{Shape} & \textbf{Size} \\
    \midrule
    DINOv2 & 93.8 & 97.1 & 87.2 & 97.6 & 96.7 & 93.6 & 96.4 & 96.2 & 97.1 & 96.7 \\
    MAE & 91.1 & 95.4 & 82.3 & 96.5 & 95.4 & 90.0 & 94.0 & 94.5 & 94.7 & 95.8 \\
    CLIP & 75.1 & 87.1 & 54.5 & 87.1 & 85.9 & 74.8 & 78.3 & 81.9 & 81.9 & 85.6 \\
    VQ-AE & 51.5 & 69.6 & 33.3 & 73.3 & 73.0 & 63.1 & 54.1 & 58.8 & 37.1 & 48.7 \\
    KL-AE & 53.5 & 71.7 & 35.6 & 75.1 & 74.3 & 65.9 & 55.2 & 58.5 & 41.7 & 49.7 \\
    ResNet50 & 60.8 & 75.8 & 40.4 & 79.4 & 78.5 & 67.0 & 61.7 & 65.2 & 60.2 & 65.2 \\
    CNN & 51.7 & 71.0 & 33.5 & 74.5 & 73.9 & 64.5 & 53.5 & 57.6 & 36.9 & 47.4 \\
    \midrule
    MultiCNN & 51.3 & 70.8 & 32.8 & 74.9 & 73.5 & 64.7 & 54.0 & 57.6 & 36.2 & 46.2 \\
    SA & 55.3 & 73.0 & 36.2 & 76.5 & 75.5 & 65.7 & 53.9 & 60.6 & 44.0 & 57.0 \\
    ResNet SA & 67.6 & 81.7 & 46.8 & 82.3 & 81.0 & 71.2 & 68.6 & 73.7 & 69.8 & 76.9 \\
    SPACE & 54.2 & 71.9 & 35.6 & 75.7 & 74.8 & 65.8 & 54.3 & 58.7 & 44.9 & 52.3 \\
    STEVE & 86.8 & 93.7 & 73.7 & 93.9 & 92.7 & 86.9 & 90.0 & 92.3 & 91.2 & 93.5 \\
    DINOSAURv2 & 93.9 & 97.3 & 87.1 & 97.5 & 97.2 & 94.6 & 96.6 & 96.0 & 96.5 & 96.8 \\
    \midrule
    VAE & 51.9 & 69.7 & 33.9 & 74.9 & 74.1 & 64.4 & 52.6 & 57.5 & 37.3 & 49.5 \\
    \bottomrule
  \end{tabular}
  \end{scriptsize}
\end{table*}

\begin{table*}[h]
  \centering
  \caption{Average accuracies on Multi-dSprites with $40$k unique training images when using T-15 as the downstream model. For pre-trained models, only one seed is available. For other models, the results are aggregated over 3 random seeds.}
  \label{tab:acc_multi40}
  \begin{scriptsize}
  \begin{tabular}{lc|ccccccccccccccc}
    \toprule
    \textbf{Model} & \textbf{Overall} & \textbf{Exist} & \textbf{Count} & \multicolumn{3}{c}{\textbf{Compare Integer}} & \multicolumn{3}{c}{\textbf{Compare Attribute}} & \multicolumn{3}{c}{\textbf{Query Attribute}} \\
    \cmidrule(lr){5-7} \cmidrule(lr){8-10} \cmidrule(lr){11-13}
    & & & & \textbf{Less} & \textbf{Greater} & \textbf{Equal} & \textbf{Shape} & \textbf{Color} & \textbf{Size} & \textbf{Shape} & \textbf{Color} & \textbf{Size} \\
    \midrule
    DINOv2 & 91.9 & 94.1 & 91.3 & 96.4 & 96.1 & 91.7 & 96.9 & 80.8 & 94.1 & 97.6 & 77.4 & 96.1 \\
    MAE & 95.9 & 96.9 & 95.2 & 98.1 & 98.2 & 94.8 & 98.2 & 91.0 & 96.7 & 98.8 & 89.9 & 98.2 \\
    CLIP & 88.4 & 91.0 & 86.6 & 92.8 & 93.1 & 86.1 & 92.3 & 77.2 & 84.8 & 95.0 & 80.8 & 92.1 \\
    VQ-AE & 63.1 & 72.8 & 62.2 & 82.6 & 83.2 & 71.8 & 53.5 & 58.8 & 59.9 & 48.6 & 41.6 & 71.5 \\
    KL-AE & 68.5 & 75.4 & 67.9 & 83.2 & 84.2 & 72.1 & 56.9 & 66.4 & 65.0 & 54.2 & 56.8 & 76.5 \\
    ResNet50 & 74.4 & 79.9 & 73.7 & 87.8 & 87.3 & 78.8 & 73.4 & 60.9 & 70.0 & 77.4 & 53.5 & 79.8 \\
    CNN & 65.3 & 74.0 & 65.9 & 82.3 & 82.6 & 71.6 & 56.3 & 62.1 & 59.9 & 49.8 & 47.4 & 72.5 \\
    \midrule
    MultiCNN & 61.7 & 72.6 & 62.6 & 80.7 & 80.7 & 71.3 & 58.3 & 59.0 & 58.0 & 46.6 & 32.2 & 68.4 \\
    SA & 94.2 & 95.4 & 93.1 & 97.2 & 96.4 & 92.2 & 96.2 & 87.8 & 94.7 & 97.4 & 89.2 & 97.2 \\
    ResNet SA & 87.4 & 90.6 & 87.2 & 95.5 & 93.6 & 89.0 & 94.4 & 65.0 & 90.4 & 94.7 & 65.3 & 95.7 \\
    MONet & 90.0 & 92.4 & 88.2 & 95.6 & 93.9 & 88.5 & 84.1 & 88.6 & 88.0 & 86.4 & 89.8 & 94.6 \\
    SPACE & 89.2 & 90.3 & 88.8 & 89.3 & 89.2 & 76.3 & 93.5 & 85.9 & 89.8 & 93.2 & 85.8 & 92.0 \\
    STEVE & 96.6 & 97.5 & 95.5 & 98.0 & 97.1 & 93.5 & 97.6 & 92.6 & 97.9 & 98.5 & 95.1 & 98.8 \\
    DINOSAURv2 & 91.1 & 91.0 & 87.8 & 95.3 & 94.2 & 89.4 & 94.3 & 64.8 & 92.8 & 95.8 & 63.3 & 95.9 \\
    \midrule
    VAE & 68.0 & 74.5 & 68.9 & 83.8 & 83.3 & 71.0 & 59.3 & 60.9 & 68.7 & 57.7 & 47.7 & 75.6 \\
    \bottomrule
  \end{tabular}
  \end{scriptsize}
\end{table*}

\begin{table*}[h]
  \centering
  \caption{Average accuracies on Multi-dSprites with $80$k unique training images when using T-15 as the downstream model. For pre-trained models, only one seed is available. For other models, the results are aggregated over 3 random seeds.}
  \label{tab:acc_multi80}
  \begin{scriptsize}
  \begin{tabular}{lc|ccccccccccccccc}
    \toprule
    \textbf{Model} & \textbf{Overall} & \textbf{Exist} & \textbf{Count} & \multicolumn{3}{c}{\textbf{Compare Integer}} & \multicolumn{3}{c}{\textbf{Compare Attribute}} & \multicolumn{3}{c}{\textbf{Query Attribute}} \\
    \cmidrule(lr){5-7} \cmidrule(lr){8-10} \cmidrule(lr){11-13}
    & & & & \textbf{Less} & \textbf{Greater} & \textbf{Equal} & \textbf{Shape} & \textbf{Color} & \textbf{Size} & \textbf{Shape} & \textbf{Color} & \textbf{Size} \\
    \midrule
    DINOv2 & 93.1 & 95.3 & 92.9 & 97.0 & 96.0 & 91.6 & 97.7 & 81.2 & 94.7 & 98.2 & 81.1 & 97.4 \\
    MAE & 97.1 & 97.7 & 96.6 & 98.2 & 98.3 & 96.5 & 98.3 & 93.0 & 98.7 & 98.7 & 93.5 & 99.1 \\
    CLIP & 91.7 & 94.3 & 90.5 & 95.2 & 95.6 & 89.5 & 95.3 & 84.0 & 89.0 & 95.8 & 83.9 & 95.2 \\
    VQ-AE & 65.5 & 73.0 & 64.6 & 83.3 & 83.5 & 73.4 & 57.8 & 62.8 & 62.5 & 50.0 & 49.0 & 73.0 \\
    KL-AE & 70.6 & 76.9 & 67.5 & 84.3 & 85.1 & 70.8 & 60.8 & 70.7 & 70.1 & 57.1 & 63.6 & 80.3 \\
    ResNet50 & 78.8 & 82.7 & 77.6 & 90.6 & 89.4 & 83.5 & 76.5 & 66.3 & 74.5 & 81.7 & 62.5 & 85.0 \\
    CNN & 67.5 & 75.3 & 67.4 & 83.4 & 83.7 & 72.8 & 58.9 & 65.1 & 63.5 & 51.2 & 53.0 & 75.1 \\
    \midrule
    MultiCNN & 62.8 & 72.7 & 63.6 & 81.2 & 82.0 & 72.4 & 59.5 & 61.3 & 59.9 & 48.2 & 34.0 & 68.9 \\
    SA & 94.3 & 95.4 & 93.6 & 97.8 & 97.0 & 93.4 & 97.1 & 89.0 & 91.8 & 97.7 & 88.2 & 96.0 \\
    ResNet SA & 90.2 & 93.1 & 89.7 & 96.0 & 95.4 & 91.0 & 96.0 & 71.9 & 93.3 & 95.5 & 73.5 & 96.3 \\
    MONet & 90.6 & 92.3 & 88.6 & 95.8 & 94.5 & 91.0 & 84.9 & 90.9 & 89.6 & 86.8 & 91.5 & 94.7 \\
    SPACE & 89.3 & 90.1 & 88.3 & 89.6 & 89.2 & 77.3 & 93.9 & 89.1 & 87.4 & 93.3 & 87.6 & 91.9 \\
    STEVE & 97.9 & 98.4 & 97.3 & 98.9 & 98.3 & 95.8 & 99.3 & 95.8 & 99.0 & 99.1 & 96.4 & 99.3 \\
    DINOSAURv2 & 92.8 & 93.6 & 90.6 & 96.1 & 96.0 & 91.7 & 97.0 & 74.5 & 95.0 & 96.9 & 75.3 & 97.2 \\
    \midrule
    VAE & 72.6 & 78.2 & 72.2 & 85.0 & 85.3 & 73.4 & 65.5 & 67.7 & 71.3 & 62.7 & 60.6 & 79.5 \\
    \bottomrule
  \end{tabular}
  \end{scriptsize}
\end{table*}

\begin{table*}[h]
  \centering
  \caption{Average accuracies on Multi-dSprites with $160$k unique training images when using T-15 as the downstream model. For pre-trained models, only one seed is available. For other models, the results are aggregated over 3 random seeds.}
  \label{tab:acc_multi160}
  \begin{scriptsize}
  \begin{tabular}{lc|ccccccccccccccc}
    \toprule
    \textbf{Model} & \textbf{Overall} & \textbf{Exist} & \textbf{Count} & \multicolumn{3}{c}{\textbf{Compare Integer}} & \multicolumn{3}{c}{\textbf{Compare Attribute}} & \multicolumn{3}{c}{\textbf{Query Attribute}} \\
    \cmidrule(lr){5-7} \cmidrule(lr){8-10} \cmidrule(lr){11-13}
    & & & & \textbf{Less} & \textbf{Greater} & \textbf{Equal} & \textbf{Shape} & \textbf{Color} & \textbf{Size} & \textbf{Shape} & \textbf{Color} & \textbf{Size} \\
    \midrule
    DINOv2 & 94.2 & 95.6 & 93.7 & 97.2 & 95.4 & 91.3 & 98.9 & 84.9 & 96.2 & 98.2 & 84.8 & 98.6 \\
    MAE & 97.0 & 97.6 & 96.7 & 98.4 & 98.8 & 96.0 & 98.3 & 92.7 & 98.7 & 99.2 & 91.9 & 98.9 \\
    CLIP & 93.4 & 95.5 & 91.6 & 96.6 & 96.5 & 91.1 & 96.7 & 90.5 & 91.5 & 97.1 & 86.9 & 96.0 \\
    VQ-AE & 66.5 & 73.2 & 65.6 & 82.5 & 84.2 & 72.5 & 58.1 & 66.6 & 64.8 & 50.1 & 52.5 & 74.2 \\
    KL-AE & 72.9 & 78.4 & 71.0 & 84.8 & 85.5 & 74.0 & 61.5 & 72.8 & 72.9 & 58.2 & 67.3 & 81.7 \\
    ResNet50 & 81.4 & 86.0 & 80.0 & 92.8 & 90.8 & 85.3 & 78.2 & 70.3 & 76.4 & 84.2 & 66.8 & 87.1 \\
    CNN & 67.8 & 75.4 & 67.4 & 83.8 & 83.8 & 72.8 & 60.0 & 65.3 & 62.7 & 51.2 & 54.8 & 75.8 \\
    \midrule
    MultiCNN & 61.8 & 72.6 & 63.1 & 81.2 & 81.7 & 72.5 & 59.0 & 61.0 & 58.9 & 42.8 & 33.2 & 67.9 \\
    SA & 95.3 & 96.0 & 94.3 & 98.2 & 97.8 & 94.7 & 97.5 & 91.0 & 94.5 & 98.0 & 91.0 & 97.2 \\
    ResNet SA & 90.9 & 93.1 & 90.6 & 96.7 & 96.0 & 91.3 & 95.8 & 75.6 & 92.9 & 96.0 & 75.6 & 96.2 \\
    MONet & 91.0 & 92.8 & 89.3 & 95.9 & 95.3 & 91.3 & 85.6 & 91.4 & 88.6 & 87.4 & 91.3 & 94.7 \\
    SPACE & 90.1 & 90.6 & 89.6 & 89.6 & 90.0 & 77.6 & 94.8 & 89.8 & 89.0 & 93.8 & 88.0 & 92.6 \\
    STEVE & 98.1 & 98.4 & 97.7 & 98.7 & 98.6 & 95.9 & 98.8 & 96.1 & 99.0 & 99.3 & 96.8 & 99.4 \\
    DINOSAURv2 & 93.2 & 93.3 & 90.4 & 96.2 & 95.7 & 91.5 & 96.5 & 73.2 & 95.2 & 96.8 & 76.0 & 97.1 \\
    \midrule
    VAE & 76.3 & 81.9 & 75.6 & 89.0 & 87.8 & 79.4 & 68.0 & 70.9 & 74.6 & 65.3 & 67.2 & 83.2 \\
    \bottomrule
  \end{tabular}
  \end{scriptsize}
\end{table*}

\begin{table*}[h]
  \centering
  \caption{Average accuracies on Multi-dSprites with $320$k unique training images when using T-15 as the downstream model. For pre-trained models, only one seed is available. For other models, the results are aggregated over 3 random seeds.}
  \label{tab:acc_multi320}
  \begin{scriptsize}
  \begin{tabular}{lc|ccccccccccccccc}
    \toprule
    \textbf{Model} & \textbf{Overall} & \textbf{Exist} & \textbf{Count} & \multicolumn{3}{c}{\textbf{Compare Integer}} & \multicolumn{3}{c}{\textbf{Compare Attribute}} & \multicolumn{3}{c}{\textbf{Query Attribute}} \\
    \cmidrule(lr){5-7} \cmidrule(lr){8-10} \cmidrule(lr){11-13}
    & & & & \textbf{Less} & \textbf{Greater} & \textbf{Equal} & \textbf{Shape} & \textbf{Color} & \textbf{Size} & \textbf{Shape} & \textbf{Color} & \textbf{Size} \\
    \midrule
    DINOv2 & 94.2 & 95.6 & 93.7 & 97.2 & 95.4 & 91.3 & 98.9 & 84.9 & 96.2 & 98.2 & 84.8 & 98.6 \\
    MAE & 97.1 & 97.7 & 96.8 & 98.4 & 98.1 & 95.1 & 98.8 & 92.5 & 98.8 & 99.2 & 92.9 & 99.1 \\
    CLIP & 94.1 & 95.2 & 92.6 & 96.9 & 96.8 & 91.9 & 97.0 & 90.3 & 93.8 & 97.7 & 89.3 & 96.4 \\
    VQ-AE & 66.6 & 74.3 & 65.5 & 83.6 & 84.5 & 72.8 & 57.9 & 64.0 & 64.9 & 50.1 & 51.9 & 74.2 \\
    KL-AE & 73.3 & 78.9 & 72.6 & 84.2 & 85.1 & 73.2 & 61.9 & 72.9 & 75.5 & 57.2 & 67.4 & 81.3 \\
    ResNet50 & 82.2 & 86.2 & 81.0 & 93.1 & 91.8 & 86.7 & 78.7 & 71.1 & 80.1 & 83.9 & 68.0 & 87.1 \\
    CNN & 68.2 & 75.5 & 68.0 & 84.1 & 84.1 & 73.1 & 59.6 & 66.2 & 64.8 & 51.2 & 55.0 & 75.9 \\
    \midrule
    MultiCNN & 62.4 & 72.6 & 63.3 & 81.2 & 82.0 & 72.4 & 59.5 & 61.0 & 59.1 & 47.0 & 33.7 & 68.3 \\
    SA & 95.4 & 96.1 & 94.4 & 98.0 & 97.0 & 93.5 & 97.4 & 90.4 & 95.3 & 98.0 & 91.2 & 97.8 \\
    ResNet SA & 91.6 & 93.7 & 91.3 & 97.0 & 96.3 & 93.0 & 96.8 & 78.4 & 95.1 & 96.3 & 75.5 & 96.9 \\
    MONet & 90.9 & 92.7 & 89.3 & 95.9 & 95.3 & 90.5 & 85.1 & 91.4 & 89.4 & 87.0 & 91.4 & 94.8 \\
    SPACE & 90.0 & 90.7 & 89.5 & 89.8 & 90.1 & 77.4 & 94.5 & 88.7 & 88.7 & 93.6 & 87.4 & 92.6 \\
    STEVE & 97.9 & 98.3 & 97.1 & 98.8 & 97.9 & 95.5 & 98.8 & 96.4 & 98.7 & 99.0 & 97.1 & 99.2 \\
    DINOSAURv2 & 94.4 & 94.7 & 92.7 & 97.6 & 97.5 & 93.3 & 98.5 & 78.0 & 96.8 & 98.3 & 78.0 & 98.6 \\
    \midrule
    VAE & 76.8 & 82.5 & 76.3 & 89.8 & 87.6 & 79.9 & 66.2 & 71.5 & 74.9 & 65.3 & 68.8 & 83.7 \\
    \bottomrule
  \end{tabular}
  \end{scriptsize}
\end{table*}

\begin{table*}[h]
  \centering
  \caption{Average accuracies on VQA-v2 when using T-2 as the downstream model. For pre-trained models, only one seed is available. For other models, the results are aggregated over 3 random seeds.}
  \label{tab:acc_vqav2}
  \begin{scriptsize}
  \begin{tabular}{lc|cc}
    \toprule
    \textbf{Model} & \textbf{Overall} & \textbf{Number} & \textbf{Yes/No} \\
    \midrule
    DINOv2 & 58.4 & 37.8 & 64.3 \\
    MAE & 56.5 & 36.8 & 62.2 \\
    CLIP & 57.2 & 36.6 & 63.2 \\
    VQ-AE & 53.2 & 30.0 & 59.9 \\
    KL-AE & 54.2 & 30.0 & 61.2 \\
    ResNet50 & 53.3 & 30.1 & 59.9 \\
    \midrule
    ResNet SA & 52.6 & 30.1 & 59.1 \\
    STEVE & 53.8 & 29.5 & 60.8 \\
    DINOSAURv2 & 61.3 & 41.7 & 67.0 \\
    \bottomrule
  \end{tabular}
  \end{scriptsize}
\end{table*}

\begin{table*}[h]
  \centering
  \caption{Average accuracies on GQA when using T-15 as the downstream model. For pre-trained models, only one seed is available. For other models, the results are aggregated over 3 random seeds.}
  \label{tab:acc_gqa}
  \begin{scriptsize}
  \begin{tabular}{lc|ccccc}
    \toprule
    \textbf{Model} & \textbf{Overall} & \textbf{Verify} & \textbf{Query} & \textbf{Choose} & \textbf{Logical} & \textbf{Compare}  \\
    \midrule
    DINOv2 & 51.3 & 71.9 & 39.2 & 66.2 & 61.6 & 59.3 \\
    MAE & 47.1 & 66.9 & 32.9 & 57.7 & 61.4 & 57.6 \\
    CLIP & 47.3 & 69.9 & 32.6 & 57.0 & 60.8 & 62.0 \\
    VQ-AE & 40.6 & 64.1 & 21.3 & 56.8 & 61.3 & 58.2 \\
    KL-AE & 41.7 & 64.7 & 24.3 & 56.7 & 62.5 & 55.5 \\
    \midrule
    ResNet50 & 44.7 & 67.2 & 28.7 & 57.8 & 63.3 & 56.2 \\
    ResNet SA & 40.8 & 66.2 & 22.3 & 55.6 & 61.3 & 59.7 \\
    STEVE & 41.6 & 65.0 & 25.1 & 57.7 & 61.4 & 56.1 \\
    DINOSAURv2 & 51.1 & 70.6 & 37.4 & 66.2 & 63.0 & 56.6 \\
    \bottomrule
  \end{tabular}
  \end{scriptsize}
\end{table*}

\end{document}